\documentclass[runningheads]{llncs}
\usepackage[T1]{fontenc}
\usepackage{graphicx}
\usepackage{booktabs}
\usepackage[misc]{ifsym}
\newcommand{\corr}{(\Letter)}
\usepackage{hyperref}
\usepackage{xcolor}
\usepackage{graphicx} 
\usepackage{tikz}
\usepackage{pdfpages}        
\usepackage{subcaption}
\usepackage{algorithm}
\usepackage[noend]{algorithmic}
\usepackage{float}
\usepackage{amsmath}
\usepackage{multicol}
\usepackage{multirow}
\usepackage{array}
\usetikzlibrary{tikzmark}
\usepackage{pgfplots} 
\usepackage{diagbox}
\usepackage{forest}
\usepackage{makecell}
\usepackage{wrapfig}
\usepackage[normalem]{ulem}

\usepackage[colorinlistoftodos]{todonotes}

\newcommand{\ptitle}[1]{{\bf#1.}}

\newcommand{\InfoClus}{\ensuremath{\texttt{InfoClus}}}
\newcommand{\Cyto}{\ensuremath{\texttt{Cytometry}}}
\newcommand{\German}{\ensuremath{\texttt{GSE}}}
\newcommand{\Mushroom}{\ensuremath{\texttt{Mushroom}}}

\newcommand{\pwx}{\textit{PwX}}
\newcommand{\infoc}{\textit{information content}}
\newcommand{\xratio}{\textit{explanation ratio}}

\begin{document}

\title{InfoClus: Informative Clustering of High-dimensional Data Embeddings}

\titlerunning{Informative Clustering of High-dimensional Data Embeddings}

\author{Fuyin Lai\inst{1} \corr \and
Edith Heiter\inst{1} \and
Guillaume Bied\inst{1} \and
Jefrey Lijffijt\inst{1} 
} 


\institute{Ghent University, Belgium \email{\{fuyin.lai, edith.heiter, guillaume.bied, jefrey.lijffijt\}@ugent.be}
}

\maketitle              

\begin{abstract}
Developing an understanding of high-dimensional data can be facilitated by visualizing that data using dimensionality reduction. However, the low-dimensional embeddings are often difficult to interpret.
To facilitate the exploration and interpretation of low-dimensional embeddings, we introduce a new concept named \emph{partitioning with explanations}. The idea is to partition the data shown through the embedding into groups, each of which is given a sparse explanation using the original high-dimensional attributes. 
We introduce an objective function that quantifies how much we can learn through observing the explanations of the data partitioning, using information theory, and also how complex the explanations are. Through parameterization of the complexity, we can tune the solutions towards the desired granularity. 
We propose \InfoClus, which optimizes the partitioning and explanations jointly, through greedy search constrained over a hierarchical clustering.
We conduct a qualitative and quantitative analysis of \InfoClus\ on three data sets. We contrast the results on the Cytometry data with
published manual analysis results, and compare with two other recent methods for explaining embeddings (RVX and VERA).
These comparisons highlight that \InfoClus\ has distinct advantages over existing procedures and methods. We find that \InfoClus\ can automatically create good starting points for the analysis of dimensionality-reduction-based scatter plots.
\end{abstract}

\section{Introduction\label{introduction}}
\ptitle{Background} Dimensionality-reduction (DR) methods are widely employed to project high-dimensional data into a two-dimensional space, such that it can be visualized. Popular methods are t-SNE \cite{maaten2008} and UMAP \cite{mcinnes2018}, which are DR methods that are distance-based, local-focused, non-linear, and transductive (terms explained in following paragraphs). Their popularity stems from their effectiveness to retain the local structure of the data, often in the form of cluster structure, which they manage by flexibly warping larger distances. However, this flexibility bring several challenges in the interpretation of embeddings \cite{chatzimparmpas2020,faust2019,sedlmaier2012,stahnke2015}:

The lack of interpretability has its origin in several factors: (1) the methods are \emph{distance-based}, meaning they operate only on the distance matrix and not directly on the feature values. The distances summarize all features with equal weight and thus there is no preference for sparsity in relations between points in the embedding. (2) They are \emph{non-linear and transductive} (i.e., non-parametric):\\the placement of points in the embedding is entirely free and not a projection. The axes are not interpretable, the embeddings are sensitive to specificities of the data and many XAI methods do not apply (requiring, e.g., the existence of gradients). (3) They are \emph{local-focused} meaning that the distances are distorted. For example, t-SNE also uses a point-specific parameter $\sigma_i$ to factor out density differences across the data, which can lead to counterintuitive distortions.

\ptitle{Idea behind InfoClus} Rather than aiming to fix these problems, we want to facilitate interpretation of the embeddings, agnostic to the dimensionality reduction algorithm. More specifically, InfoClus\ is designed to provide a useful starting point for exploring and interpreting of low-dimensional embeddings.

\begin{wrapfigure}[13]{r}{0.61\textwidth}
    \vspace{-28pt}
    \centering
    \includegraphics[page=2,width=0.60\textwidth]{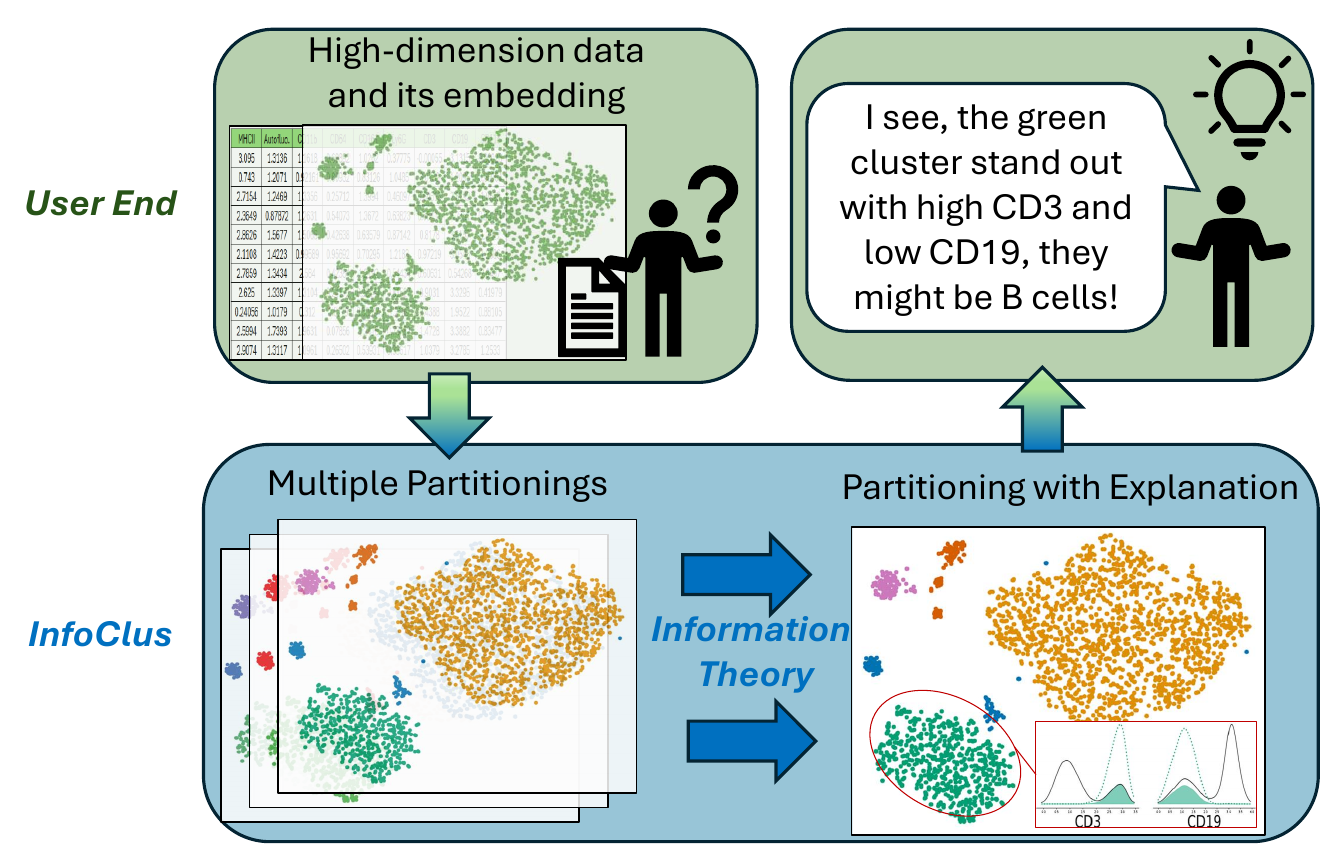}
    \caption{\InfoClus\ workflow}
    \label{fig:overview_diagram}
\end{wrapfigure}

As discussed more in-depth in the related work section below, existing systems enable users to `explain' hand-selected clusters, or give an overview of the impact of a specific attribute. They rely on the user selecting what to look at. In contrast, the idea behind \InfoClus\ is to automatically partition the embeddings into coherent clusters that have a sparse explanation. In other words, to find a clustering that is (1) cohesive on the embedding and (2) such that the points in each cluster share a relatively simple explanation. The explanations that we consider here provide information about how the points in each cluster are different from the data as a whole. Figure~\ref{fig:overview_diagram} shows a diagram that illustrates the \InfoClus\ approach.

\InfoClus\ is a parame\-trized approach for finding good \emph{partitionings with explanations} (\pwx s), given a high-dimensional dataset and a low-dimensional embedding. The explanations for a cluster are those that provide most information, as quantified through information theory. 

\ptitle{How InfoClus works}
We first formalize the problem setting (Section~\ref{sec:method}) and define \emph{partitioning with explanations}.
We then use information theory to quantify the amount of information communicated with a \pwx\ and we propose a quantification for the `complexity' of a \pwx\ (Section~\ref{sec:xr}). A \pwx\ that conveys more information and/or has a lower complexity is better. The complexity is parameterized, enabling users to tune the algorithm to yield simpler or more complex solutions, as desired.

In Section~\ref{sec:search_space}, we consider how to algorithmically find good \pwx s. Maximizing the informativeness will lead to \pwx s that are cohesive in the high-dimensional space, but we need to also ensure that the clustering is cohesive on the two-dimensional embedding. Hence, we constrain possible solutions to clusterings on the embedding from existing clustering methods. We consider the use of hierarchical and $k$-means clustering. Such constraints on solutions also have the advantage of limiting the search space of possible partitionings. 

For hierarchical clustering, the number of possible partitionings is still very large and optimizing informativeness is not straightforward. We show this problem can be optimized efficiently and effectively using a greedy search algorithm that cuts off branches from the tree induced by the hierarchical clustering.
This yields an algorithm that empirically runs in linear time in the number of data points and attributes, per iteration that adds a cluster to the solution. 

\ptitle{Evaluation} We evaluate the approach using three different datasets, and study the results qualitatively and quantitatively (Section~\ref{sec:exp}). We include a comparison with the two closest existing methods (RVX \cite{thijssen2023} and VERA \cite{policar2024}). We consider a Cytometry dataset for which we have a manual expert labeling available based on visualization of the data over feature pairs. 
We compare the features used in the experts' approach and by \InfoClus.
We additionally study a dataset with demographic and voting statistics, and the UCI Mushroom data.

Finally, we study the impact of the hyperparameters (Section~\ref{sec:hyperparameters}) and the runtime of \InfoClus\ (Section~\ref{sec:scalibility}). We conclude the paper with a discussion of the capabilities and limitations of the approach (Section~\ref{sec:discussion}).


\ptitle{Reproducibility} All code and results will be made available open source through GitHub upon publication. Code is shared also through CMT for review.

\section{Related work}
\label{sec:related_work}

For brevity, we discuss only the most important examples of work on the more general topic of explaining embeddings and then also the most similar  methods.

\ptitle{Related work on explaining embeddings} Several approaches have been developed for explaining the specific positions of points in an embedding, e.g., using inverse projections \cite{cavallo2018,espadoto2023} or counterfactuals \cite{artelt2022}. Another popular approach to explaining embeddings is feature-centric. We can show the values for an individual feature, but patterns may be found more easily through aggregates such as contour lines \cite{faust2019}. Similarly, one may construct `rangesets' for class labels \cite{sohns2022}.

The above approaches can provide insight into the impact of each attribute on the embedding, but do not lead to discovery of the most important features. The second type of approach that has been explored is to start from a user-selected cluster and show how this cluster stands out against the overall data or against another cluster. Stahnke et al. introduced this as `probing' of projections \cite{stahnke2015}. In their paper, they showed the distributions of all attributes to users. Recent papers have considered the use of decision trees \cite{bibal2021}, boosting machines \cite{salmanian2024}, and predicate logic \cite{montambault2024} to provide sparse explanations for manually-selected clusters (i.e., explanations using a small subset of features). The latter approach named DimBridge \cite{montambault2024} is notable for also supporting the explanation of hand-drawn lines, by constructing a regression model for the progression along the line.

Finally, there are other DR algorithms, including approaches that aim to directly create coherent---and thus arguably more interpretable---clusters in the embedding. For example, by integrating label information in classification settings \cite{debodt2019} or contrastive DR to visualise a clustering \cite{xia2024}. However, in this paper we do not aim to evaluate DR approaches and the approach is agnostic to the DR algorithm. We aim to facilitate the interpretation of any given embedding.

\ptitle{Methods similar to \InfoClus} There are no approaches directly similar to \InfoClus\ that partition an embedding of the data and generate explanations for the partitioning. The most closely related work is Visual Explanations via Region Annotation (VERA) by Policar and Zupan \cite{policar2024}. VERA provides two different types of `explanations' for embeddings: descriptive and contrastive. They both are built up by finding clusters on the embedding for each data attribute separately. Contrastive explanations select an attribute and value splits that lead to cohesive clusters on the embedding. Descriptive explanations are constructed by merging clusters across attributes that are similar, and then selecting a set that has small overlap.

Another method that does not explicitly seek clusters but may end up giving a similar result to \InfoClus\ is by Thijssen et al. \cite{thijssen2023}. We will refer to it as Relative Value eXplanations (RVX) and $\mathrm{RV_\sigma X}$ for the variance-based variant. RVX provides an explanation in the form of a feature per data point that stands out most, in terms of value or variance. It employs smoothing, leading to visual clusters often sharing a single most-outstanding attribute. We include an empirical comparison between \InfoClus, VERA, and RVX in Section~\ref{sec:exp}.

\section{InfoClus -- Step by Step\label{sec:method}}

\ptitle{Preliminaries} We assume given a dataset with $n$

\begin{wrapfigure}[8]{r}{0.3\textwidth}
\vspace{-36pt}
\centering
        \begin{tikzpicture}
            \begin{axis}[
                axis lines = middle,
                xlabel = $a_1$,
                ylabel = $a_2$,
                xlabel style={yshift=-3ex, xshift=1ex},
                ylabel style={xshift=-3.5ex},
                width=4.5cm,
                height=4.5cm,
                xmin=0, xmax=8,
                ymin=0, ymax=8,
                xtick={0,2,4,6},
                ytick={0,2,4,6},
                scale=0.8,
            ]
            \addplot[
                only marks,
                mark=*,
                color=green!50!black,
                mark size=1.2pt,
            ] coordinates {
                (4,1)
                (5,2)
                (6,1)
                (7,2)
                (1,4)
                (2,5)
                (1,6)
                (2,7)
            };
            \node at (axis cs:4,1) [anchor=east] {$p_0$};
            \node at (axis cs:5,2) [anchor=east] {$p_1$};
            \node at (axis cs:6,1) [anchor=east] {$p_2$};
            \node at (axis cs:7,2) [anchor=east] {$p_3$};
            \node at (axis cs:1,4) [anchor=west] {$p_4$};
            \node at (axis cs:2,5) [anchor=west] {$p_5$};
            \node at (axis cs:1,6) [anchor=west] {$p_6$};
            \node at (axis cs:2,7) [anchor=west] {$p_7$}; 
            \end{axis}
        \end{tikzpicture}  
    \caption{Example dataset.}
    \label{fig:example}
\end{wrapfigure}
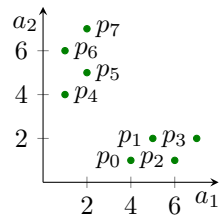
\noindent points $X = \{x_1, x_2, \cdots, x_n\}$, $m$ attributes $A = \{a_1, a_2, \cdots, a_m\}$, and its corresponding two-dimensio\-nal embedding $Y$. 
We refer to a \emph{partitioning} of $Y$ as a list of sets $\mathcal{C} = [c_1,\ldots,c_r]$ such that each point in $Y$ belongs to exactly one $c_i$. We refer to an \emph{explanation} for a \emph{partitioning} as choice of representative attributes and their corresponding distribution for each cluster in the partitioning.



\begin{definition}{\textrm
    A \emph{partitioning with explanations} (\pwx) of dataset $X$ is a partitioning $\mathcal{C}$ of its embedding $Y$ with associated explanations, represented as a tuple $(\mathcal{C}, \mathcal{E})$, where $\mathcal{C} = [c_1,\ldots,c_r], \mathcal{E} = [e_1, \ldots, e_r]$. $\mathcal{E}$ denotes the list of explanations, where $e_i = \{a_i^1, a_i^2, \ldots \} \subseteq A$ is the set of attributes selected to explain cluster $c_i$.
    }
\label{def:pwx}
\end{definition}

Fig. \ref{fig:example} shows a small dataset (8 points, 2 attributes) that will serve as a running example to illustrate the concepts in this section (note here that $Y=X$).

\emph{Example.} $(\mathcal{C'},\mathcal{E'})$
with $\mathcal{C'}=[c_0:\{p_0,p_1,p_2,p_3\}, c_1:\{p_4,p_5,p_6,p_7\}]$, $\mathcal{E'}=[e_1:\{a_1\},e_2:\{a_2\}]$ is a \pwx\ of the dataset shown in Fig. \ref{fig:example}.


\ptitle{Section outline} Section \ref{sec:xr} presents measures for how much information users may learn from a \pwx\ and the complexity of a \pwx, which together define what is an informative clustering. 
Section \ref{sec:search_space} introduces \InfoClus, a methodology aimed at finding coherent and informative \pwx s.





\subsection{Explanation Ratio\label{sec:xr}}


We assume that the goal of interpreting the embedding is to learn about the distribution of the data. The visualization of the embedding enables users to identify groups of points that form a cluster. For example, t-SNE indeed tends to produce embeddings with meaningful cluster structure \cite{kobak2019}. Hence, we can model the interpretation as a communication process where the user learns about the data through visualizations of distributions for clusters present in the embedding. 

To arrive at a quantification for what a user learns, we take inspiration from the framework on subjective interestingness \cite{debie2011}. We can quantify how much a user learns using information theory, by quantifying the reduction in uncertainty a user has about the data. This means we have to make an assumption about what the user knows already---which is known as the \emph{prior distribution}---and what the user knows after observation of a \pwx.

Following De Bie \cite{debie2011}, the \infoc\ of an observation is given by the surprisal of this observation: $IC(X) = -\log P(X)$. Equivalently, but from a different perspective, if we consider how much we learn by updating a prior distribution to a posterior, this is given by the KL-divergence between the two distributions. The amount of information that we learn by updating a prior to a more specific posterior distribution is quantified as follows:



\begin{definition}{\textrm
    The \emph{information content} $I$ for a \pwx\ is the sum of \emph{information content} for each cluster $c_i$ and each attribute $a_i^j$ in the \pwx, which is given by
\begin{align}\label{eqn:I}
    I_i^j = |c_i| * D_{KL}\big(P_i^j \Vert Q^j\big)\,,
\end{align}
    where $|c_i|$ is the size of the cluster, $P_i^j$ is the distribution of attribute $a^j$ in cluster $c_i$, and $Q^j$ is the distribution of attribute $a^j$ in the full dataset.
}
\label{def:ic}
\end{definition}


Here we assumed that the prior information a user has about the data are the marginal distributions for each attribute in the data. We present the distributions for clusters in visual form, but we assume users learn about the mean and variance statistics only, but with infinite precision. In practice, we may not know either, but the approach may still be helpful even if the details of how we model the beliefs of the user are wrong.


In \InfoClus, 
we model real-valued attributes with Gaussian distributions, and categorical attributes with a categorical distribution as a default.
The implementation of \InfoClus\  supports both numerical and categorical attributes, but not mixed data, because the KL-divergence is defined for discrete and continuous variables, but they are not comparable.

\textit{Example.}
We assume continuous attribute $a_2$ follows a Gaussian distribution. By computing the empirical mean and variance in the full dataset and the cluster $c_0$ respectively, we have $a_2$ follows $\mathcal{N}(3.5,4.75)$ on the full dataset and  $\mathcal{N}(1.5,0.25)$ in $c_0$ of \pwx\ $(\mathcal{C}',\mathcal{E}')$.
The information carried by $a_2$ in $c_0$ is: $I_0^1$ = $4 \times D_{KL}(\mathcal{N}(1.5,0.25)\Vert\mathcal{N}(3.5,4.75))$ = $7.38$.

Yet, the more attributes are selected to inform a user about a partitioning, the more effort is needed for users to understand them. This leads us to define the following notion of complexity, with $\alpha$ and $\beta$ parameterizing a user's learning sensitivity for information:

\begin{definition}{\textrm
    The \emph{complexity} of a \pwx\ $(\mathcal{C},\mathcal{E})$ is
\begin{align}\label{eqn:dl}
        \alpha + (\sum_{i=1}^r{\sum_{j=1}^{|e_i|}{|a_i^j|}})^\beta\,,
\end{align}
    where $a_i^j$ is an attribute belonging to $e_i\in \mathcal{E}$, and $|a_i^j|$ refers to the count of statistics needed to describe the distribution $a_i^j$ follows. $\alpha$ and $\beta$ are hyper-parameters.
} 
\end{definition}

\textit{Example.} Assuming attributes $a_1,a_2$ in Fig. \ref{fig:example} follow a Gaussian distribution, we have $|a_1|=|a_2|=2$, since for a Gaussian distribution, the mean and variance statistics define the probability distribution. Then, given $\alpha=1,\beta=2$, the \emph{complexity} for \pwx\ $(\mathcal{C}',\mathcal{E}')$ is $1+4^2=17$. 


To balance the information carried by a \pwx\ and the human cognitive effort necessary to interpret it, we introduce the concept of an explanation ratio:

\begin{definition}{\textrm
    The \emph{explanation ratio} $R_{\alpha,\beta}$ of a \pwx\ $(\mathcal{C}, \mathcal{E})$ with $r$ clusters is
    \begin{align}\label{eqn:xr}
             R_{\alpha, \beta}(\mathcal{C}, \mathcal{E}) = \frac{\sum_{i=1}^r{\sum_{j=1}^{|e_i|}{I_i^j}}}{\alpha + (\sum_{i=1}^r{\sum_{j=1}^{|e_i|}{|a_i^j|}})^\beta}\ ,
    \end{align}
where $i$ indexes the clusters and $j$ the attributes of explanation $e_i$.
}
\label{def:r}
\end{definition} 

\textit{Example.} Given $\alpha=1,\beta=2$, the \xratio\ for \pwx\ $(\mathcal{C'},\mathcal{E'})$ is $R_{1,2}(\mathcal{C'},\mathcal{E'})=({I_0^1+I_1^0})/({1+(|y|+|x|)^2})=0.87 $.

\paragraph{\textbf{Problem statement.}}  \InfoClus\ aims to partition the dataset into clusters that are cohesive in the embedding and the high-dimensional space. The clustering is shown on the embedding; information on the high-dimensional space is provided through a few selected attributes. The partitioning and explanations are chosen such that they maximize the \xratio\, aiming to optimize the ratio of the amount of information provided over the effort required to interpret it.

\subsection{InfoClus: finding cohesive and informative PwXs efficiently\label{sec:search_space}}

\InfoClus\ seeks to partition the embeddings into visually coherent clusters with high \xratio. To do so, existing clustering algorithms (e.g. hierarchical clustering or $k$-means) are used to generate candidate partitionings (ensuring visual coherence), among which the \pwx\ with highest \xratio\ is retained. 
In the following, we focus on hierarchical clustering, which we advise using as a default candidate generator.

Hierarchical clustering strikes a suitable balance with a large number of possible partitionings, but since these are structured in a tree we can still optimize efficiently over the space of candidates\footnote{Its use will be compared to that of $k$-means in Section \ref{sec:case_study}.}. We proceed to detail the procedure used to navigate the search space for partitionings implied by hierarchical clustering, and sketch algorithmic details.



\begin{figure}[h]
    \vspace{-16pt}
    \centering
    \begin{tabular}{c|c}
        \begin{forest}
        for tree={
            edge path={\noexpand\path[\forestoption{edge}] (\forestOve{\forestove{@parent}}{name}.parent anchor) -- +(0,-12pt)-| (\forestove{name}.child anchor)\forestoption{edge label};},
            l sep=1mm,
            s sep=1mm
        }
          [Root
            [10, draw, circle,
              [9, draw, circle,
                [8 , draw, circle, [$p_0$, text=green!50!black, tier=terminal] [$p_1$, text=green!50!black, tier=terminal] ]
                [$p_2$, text=green!50!black, tier=terminal]
              ]
              [$p_3$, text=green!50!black, tier=terminal]
            ]
            [13, draw, circle,
                [12 , draw, circle,
                    [11, draw, circle, [$p_4$, text=green!50!black, tier=terminal] [$p_5$, text=green!50!black, tier=terminal]]
                    [$p_6$, text=green!50!black, tier=terminal]
                ]
                [$p_7$, text=green!50!black, tier=terminal]
            ]
          ]
        \end{forest}
        &  
        \includegraphics[width=0.5\textwidth]{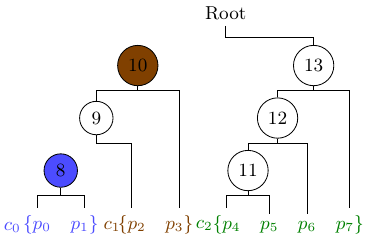}
        \\
        (a) Dendrogram & \makecell{(b) Partitioning $\mathcal{C}=[c_0,c_1,c_2]$ \\ generated by splitting nodes 8, 10}
    \end{tabular}
    \caption{How hierarchical clustering works as a candidate partitioning generator.
    \label{fig:dendrogram}}
    \vspace{-12pt}
\end{figure}
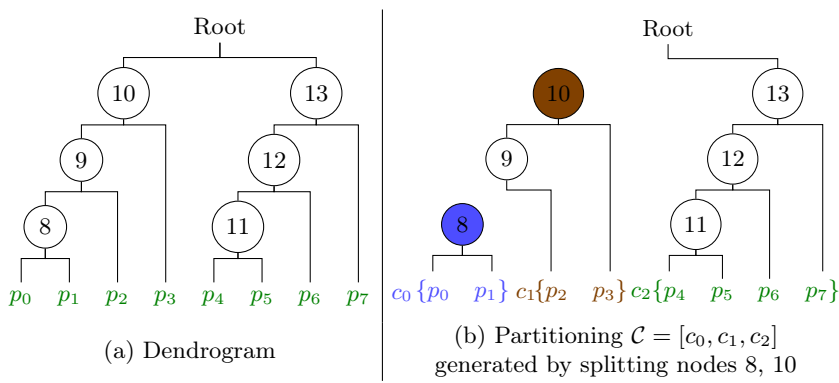

\ptitle{Search for Partitionings} Hierarchical clustering constructs a hierarchy of clusters, which may be seen as organizing data in a tree structure (dendrogram). For instance, agglomerative hierarchical clustering builds a tree from the ground up: each observation initially defines its own cluster, and clusters are progressively greedily merged based on a chosen linkage. We constrain our search space to partitionings compatible with the hierarchical clustering. 
More specifically, rather than relying solely on partitionings formed by slicing the dendrogram at a specific level, we consider a compatible partitioning with $r$ clusters as consisting of $r-1$ clusters obtained from cutting nodes at any level of the dendrogram, and the final cluster encompassing the remaining points.
For instance, Fig. \ref{fig:dendrogram} illustrates a partitioning $\mathcal{C}$ compatible with a dendrogram in our toy example. 

Yet, the number of partitionings compatible with a dendrogram is too large for enumeration\footnote{There are 56 different partitionings in a dendrogram generated by a 8 points dataset.}. Accordingly, \InfoClus\ resorts to a greedy strategy to iteratively find a sequence of \pwx s with an increasing number of clusters and high \xratio s, where each \pwx\ is obtained by splitting one node extra from the previous \pwx.
More precisely, at iteration $k$, \InfoClus\ considers partitionings with $k+1$ clusters by enumerating all \pwx s formed by cutting any subtree of the \pwx\ we chosen at $k-1$ iteration. The partitioning with highest \xratio\ is selected as the \pwx\ at iteration $k$.
Thus, $K$ iterations of the greedy procedure yield a sequence of $K$ $PwX$s of size 2, $\ldots$, $K+1$. The \pwx\ in this sequence with highest \xratio\ is the one finally selected.

\textit{Example.} Fig. \ref{fig:dendrogram}a presents a dendrogram generated for the example dataset by hierarchical clustering. Given $\alpha=1,\beta=2$, we have 5 candidate partitionings with 2 clusters: those obtained by splitting subtrees with root of node 8, 9, 10, 11 or 12. After computing the best explanations for each of these partitionings, we find the associated \xratio s to be $0.36,0.52,0.87,0.36$ and $0.52$ respectively.  The partitioning generated by splitting node 10 is the best one with $R_{1,2}=0.87$; this corresponds to \pwx\ $(\mathcal{C}',\mathcal{E}')$. The next iteration considers partitionings with 3 clusters that could be obtained by splitting one more node (node 8, 9, 11 or 12) out of \pwx\ $(\mathcal{C}',\mathcal{E}')$.
For instance, splitting node 8 out of \pwx\ $(\mathcal{C}',\mathcal{E}')$ would lead to a new partitioning $[c_0:\{p_0,p_1\}, c_1:\{p_2,p_3\}, c_2:\{p_4...p_7\}]$.

\textit{Time budget.} In the following, we call an \textbf{iteration} the process through which \InfoClus\ tries different nodes to find a good partitioning with $k$ clusters based on the partitioning with $k-1$ previously selected clusters. Iterations can be conducted as long as there are possible nodes to consider, but a partitioning of a dataset with almost as many clusters as points might not be desirable.
Therefore, in the implementation of \InfoClus\, we introduce a time budget $t$, which imposes \InfoClus\ to terminate before time $t$.

\ptitle{Search for Explanations}
The best explanations for a partitioning in a dataset with homogeneous attributes (e.g. only numeric or categorical) can be found by applying a greedy strategy in each cluster: iteratively adding attributes with highest \infoc\ to a cluster's explanations until the \xratio\ $R_{\alpha,\beta}$ decreases when a new attribute is added.
In the code implementation, the number of attributes selected for each cluster can be controlled by hyper-parameters \textit{minatt} and \textit{maxatt}, which serve as lower and upper bounds respectively for the number of attributes used to explain each cluster.

\begin{algorithm}[ht]
\caption{\textsc{InfoClus}}
\label{algo:infoclus}
    \begin{algorithmic}[1]
        \STATE \textbf{Input:} dataset $X$, its embedding $Y$, attributes $A$, \textit{minAtt}, \textit{maxAtt}, $\alpha, \beta$, and $t$
        \STATE \textbf{Output:} a \pwx\ $(\mathcal{C}, \mathcal{E})$
        \STATE Dendrogram = hierarchicalClustering($Y$)
        \STATE Nodes = getSplittingNodes(Dendrogram)\textcolor{gray}{ \% Potential splits in dendrogram}
        \STATE best$R$, best$R^1$ = 0, best$\mathcal{C}$, best$\mathcal{C}^1$ = the full dataset
        \STATE $k$ = 1
        \WHILE{$t$ has not been used up \& Nodes is not None}
            \STATE \textcolor{gray}{\% iteration starts}
            \STATE $k=k+1$
            \STATE best$R^k=0$, best$Node^k=None$
            \FOR{$Node$ in Nodes}
                \STATE $\mathcal{C}^k$ = split $Node$ out of best$\mathcal{C}^{k-1}$ to get partitioning 
                \STATE $\mathcal{E}^k$ = find best explanation for $\mathcal{C}^k$ under bounds \textit{minAtt} and \textit{maxAtt}
                \STATE compute $R_{\alpha,\beta}(\mathcal{C}^k,\mathcal{E}^k)$
                \IF{best$R^k$ $\leq$ $R_{\alpha,\beta}(\mathcal{C}^k,\mathcal{E}^k)$}
                    \STATE best$R^k=R_{\alpha,\beta}(\mathcal{C}^k,\mathcal{E}^k)$, best$\mathcal{C}^k=\mathcal{C}^k$, best$Node^k=Node$
                \ENDIF
                \IF{best$R$ $\leq$ $R_{\alpha,\beta}(\mathcal{C}^k,\mathcal{E}^k)$}
                    \STATE best$R$, best$\mathcal{C}$, best$\mathcal{E}$ = $R_{\alpha,\beta}(\mathcal{C}^k,\mathcal{E}^k)$, $\mathcal{C}^k$, $\mathcal{E}^k$
                \ENDIF
            \ENDFOR
            \STATE Update Nodes by removing best$Node^k$ and its ancestors
        \ENDWHILE
        \RETURN \pwx\ (best$\mathcal{C}$, best$\mathcal{E}$)
    \end{algorithmic}
\end{algorithm}



\textit{Example.} Given $\alpha=1,\beta=2$, consider computing the best explanation $\mathcal{E}$ for partitioning $\mathcal{C}'=[c_0:\{p_0,p_1,p_2,p_3\},c_1:\{p_4,p_5,p_6,p_7\}]$ in the example dataset (Fig. \ref{fig:example}). We first compute \infoc\ for both attributes $x, y$, and both clusters $c_0$ and $c_1$. We get $I_0^x, I_0^y, I_1^x, I_1^y = 4.25,7.38,7.38,4.25$. Next, we assign $a_2$ to $c_0$ and $a_1$ to $c_1$ to make sure that at least one attribute is assigned to each cluster. At this stage, the \xratio\ $R_{1,2}$ is $0.87$. Then we assign the next attribute ($a_1$) with highest $I$ to its corresponding cluster ($c_0$). This results in $R_{1,2}=0.51$. Since $0.51<0.87$, we stop searching, and serve $\mathcal{E}':[e_1=\{y\},e_2:\{x\}]$ as the best explanation for $\mathcal{C}'$.

\ptitle{Pseudo-code} Full pseudo-code for \InfoClus\ is shown in Algorithm \ref{algo:infoclus}.

\section{Experiments\label{sec:exp}}


In Section \ref{sec:case_study} we first present experiments on three datasets to evaluate the ability of \InfoClus\ to generate useful insights, including a comparative assessment with respect to the state of art \cite{policar2024,thijssen2023}. We then study the sensitivity of \InfoClus\ to the hyper-parameters (Section \ref{sec:hyperparameters}), and its scalability (Section \ref{sec:scalibility}).



\subsection{Case study\label{sec:case_study}}

To evaluate the usefulness of \InfoClus, we conduct case studies on three datasets covering different domains and attribute types: the \Cyto, \German\ (German Socio-Economic) and \Mushroom\ data; details of which will be introduced below.
For the \Cyto\ data, we compare insights given by \InfoClus\ to manual annotation by domain experts, to \InfoClus\ with $k$-means, and with RVX \cite{thijssen2023} and VERA \cite{policar2024}.
Embeddings studied in the case studies are generated with t-SNE. 


\ptitle{Cytometry data \cite{saeys2016}}
The \Cyto\ dataset  represents different cell types of mouse splenocytes, described by 9 columns/features recording 
the stain of different markers.
We conduct the case study on a random sample of 2500 instances from the original dataset, which contains more than 300K instances\footnote{As in \cite{saeys2016}, the analysis is conducted on a subsample for the sake of limiting t-SNE's runtime; the scalability of \InfoClus\ with larger data is investigated in Section \ref{sec:scalibility}.}.

Advantageously for this case study, the data has been annotated by domain experts through manual gating \cite{saeys2016}. This enables us to compare insights derived by \InfoClus\ with an expert procedure. 

\begin{figure*}[h!]
\vspace{-16pt}
\centering
    \begin{tabular}{c|ccccc}

        \tikzmark{expert}
        \begin{tikzpicture}[overlay, remember picture]
        \node[anchor=south west] at ([xshift=-2.5cm, yshift=-1.5cm] pic cs:expert)
        {\includegraphics[width=0.4\linewidth]{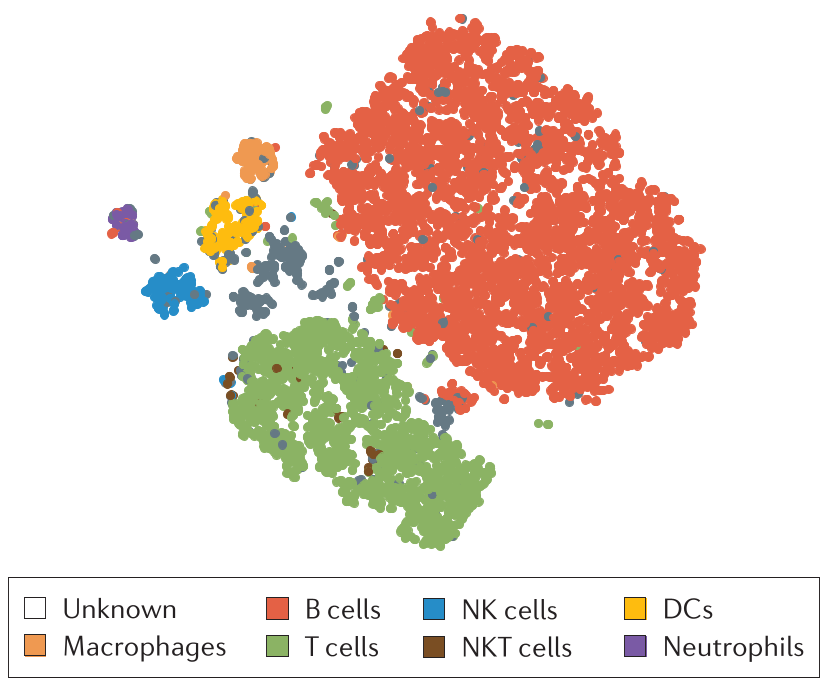}};
        \end{tikzpicture}
        & \multicolumn{5}{c}{\includegraphics[width=0.5\linewidth]{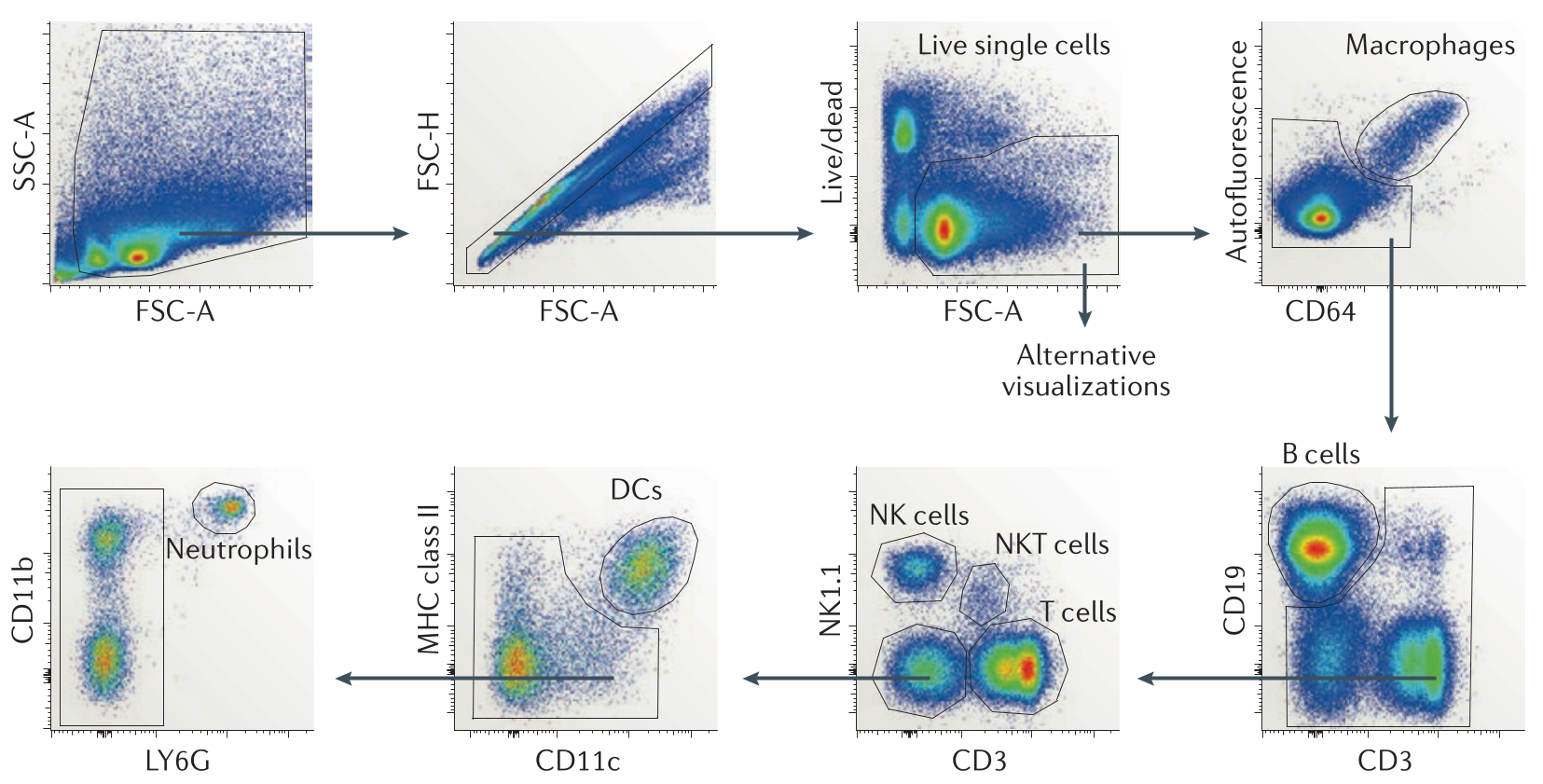}
       } \\

        & \multicolumn{5}{c}{ (b1) Manual gating process \cite{saeys2016}} \\
       
        \cline{2-6}
        & & & & & \\
        
         (a1) Labeling by manual gating \cite{saeys2016}&  \includegraphics[width=0.1\linewidth]{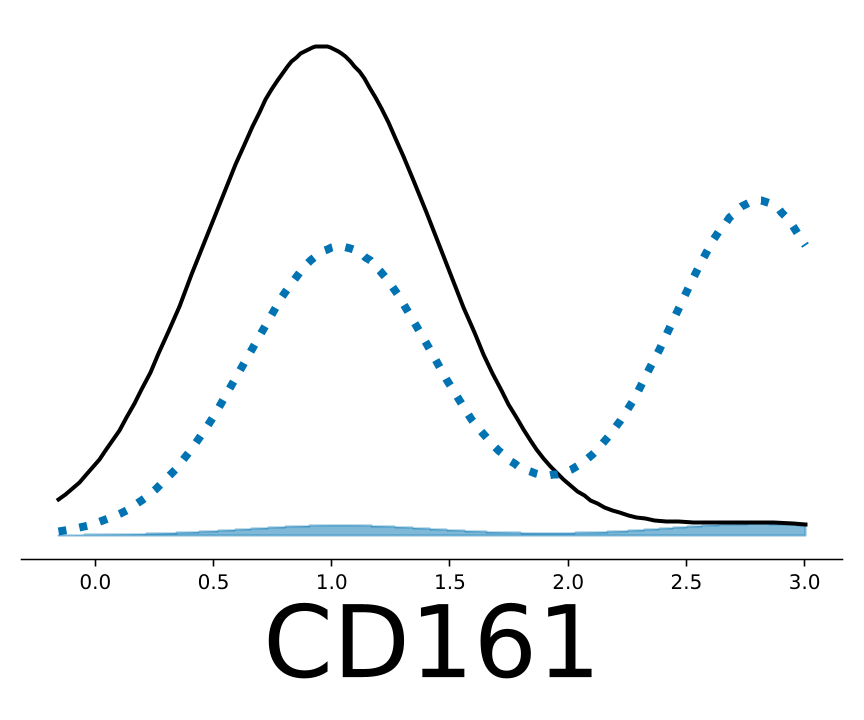}&\includegraphics[width=0.1\linewidth]{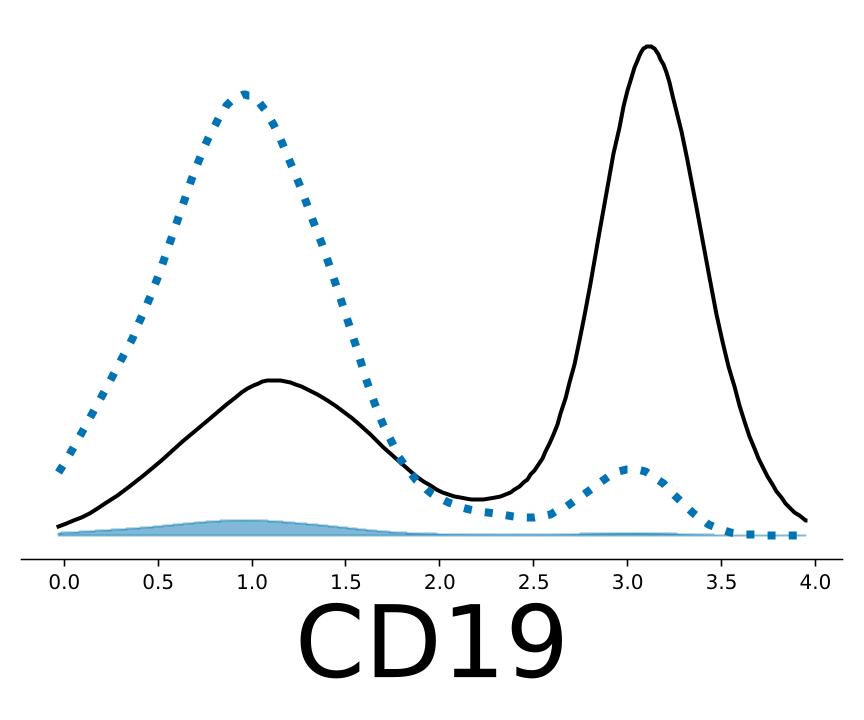} & & & \\ 
         
        \tikzmark{infoc}
        \begin{tikzpicture}[overlay, remember picture]
        \node[anchor=south west] at ([xshift=-2.5cm, yshift=-4cm] pic cs:infoc)
        {\includegraphics[width=0.4\linewidth]{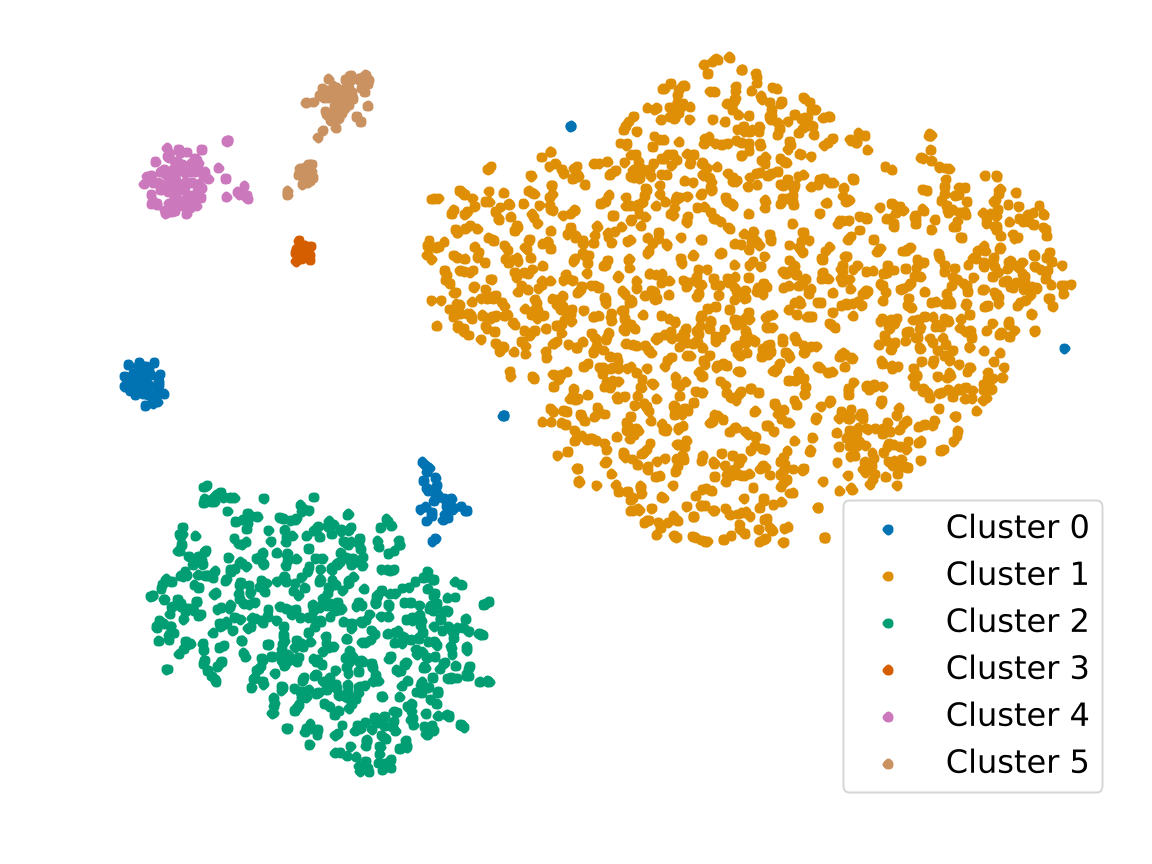}};
        \end{tikzpicture}& \includegraphics[width=0.1\linewidth]{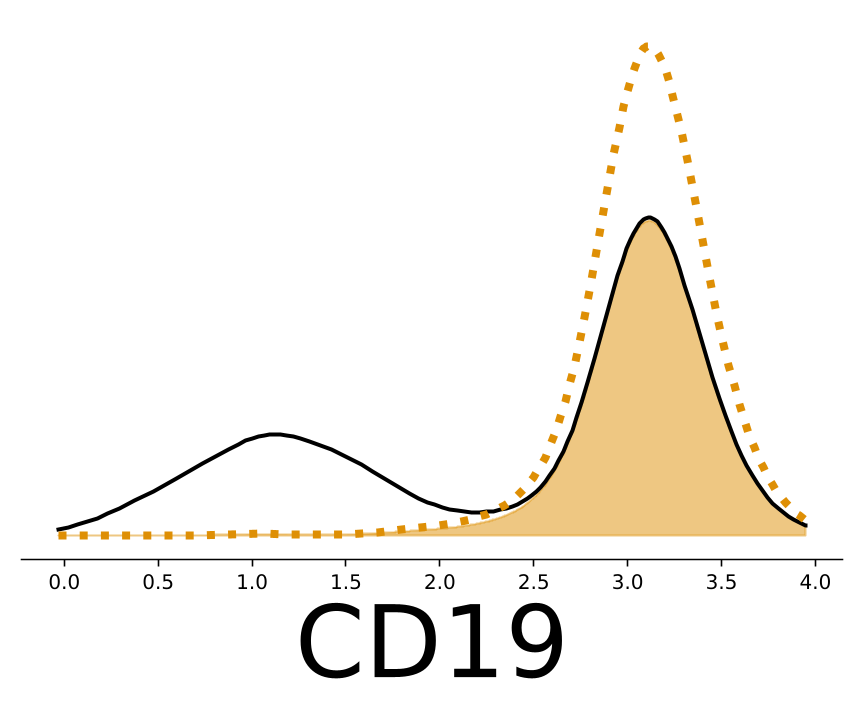} &\includegraphics[width=0.1\linewidth]{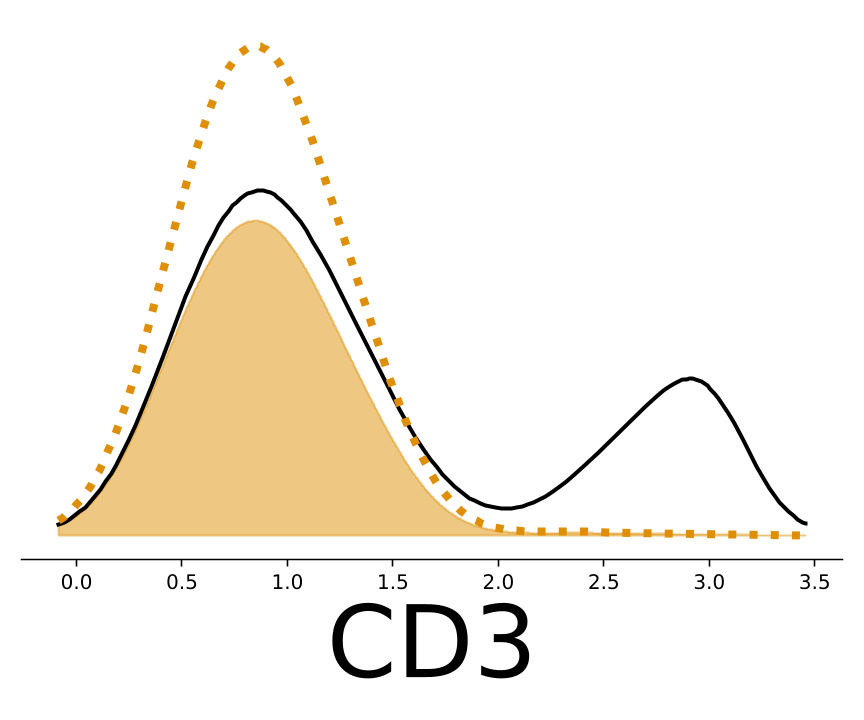} &\includegraphics[width=0.1\linewidth]{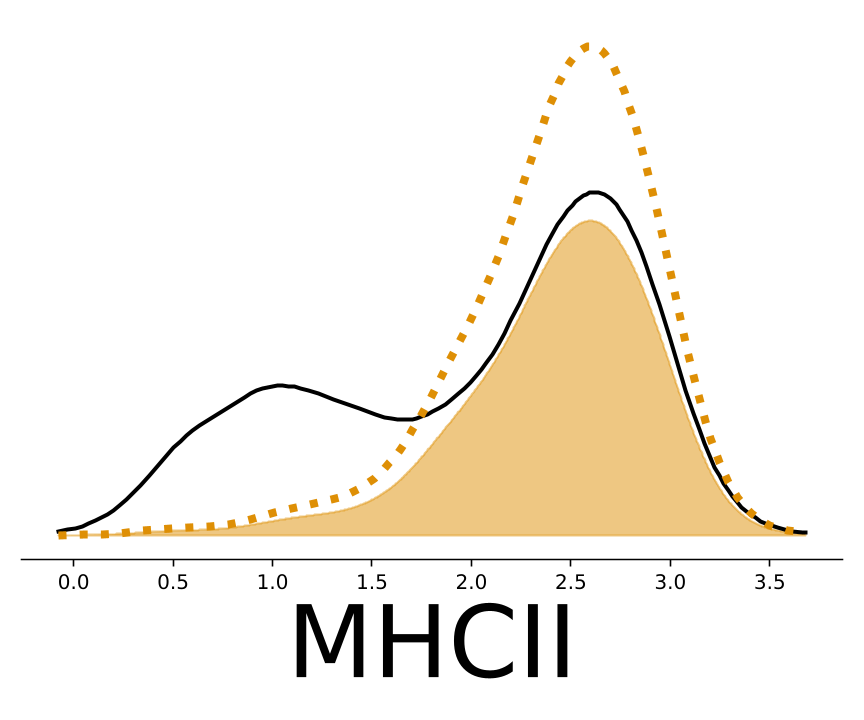} &\includegraphics[width=0.1\linewidth]{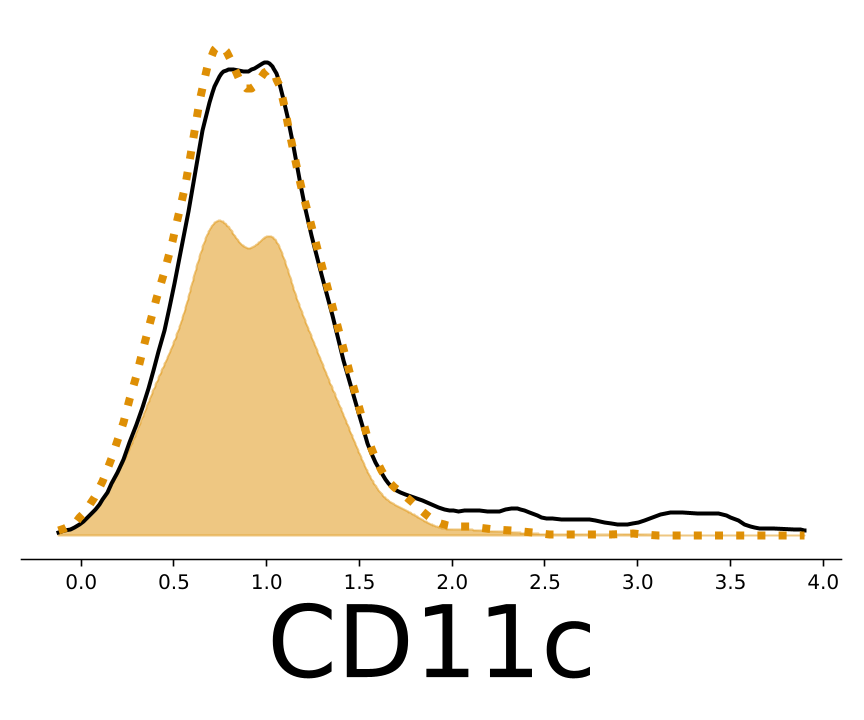} &\includegraphics[width=0.1\linewidth]{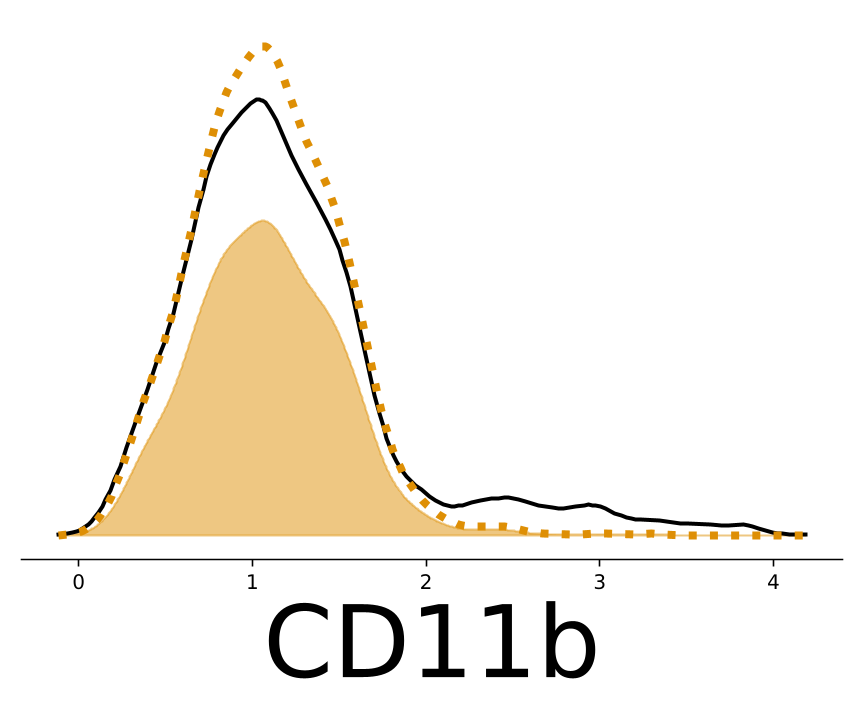} \\ 
         
         &\includegraphics[width=0.1\linewidth]{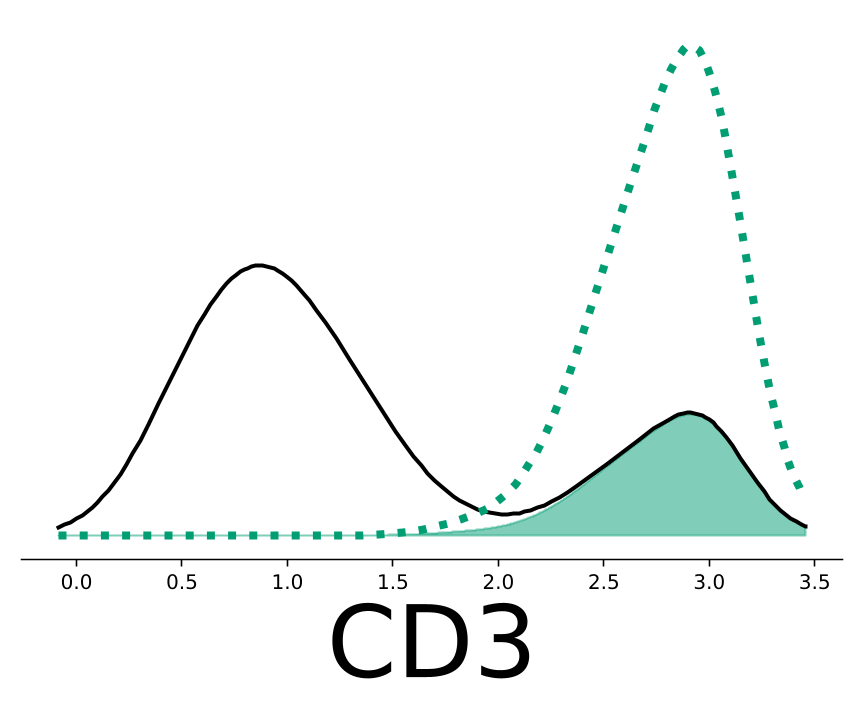} &\includegraphics[width=0.1\linewidth]{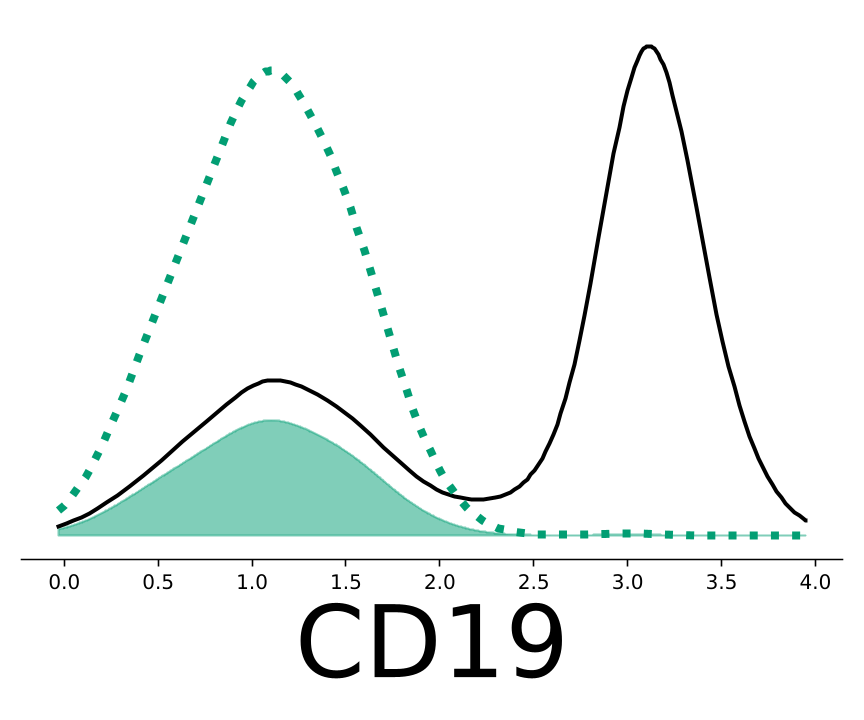} &
         \tikzmark{larged}
         \includegraphics[width=0.1\linewidth]{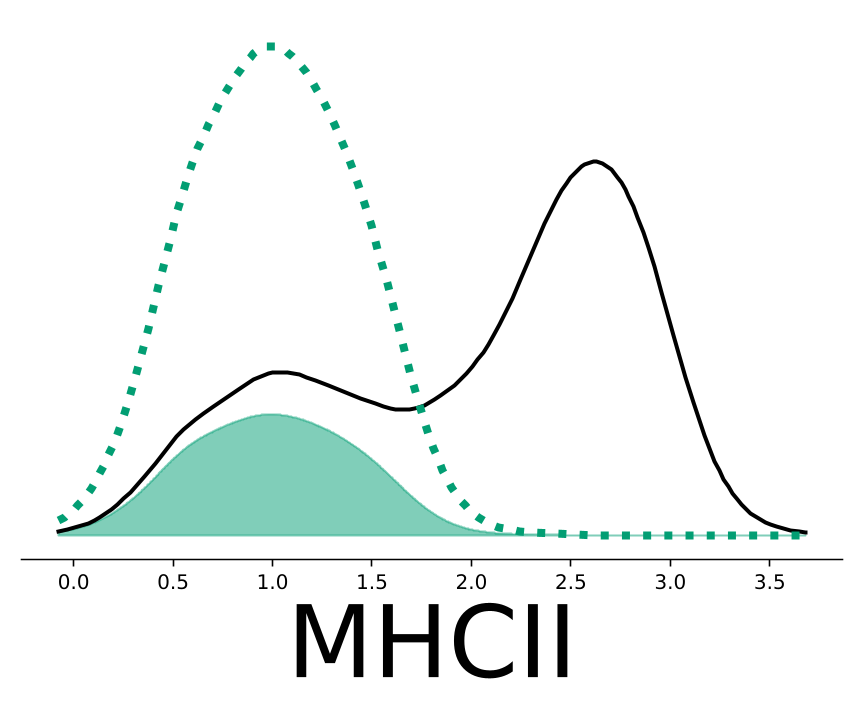}     
          \begin{tikzpicture}[overlay, remember picture]
        \node[anchor=south west] at ([xshift=0.6cm, yshift=-3cm] pic cs:larged)
        {\includegraphics[width=0.25\linewidth]{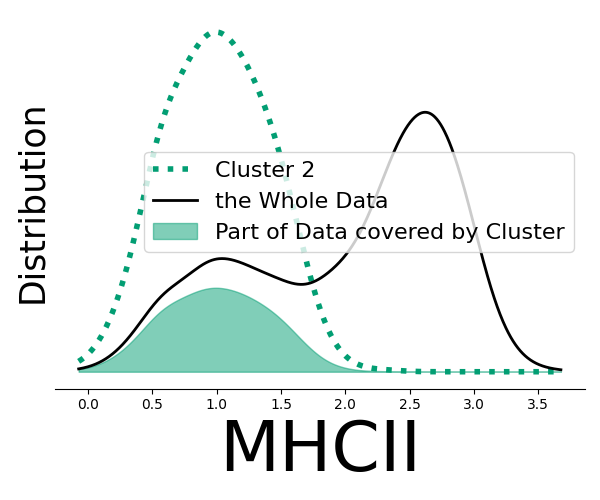}};
         \draw[thick,red, ->] (-0.7,0) .. controls +(0,0) and +(0,0) .. (0.1,-0.4);
        \end{tikzpicture}
         &\includegraphics[width=0.1\linewidth]{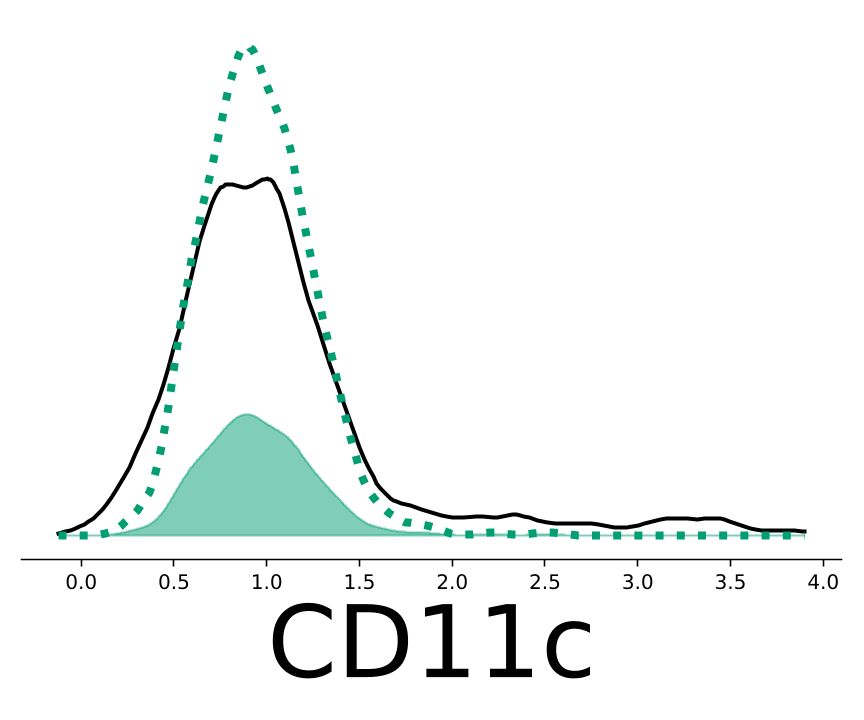} &\includegraphics[width=0.1\linewidth]{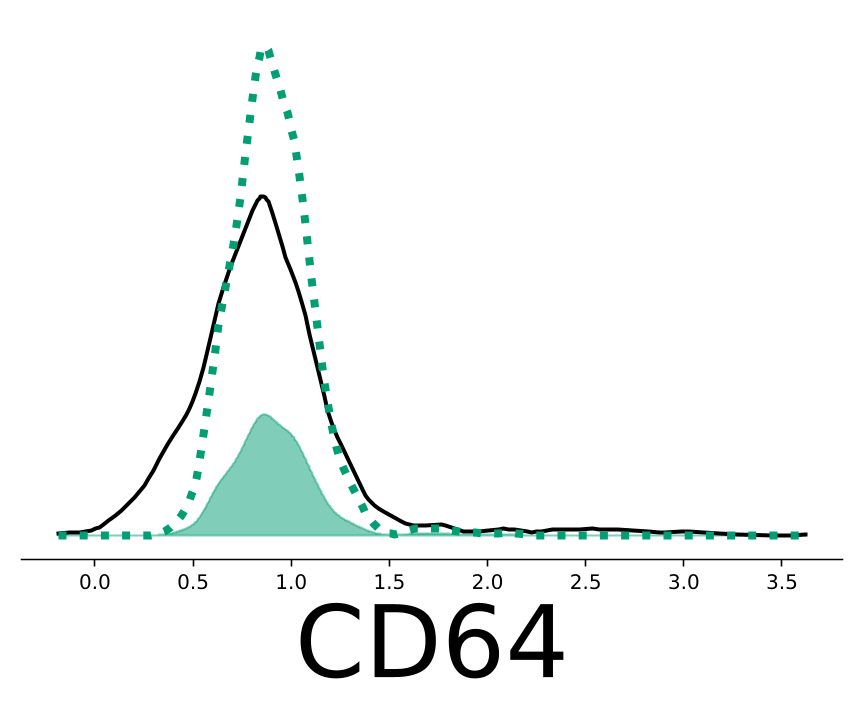} \\ 

         &\includegraphics[width=0.1\linewidth]{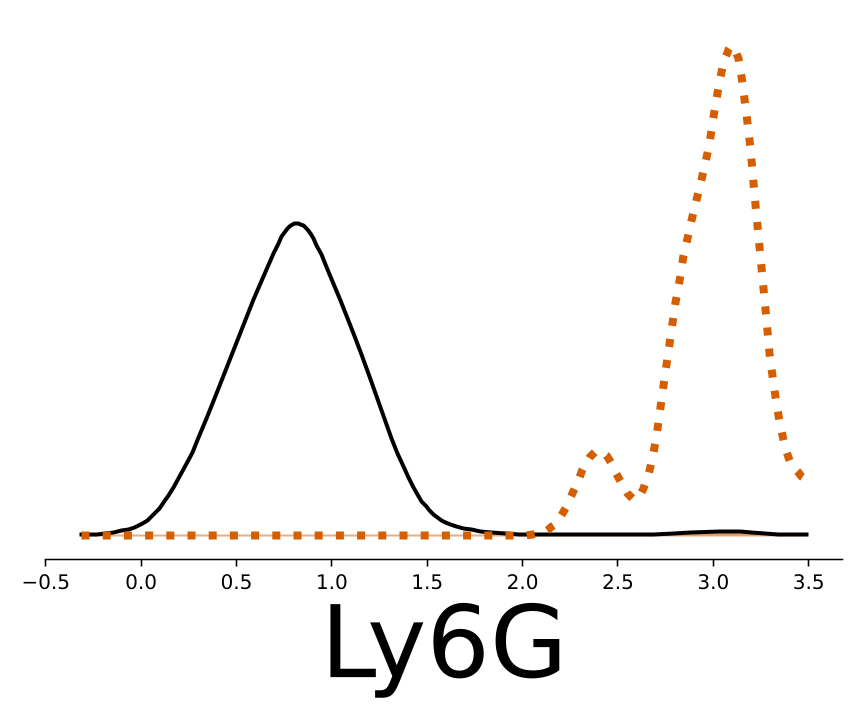}
         &\includegraphics[width=0.1\linewidth]{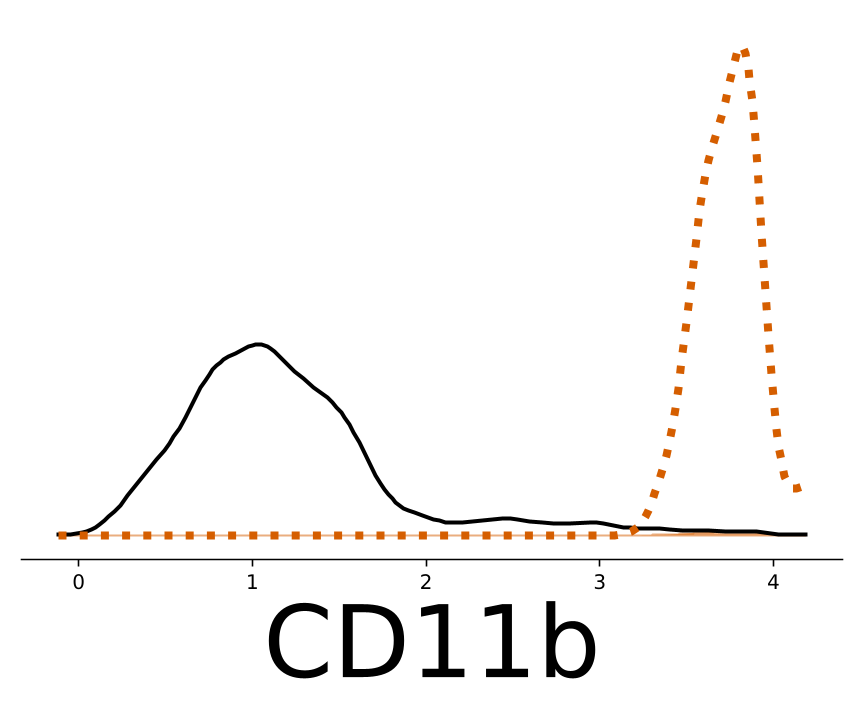}
         &&& \\
         
       \\
         
         &\includegraphics[width=0.1\linewidth]{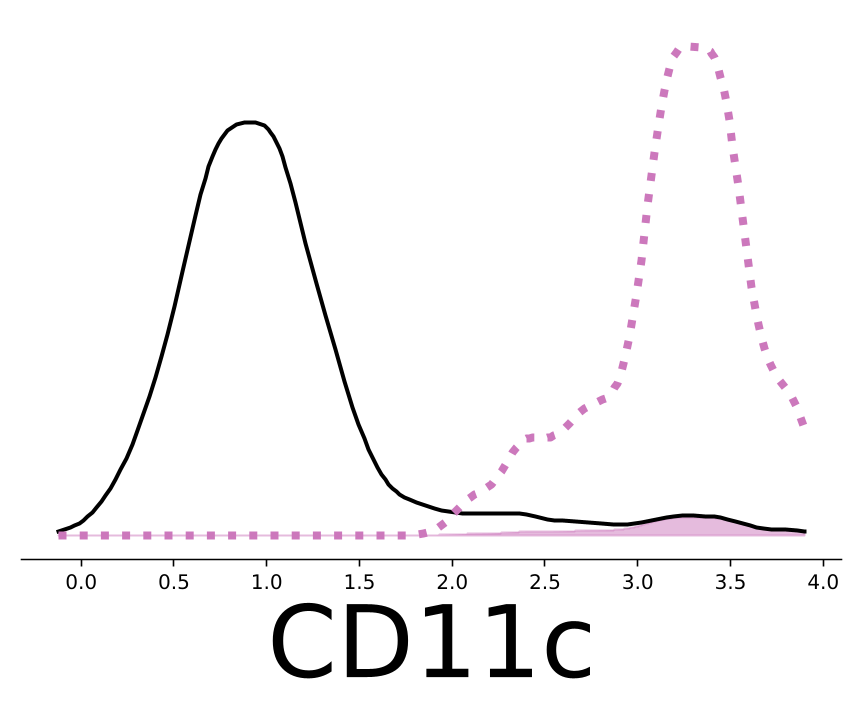} &\includegraphics[width=0.1\linewidth]{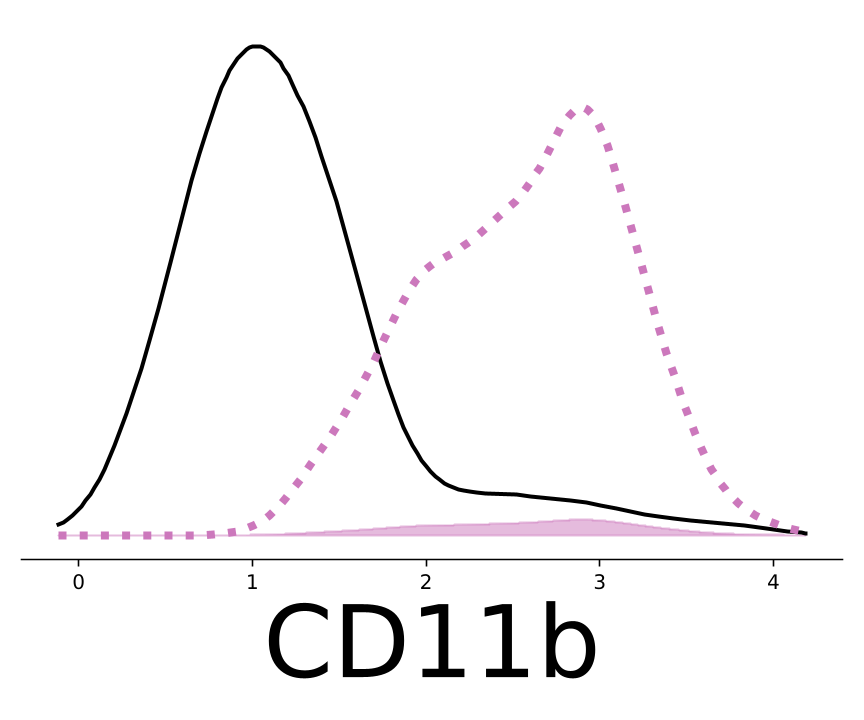} & &  \\
         
         (a2) Clustering by \InfoClus &\includegraphics[width=0.1\linewidth]{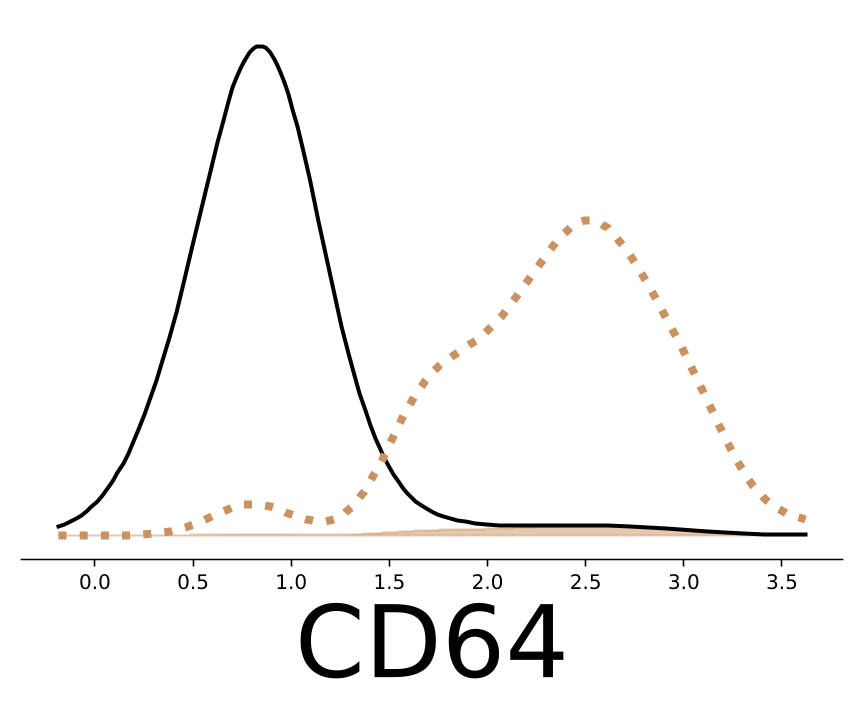} &\includegraphics[width=0.1\linewidth]{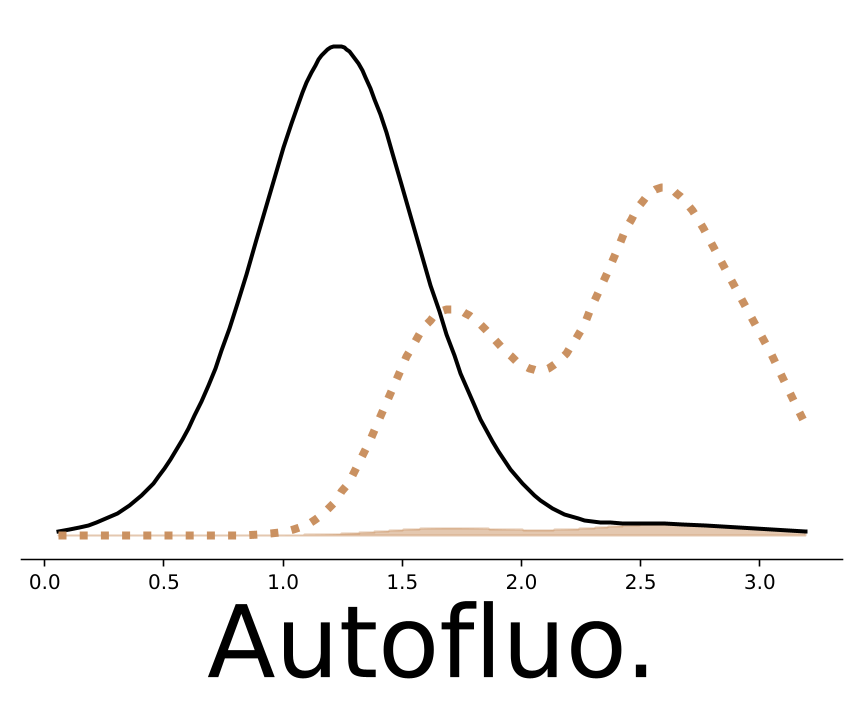} &\includegraphics[width=0.1\linewidth]{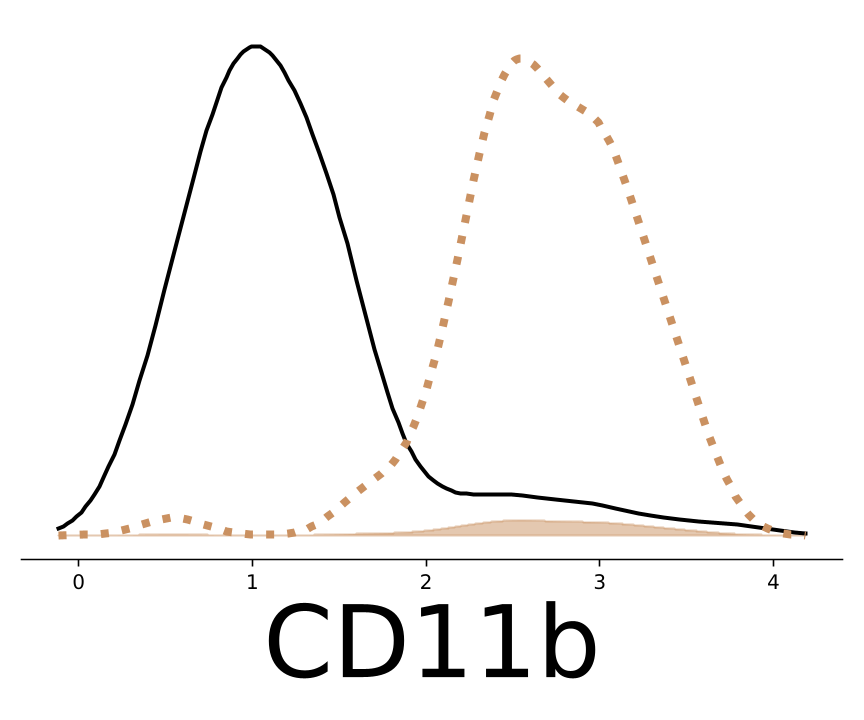} & & \\
         
           & \multicolumn{5}{c}{(b2) Explanation of (a1) given by \InfoClus} \\  
         
    \end{tabular}
    \caption{\Cyto\ dataset. Panels \underline{a1} and \underline{a2} show t-SNE embeddings colored by manual gating \cite{saeys2016} and by \InfoClus\ respectively. 
     Panel \underline{b1} shows how manual gating proceeds to label cells in a2. Panel \underline{b2} displays the explanations selected by \InfoClus\ for each cluster (the color coding maps explanations to their respective clusters). Each plot displays Kernel Density Estimates (KDE) of the distributions of the attribute in the cluster (dotted line) and on the full dataset (solid line). Filled color represents a KDE of the cluster scaled by how many points are covered by the cluster in the full data.
     }
    \label{fig:cytometry_infoclus}
    \vspace{-14pt}
\end{figure*}

Manual gating, routinely used in the analysis of cytometry data \cite{liu2025comprehensive,saeys2016}, proceeds by iteratively plotting cells on two-dimensional scatter plots with respect to a chosen subset of features, and selecting at each step a subset of cells on which to further focus in the next iteration \cite{saeys2016}. The sequence of features studied at each step is determined by expert knowledge. Fig. \ref{fig:cytometry_infoclus}a1, reproduced from \cite{saeys2016}, represents the results of this process in a t-SNE embedding of the \Cyto\ dataset. The manual gating process is illustrated by Fig. \ref{fig:cytometry_infoclus}b1, in which arrows indicate the sequence in which the analysis is conducted.

\ptitle{Qualitative analysis} We applied \InfoClus\ on \Cyto\ with hyper-parameters $\alpha = 1500$, $\beta = 1.5$, $minatt=2$ and $maxatt=5$. Given a 5 second limit for iterations ($t$ in Algorithm \ref{algo:infoclus}), \InfoClus\ conducted 9 iterations. Fig. \ref{fig:cytometry_infoclus}a2 and Fig. \ref{fig:cytometry_infoclus}b2 display the resulting 6 clusters, and their explanations (i.e., the distributions of features selected as explanations for each cluster).


\begin{figure*}[htp]
\centering
    \begin{tabular}{ccc}
          \raisebox{-0.15\height}{\includegraphics[width=0.3\linewidth]{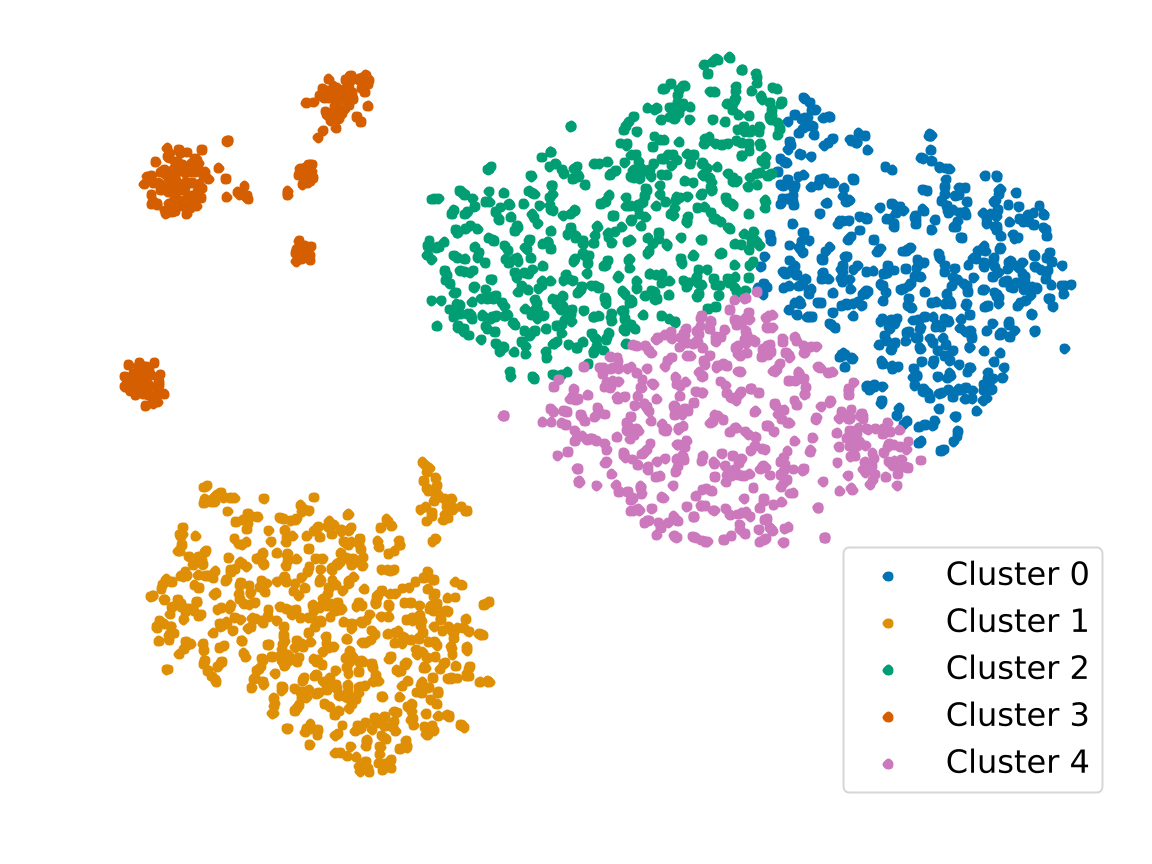}} & \raisebox{-0.15\height}{\includegraphics[width=0.3\linewidth]{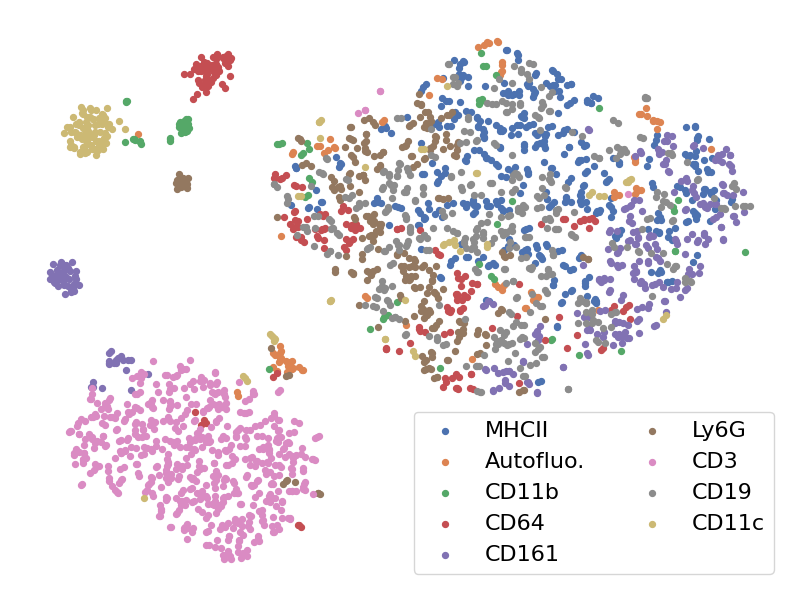}}&
          \raisebox{-0.15\height}{\includegraphics[width=0.35\linewidth]{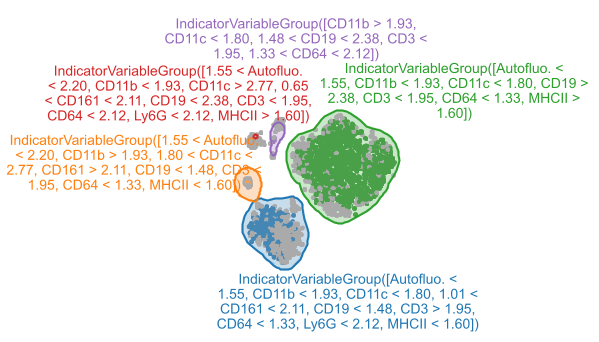}} \\
           (a) \InfoClus\ with $k$-means &(b) RVX &(c) VERA \\
          
    \end{tabular}
    \caption{
    Results of other methods aimed at explaining embeddings on \Cyto
    }
    \label{fig:cytometry_comparison}
    \vspace{-20pt}
\end{figure*}



Combining the partitioning and explanations provided by \InfoClus\
(Fig. \ref{fig:cytometry_infoclus}a2, b2), cluster 1 and 2 stand out---the former being characterized by high CD19 and low CD3, and the latter by high CD3 and low CD19.
These findings align with those of experts, which also use CD19 and CD3 to identify B and T cells---as displayed in the 2\textsuperscript{nd} row, 3\textsuperscript{rd} and 4\textsuperscript{th} sub-figures in Fig. \ref{fig:cytometry_infoclus}b1. Similar conclusions apply to cluster 3 (orange in a1; characterized by higher Ly6G and CD11b; labeled by experts as Neutrophils); cluster 4 (purple; higher CD11c; labeled as DCs), and cluster 5 (Ochre; higher CD64 and Autofluo; labeled as Macrophages). Altogether, the clusters and explanations generated by \InfoClus\ align with those of domain experts. \InfoClus\ may even help in uncovering possible inconsistencies in labels generated by manual gating: for instance, points in a cluster in the North-West of Fig. \ref{fig:cytometry_infoclus}a1 share similar values in high dimension, but are labeled as different cells---red (B cells) and purple (Neutrophils) labels.



\noindent \textbf{Comparison} 
We now compare \InfoClus\ to other methods aimed at explaining embeddings: \InfoClus\ with $k$-means, RVX \cite{thijssen2023}, and VERA \cite{policar2024}. \InfoClus\ with $k$-means proceeds by running $k$-means on a dataset for a range of $K$ different values ($k=3, 4, 5, \ldots, 32$ here), generating $K$ partitionings. Each partitioning is then associated to a set of explanations with best \pwx\ in terms of \xratio; the final partitioning selected is the one with highest \pwx\ among the $K$ candidates.
Descriptions of RVX and VERA have been presented in Section \ref{sec:related_work}.

Fig. \ref{fig:cytometry_comparison} shows the partitionings and explanations provided by \InfoClus\ with $k$-means, RVX, and VERA on the \Cyto\ dataset. We compare these to the \InfoClus\ result of Fig. \ref{fig:cytometry_infoclus}a1.
Explanation ratios for \InfoClus\ and \InfoClus\ with $k$-means are 5.61 and 5.39 respectively  (Fig. \ref{fig:cytometry_comparison}a). With $k$-means, clusters are roughly evenly sized and visually less in line with the structure of the embedding than those generated by \InfoClus: the North-East block of points in the embeddings is split in three (blue, green and pink clusters), while the disconnected point sets in the North-West corner are considered a single cluster (red).

In Fig. \ref{fig:cytometry_comparison}b we find that RVX also shows clear cluster structure, even though it is not a clustering method. However, the largest cluster lacks any clear visual structure. Note also that by definition of the method, each point is associated to exactly one attribute and not more. This is probably why the B cells that comprise the North-East cluster do not form a visually coherent cluster, they are not defined through any single attribute.

The clusters generated by VERA (Fig. \ref{fig:cytometry_comparison}c) are visually coherent, but the annotations involve almost all attributes in the data. The threshold-based explanations given by VERA also cause the coverage of the clusters to be low, i.e., the explanations cover only part of the points in the identified regions.

\begin{figure*}[ht]
\vspace{-16pt}
\centering
    \begin{tabular}{c|c@{}c@{}cc}
        \multirow{3}{*}{\includegraphics[width=0.35\linewidth, trim=0mm 10mm 0mm 22mm, clip]{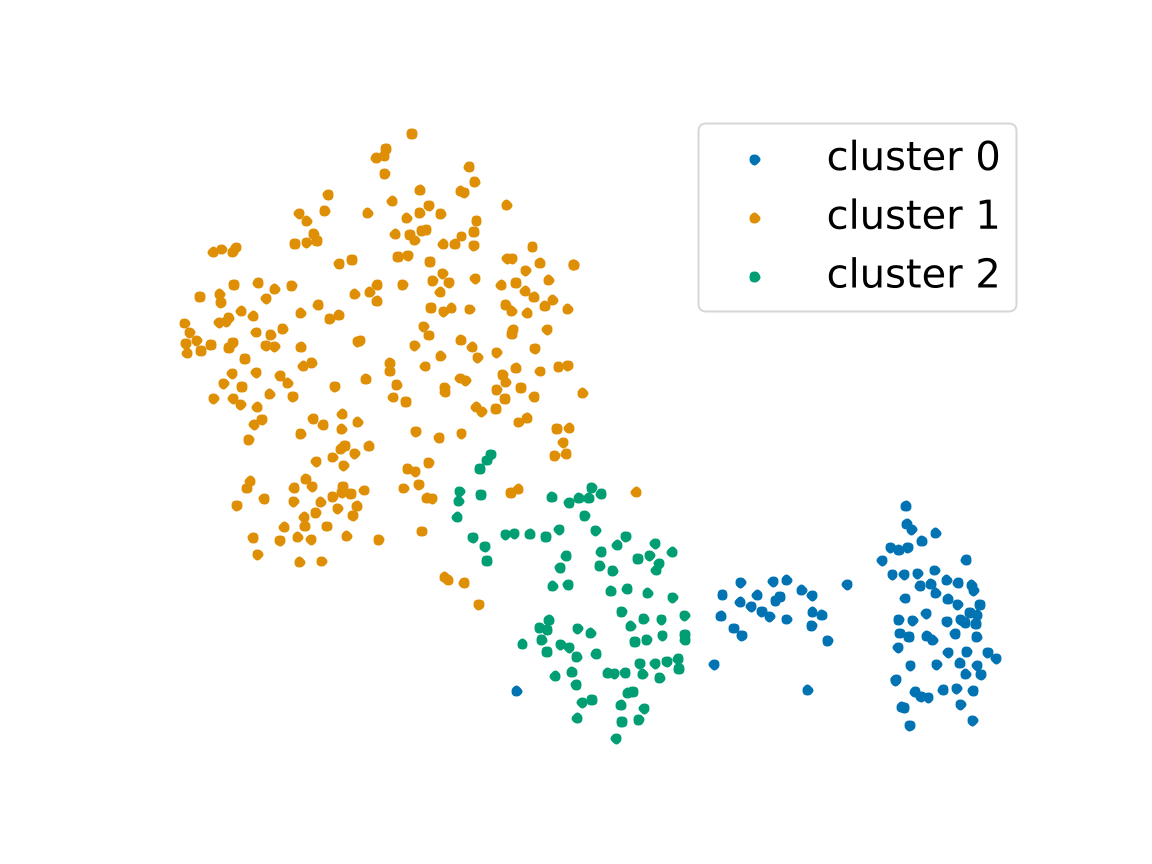}} & \includegraphics[width=0.1\linewidth]{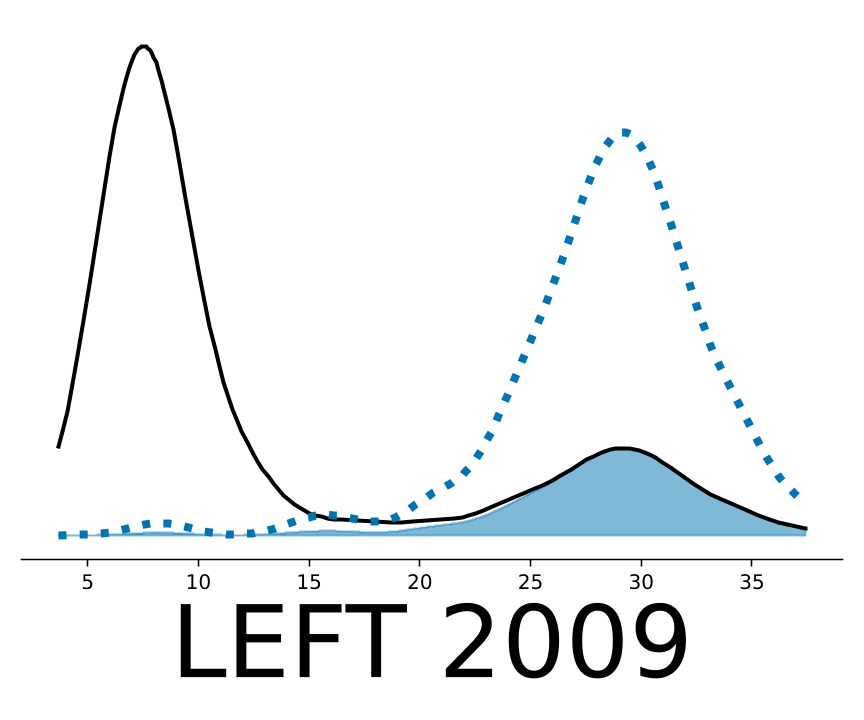} & \includegraphics[width=0.1\linewidth]{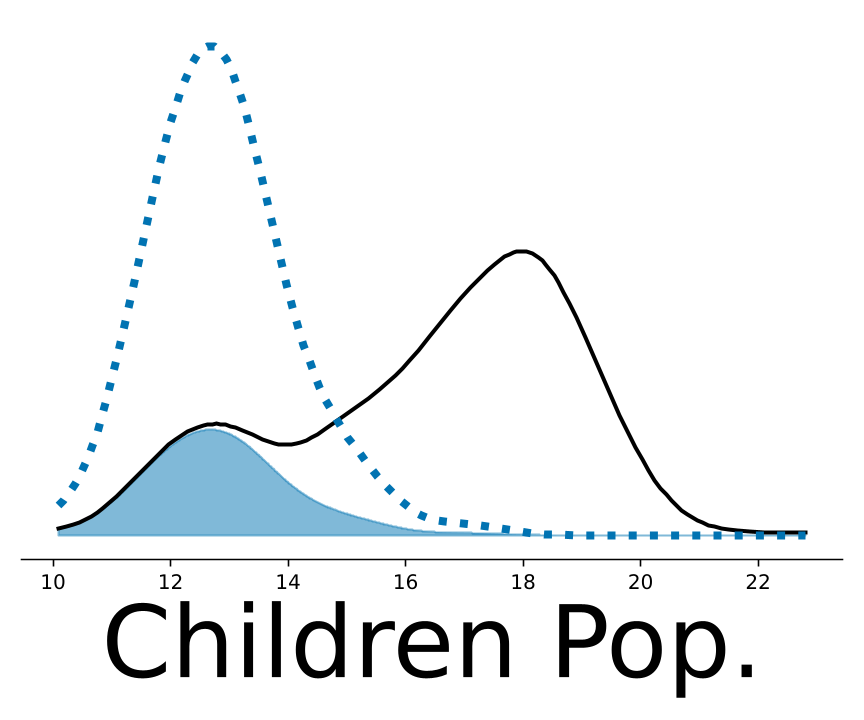} & &  \multirow{3}{*}{\includegraphics[width=0.25\linewidth, trim=0mm 0mm 0mm 5mm, clip]{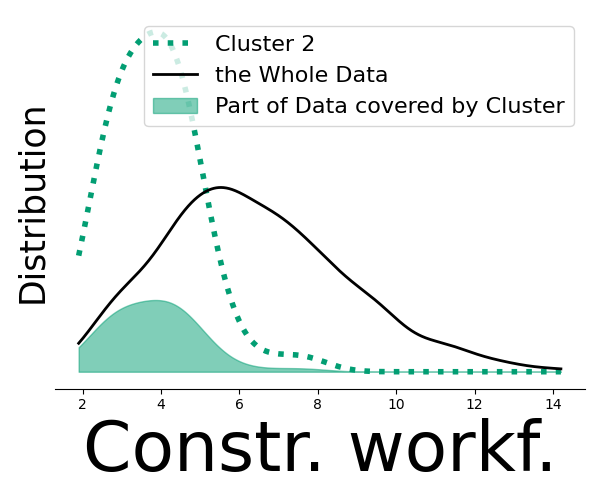}}\\
        &\includegraphics[width=0.1\linewidth]{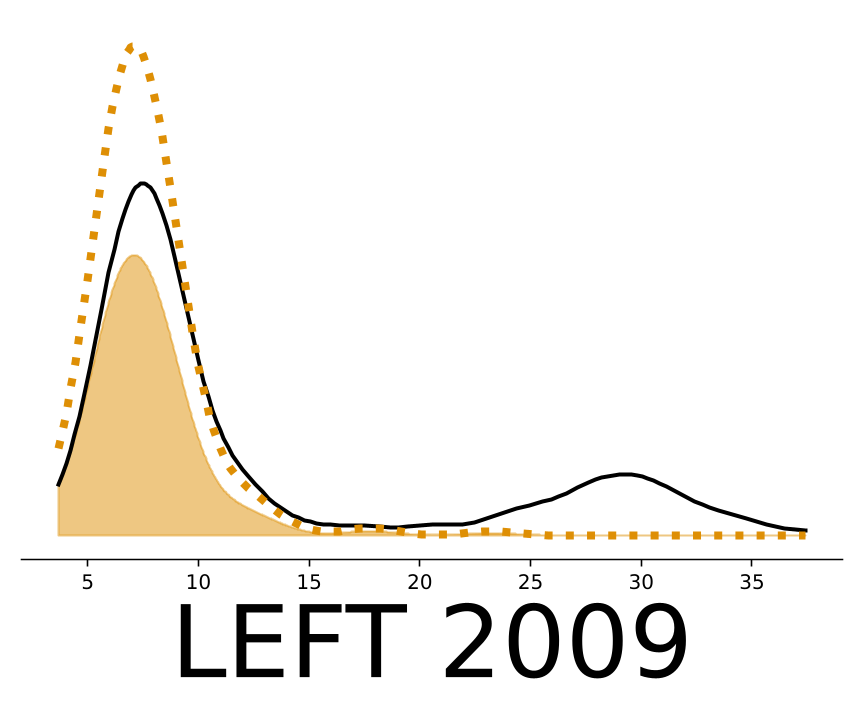} &\includegraphics[width=0.1\linewidth]{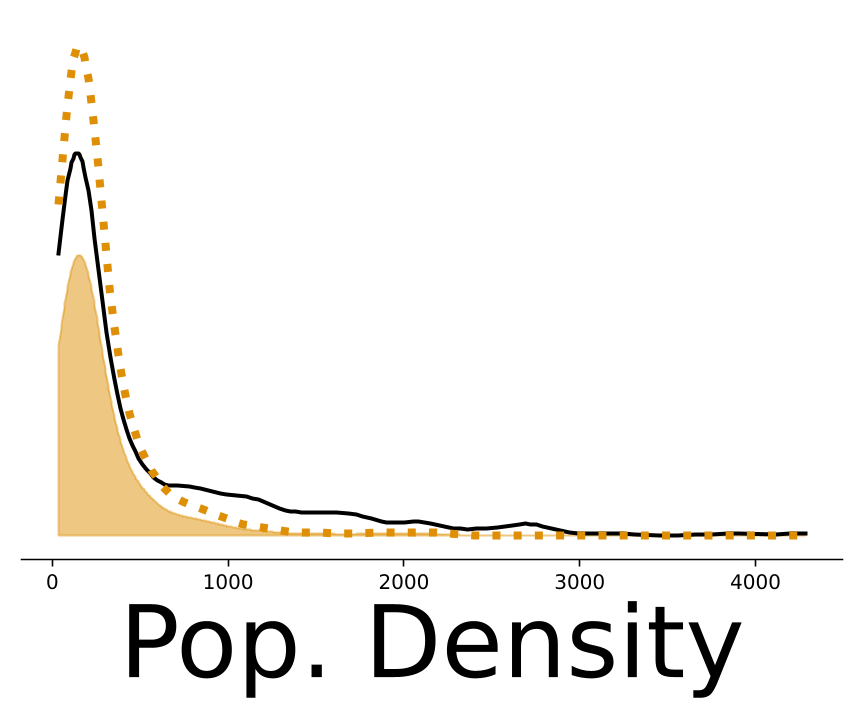} &\includegraphics[width=0.1\linewidth]{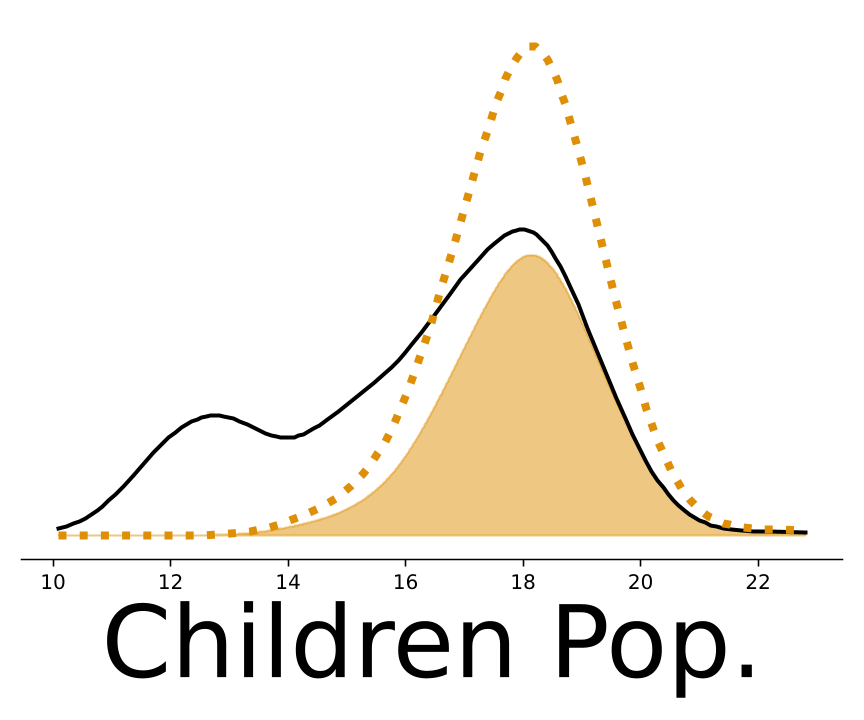} & \\
        &\includegraphics[width=0.1\linewidth]{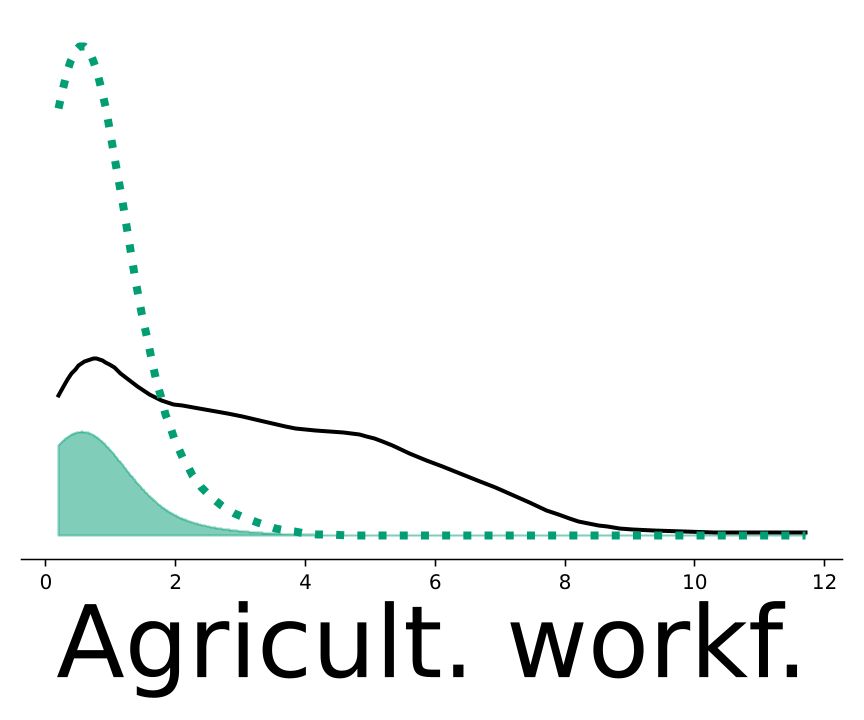} &\includegraphics[width=0.1\linewidth]{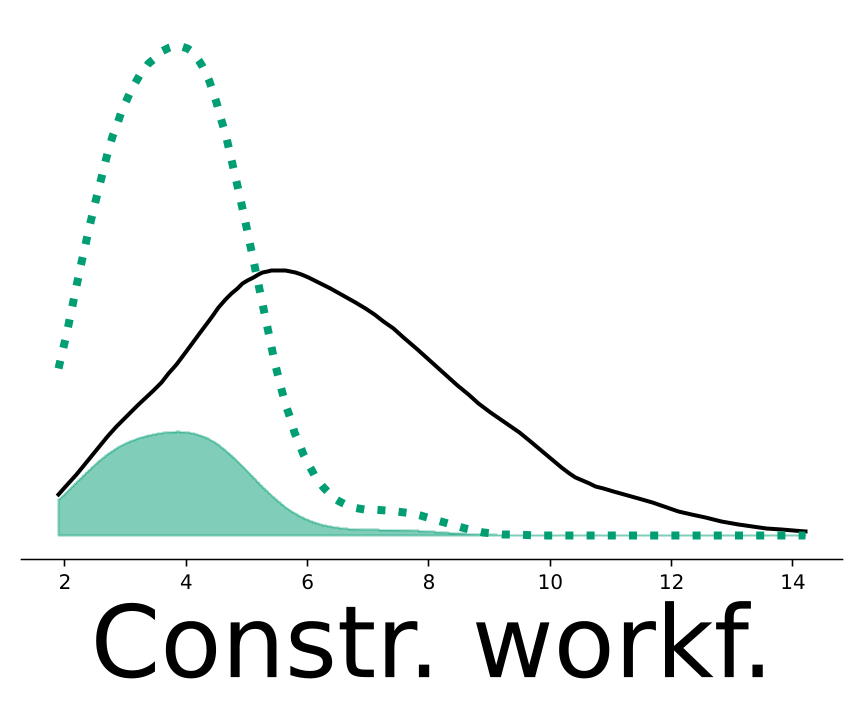} & & \\
        (a) \InfoClus\ partitioning &  \multicolumn{4}{c}{(b) Explanation for the left partitioning} \\
    \end{tabular}
    \begin{tikzpicture}[overlay, remember picture]
    \draw[thick,red, ->] (-4.9,-0.9) .. controls +(0,0) and +(0,0) .. (-3,-0.5);
    \end{tikzpicture}
    \caption{\InfoClus\ analysis in GSE}
    \label{fig:german_infoclus}
    \vspace{-16pt}
\end{figure*}

\ptitle{German Socio-Economics (\German) data \cite{Boley2013GermanDataset}}
This dataset describes 412 administrative districts of Germany in terms of 31 socio-demographic attributes, such as distributions of population age and labor sector, and voting percentages, recorded at the 2009 elections.
We set hyper-parameters on this dataset as follows: $\alpha = 50$, $\beta = 1.5$, $minatt=2$ and $maxatt=5$. Given a 5 second limit for iterations, \InfoClus\ conducted 18 iterations and returned a partitioning with 3 clusters, presented alongside their explanations in Fig. \ref{fig:german_infoclus}.

\begin{figure*}[ht]
\vspace{-22pt}
\centering
    \begin{tabular}{ccc}
        \includegraphics[width=0.25\linewidth,  trim=30mm 0mm 30mm 0mm, clip]{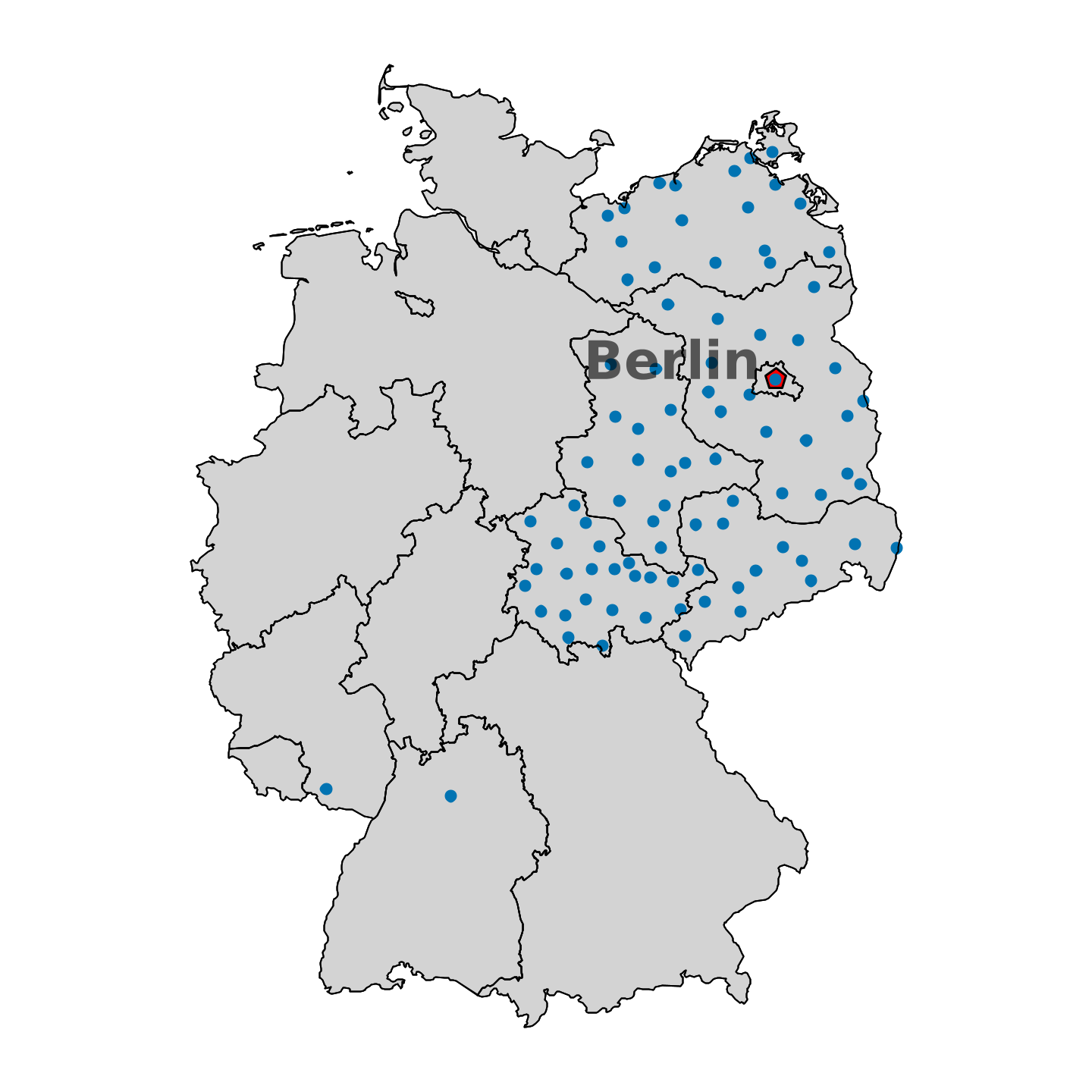} & \includegraphics[width=0.25\linewidth,  trim=30mm 0mm 30mm 0mm, clip]{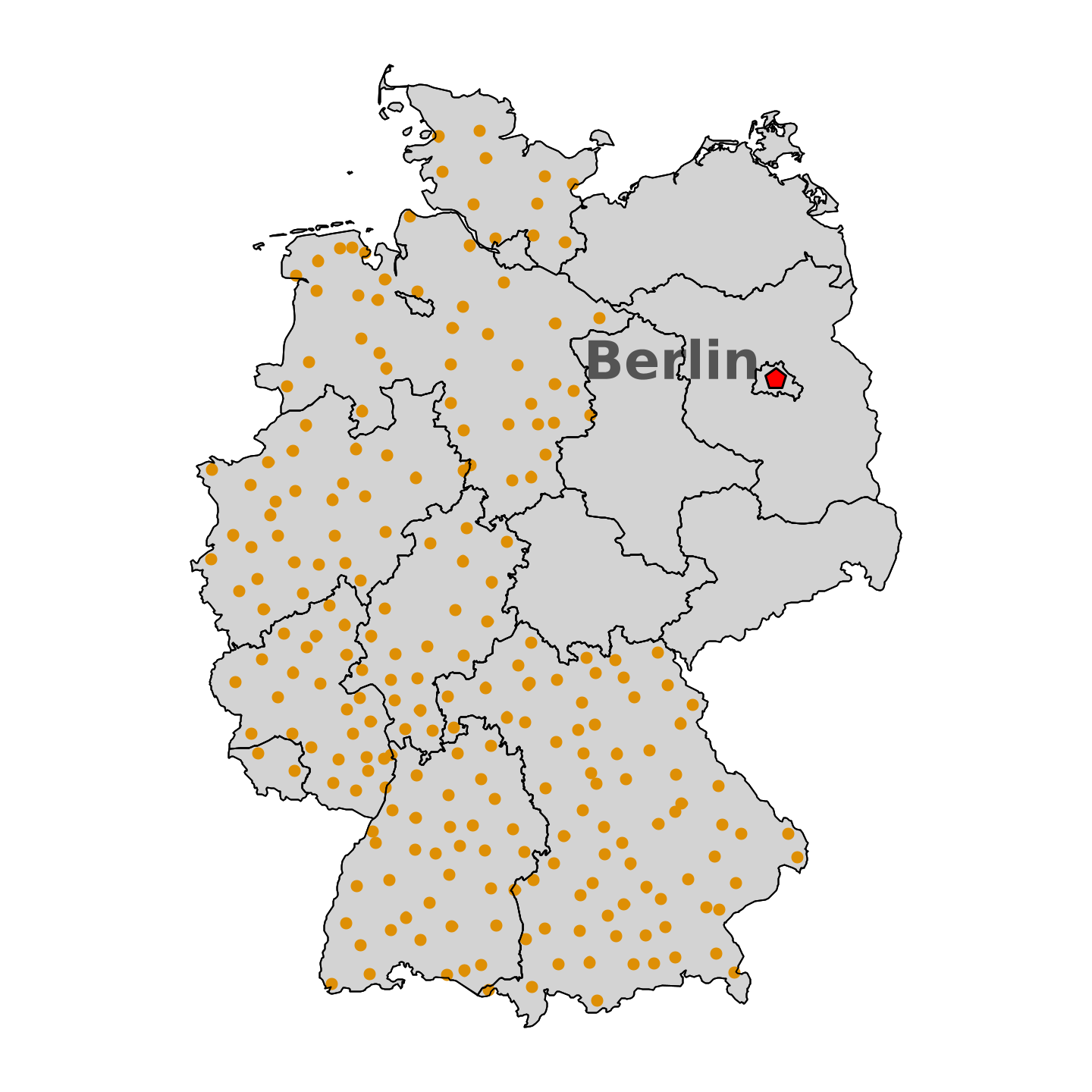} &\includegraphics[width=0.25\linewidth, trim=30mm 0mm 30mm 0mm, clip]{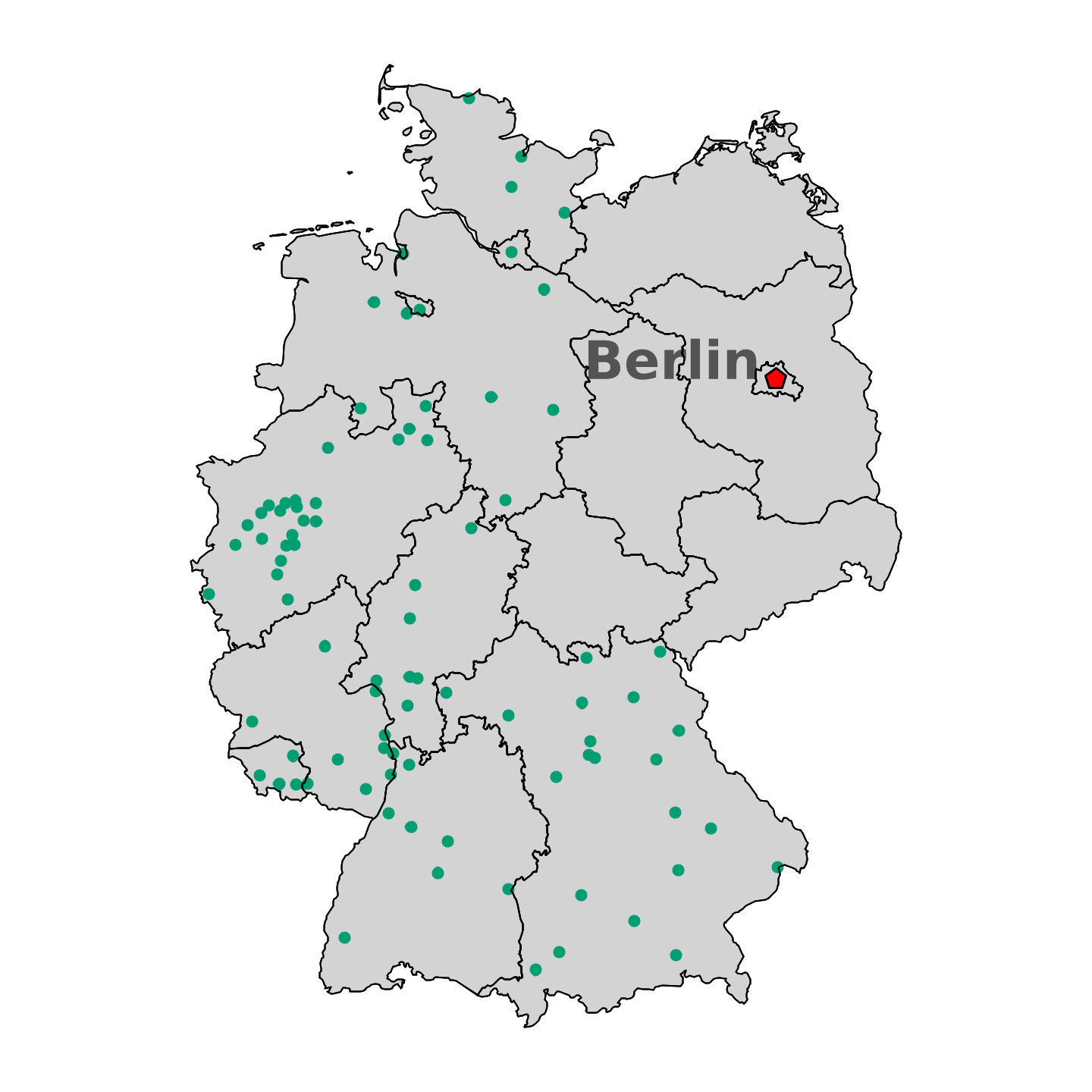} \\
        (a) Cluster 0 &(b) Cluster 1 &(c) Cluster 2 \\
    \end{tabular}
    \begin{tikzpicture}[overlay, remember picture]
    \draw[red, thick] (-2.65, 0.34) circle (0.28); 
    \end{tikzpicture}
    \caption{
    Visualization of \InfoClus\ clusters on the map of Germany.
    }
    \label{fig:german_map}
    \vspace{-11pt}
\end{figure*}
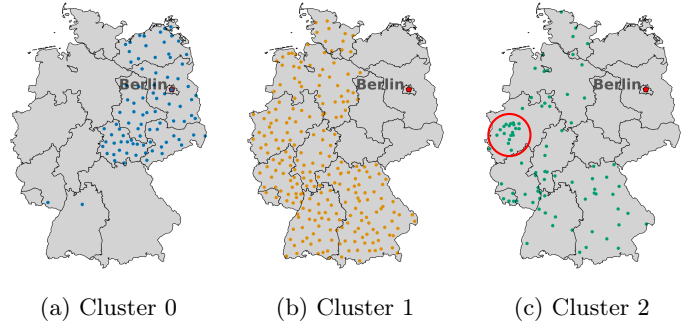

In Fig. \ref{fig:german_infoclus}, the voting ratio of the LEFT party in 2009 is selected as the most interesting attribute in both cluster 0 and 1, with high and low support respectively. 
 Cluster 2 stands out for the low shares of its workforce working in agriculture and construction, hinting that cluster 2 might correspond to urban areas. We verify this insight using metadata of \German: 86\% of districts in cluster 2 are urban, compared of 28\% of all districts in the data.
Out of interest, we visualized the partitioning of districts on the German map (Fig. \ref{fig:german_map}). Indeed, Cluster 0 and 1 are clearly geographically separate, and correspond to former East and West Germany, and cluster 2 corresponds in a large part to the Ruhr area, shown in the red circle in Fig. \ref{fig:german_map}c. Note that this geographic coherence is striking as t-SNE and \InfoClus\ do not know about the geographic location of the districts: these clusters emerged purely from demographic and voting statistics.

\begin{figure*}[ht]
\vspace{-16pt}
\centering
    \begin{tabular}{c|c@{}c@{}c@{}c@{}c}
         \multirow{3}{*}{\includegraphics[width=0.38\linewidth]{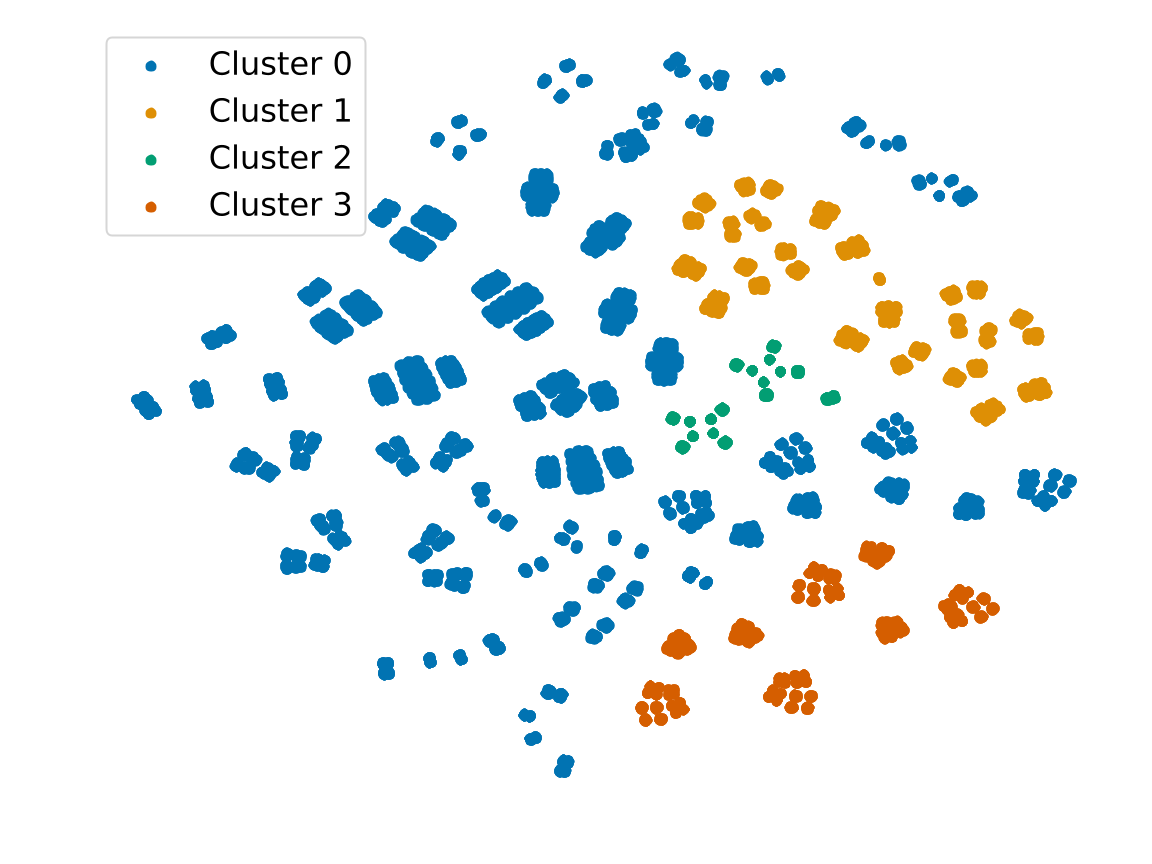}} &\includegraphics[width=0.12\linewidth]{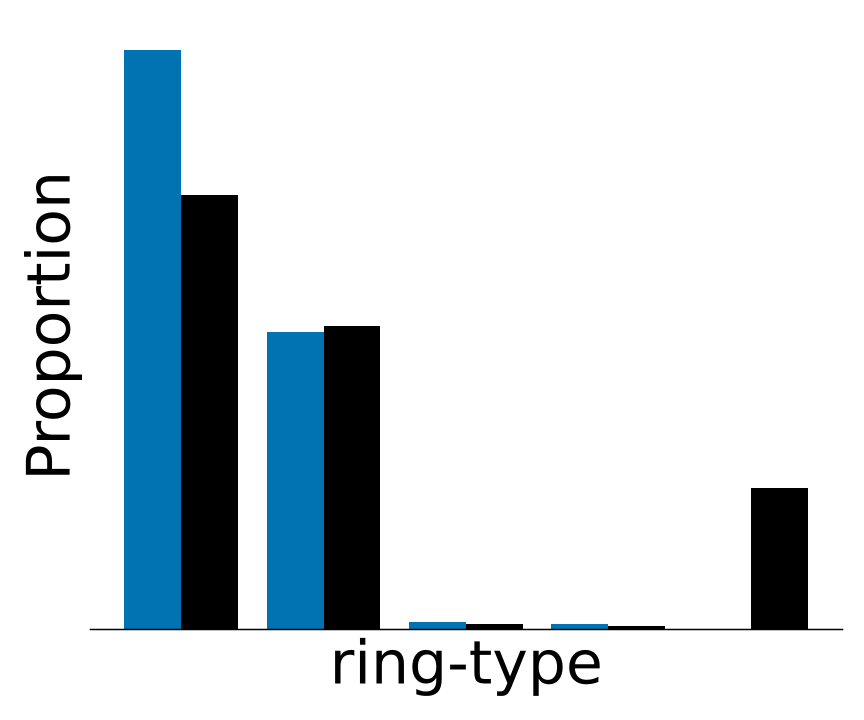} &\includegraphics[width=0.12\linewidth]{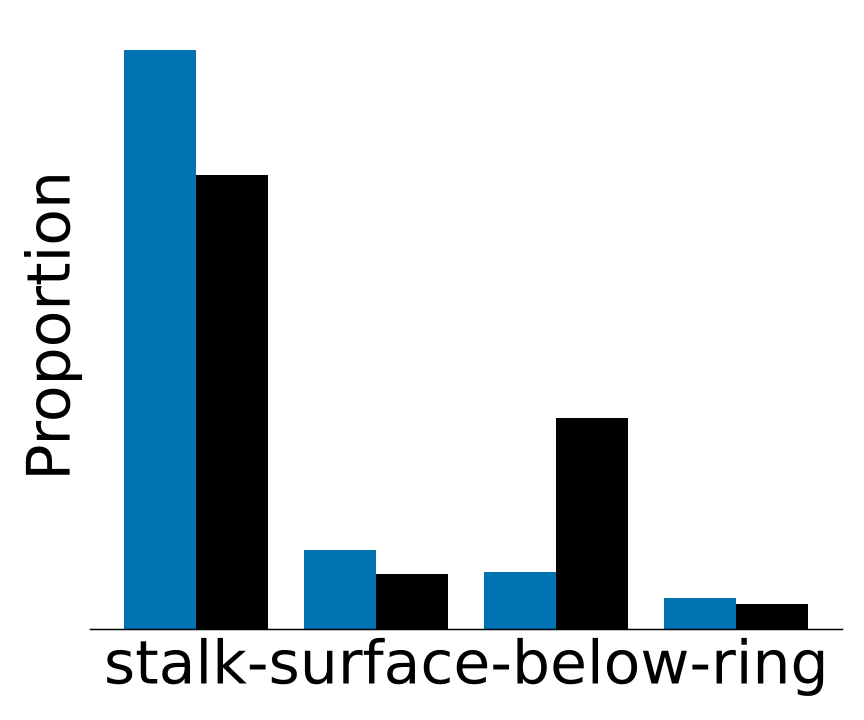} &&& \\
         &\includegraphics[width=0.12\linewidth]{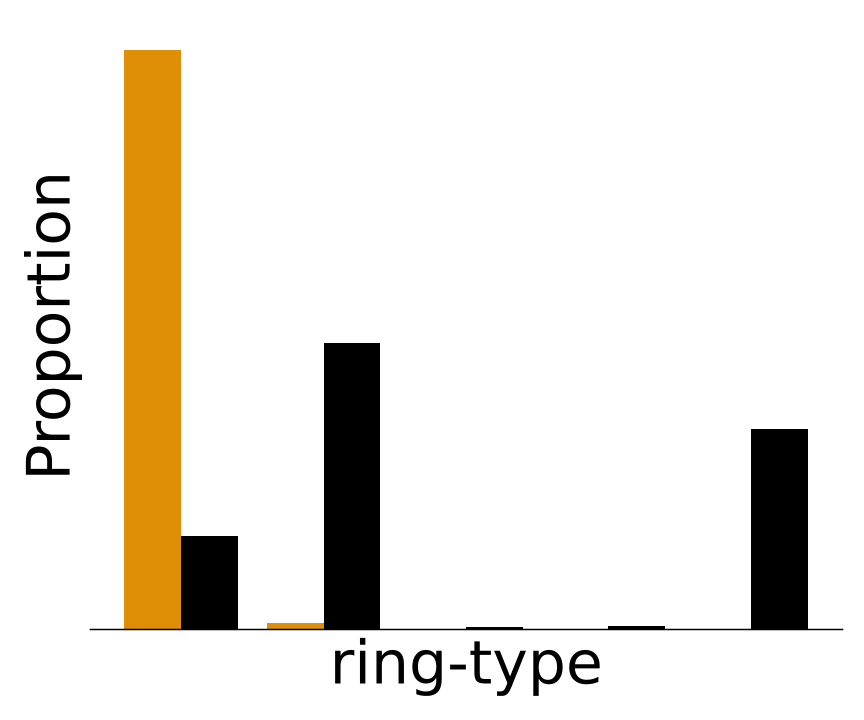} &\includegraphics[width=0.12\linewidth]{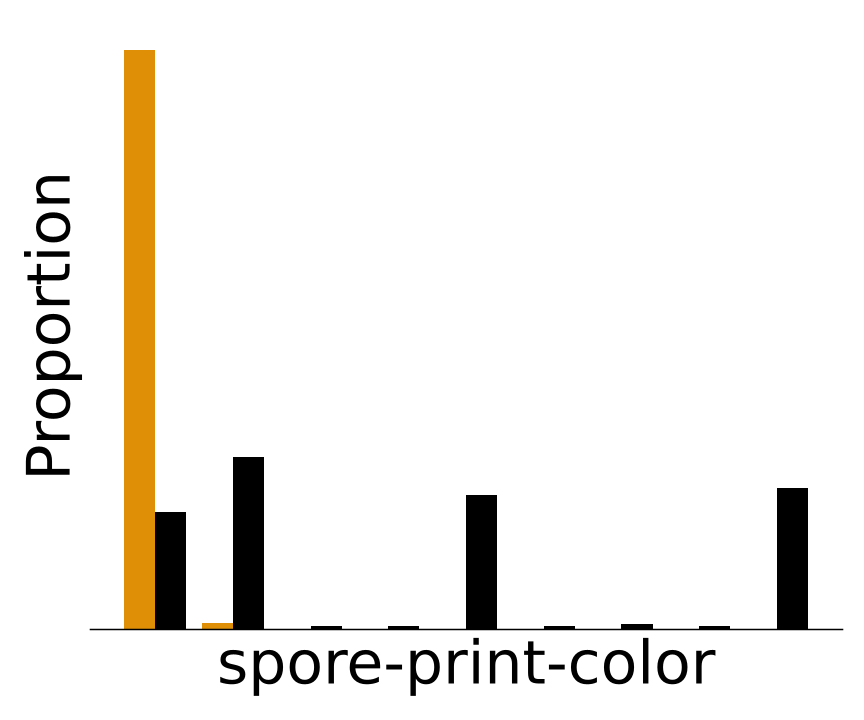} &\includegraphics[width=0.12\linewidth]{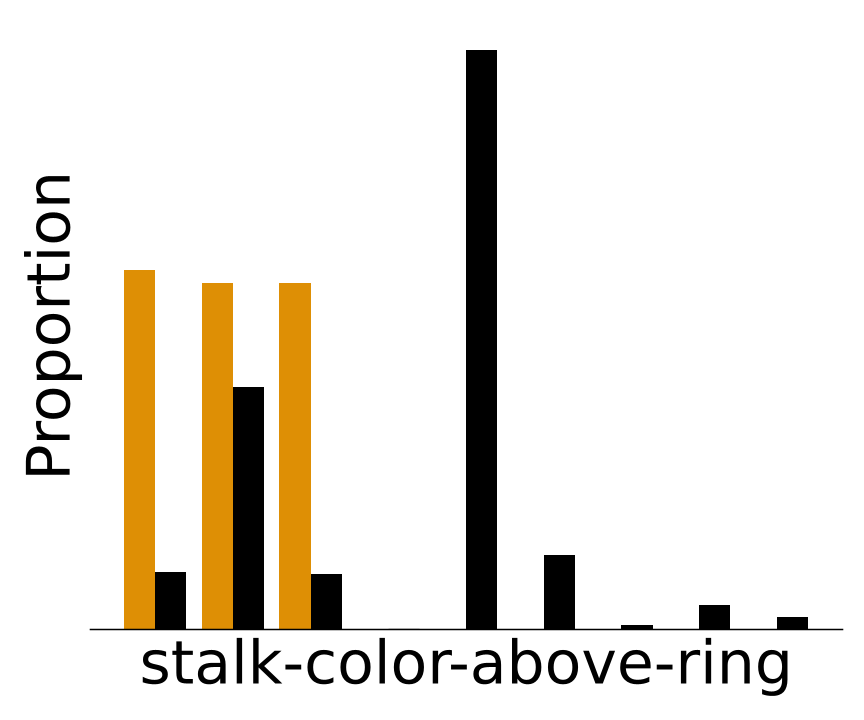} &\includegraphics[width=0.12\linewidth]{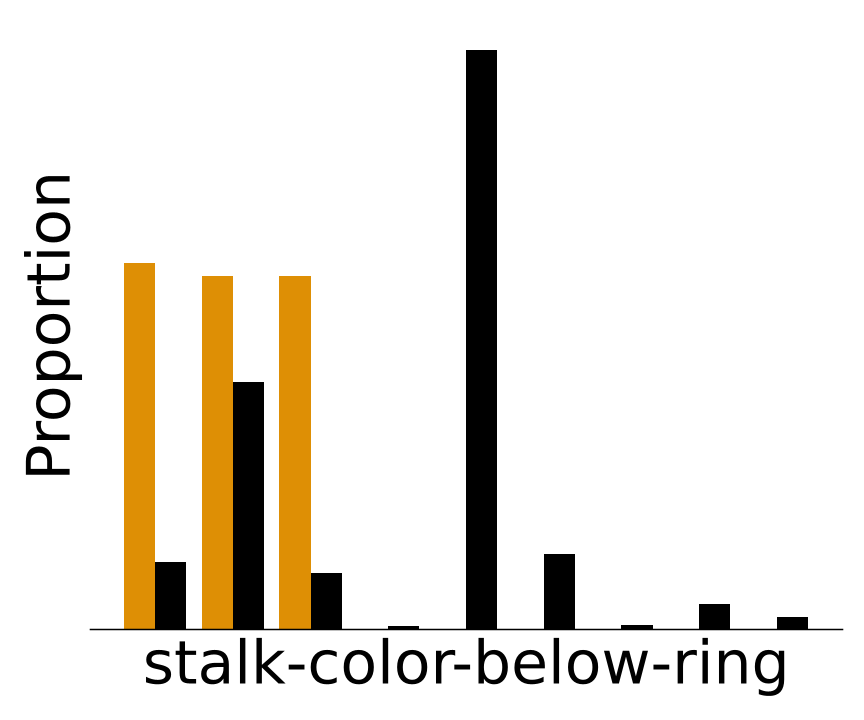} &\includegraphics[width=0.12\linewidth]{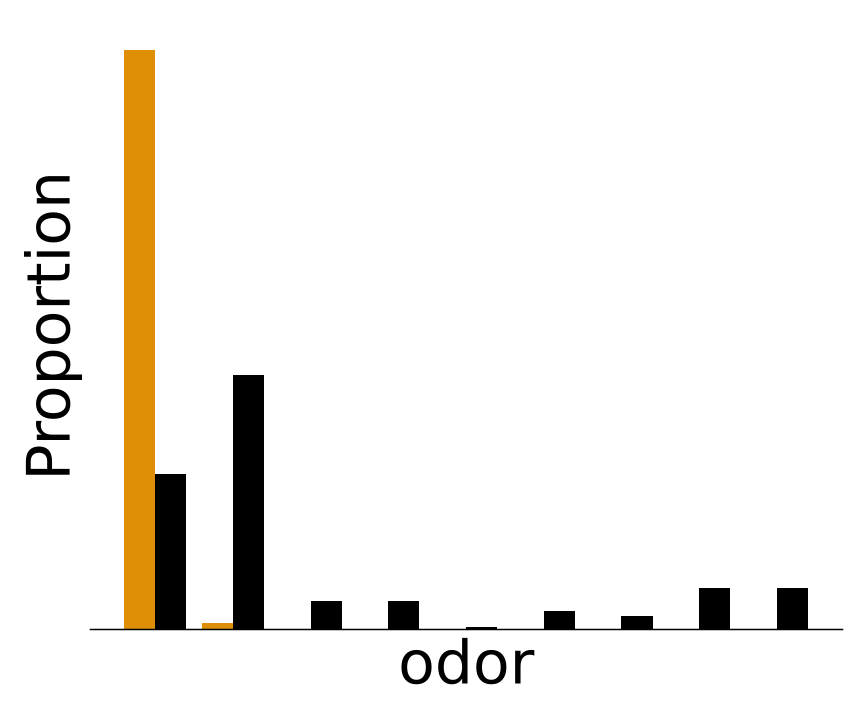} \\
         &\includegraphics[width=0.12\linewidth]{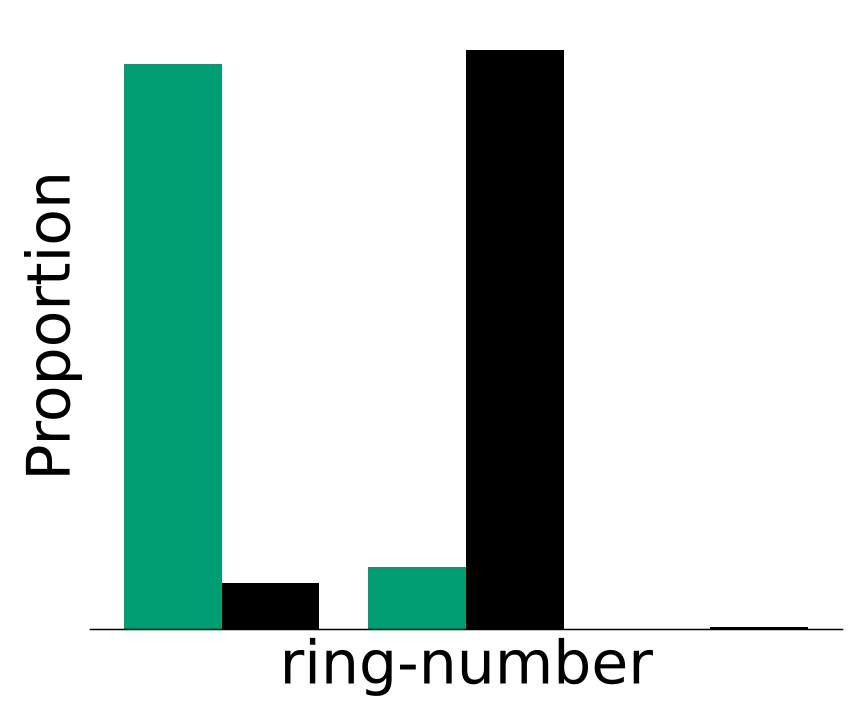} &\includegraphics[width=0.12\linewidth]{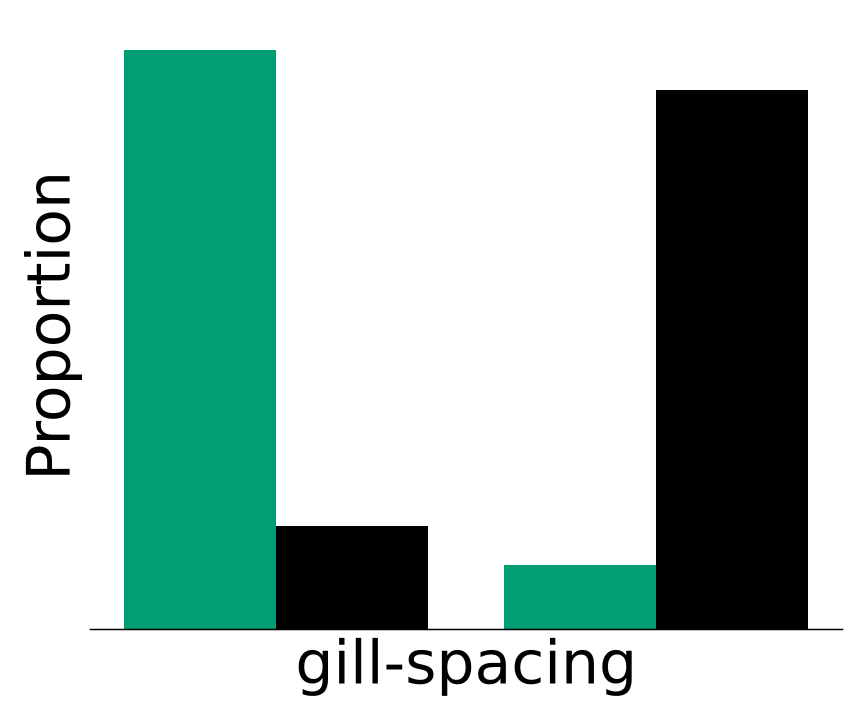} & && \\
         &\includegraphics[width=0.12\linewidth]{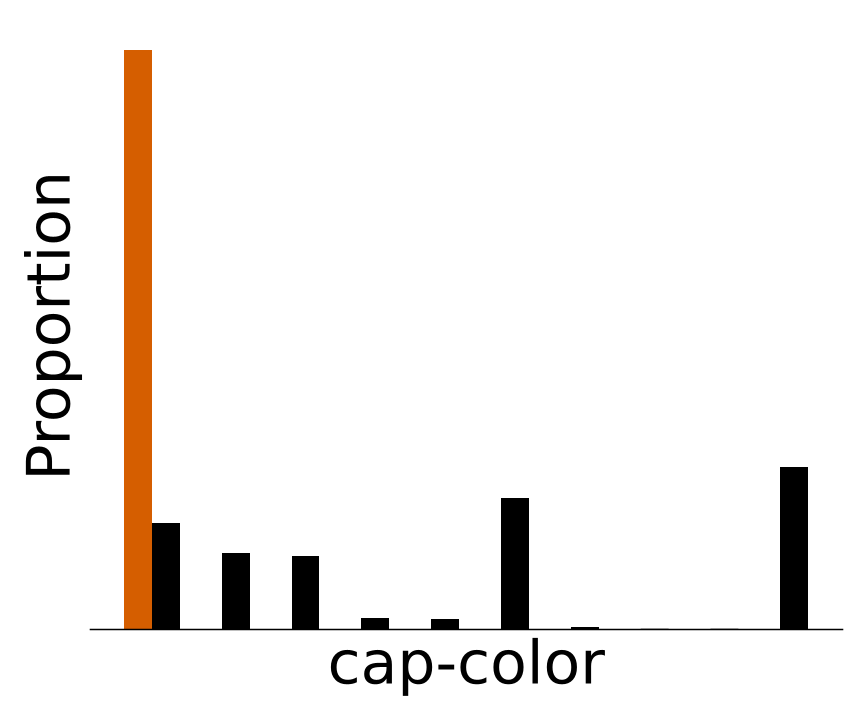} &\includegraphics[width=0.12\linewidth]{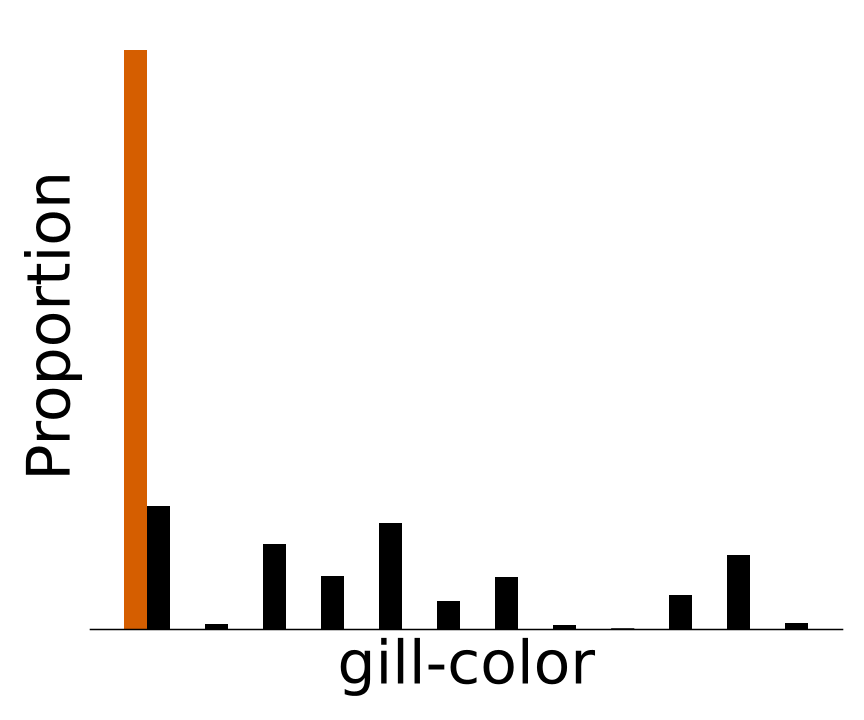} & && \tikzmark{cellrb} \\
         (a) \InfoClus\ partitioning & \multicolumn{5}{c}{(b) Explanation for the left partitioning}
    \end{tabular}
    \begin{tikzpicture}[overlay, remember picture]
    \node[anchor=south west] at ([xshift=2.5cm, yshift=0.5cm] pic cs:cellrb.south east)
    {\includegraphics[width=0.25\linewidth]{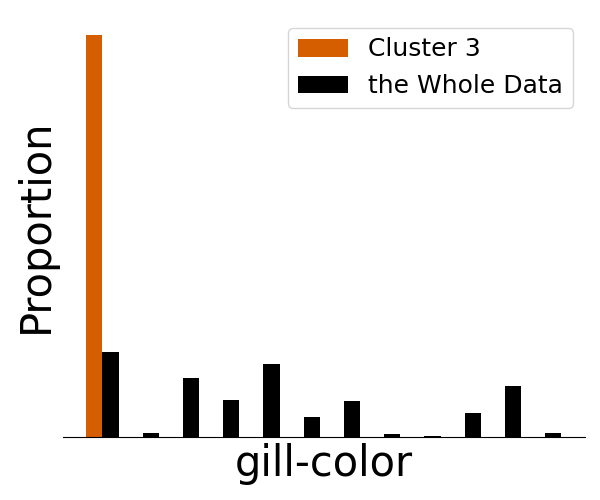}};
    \draw[thick,red, ->] (0.9,1.1) .. controls +(0,0) and +(0,0) .. (2.5,2);
    \end{tikzpicture}
    \caption{\InfoClus\ analysis on Mushroom dataset}
    \label{fig:mushroom_infoclus}
    \vspace{-18pt}
\end{figure*}

\ptitle{UCI Mushroom data \cite{mushroom_73}}
\Mushroom\ consists of 8124 instances characterized by 22 categorical 
attributes that describe 23 species of gilled mushrooms within the Agaricus and Lepiota family.
We considered this data to investigate \InfoClus' ability to interpret embeddings of categorical data.
We set hyper-parameters on this dataset as follows: $\alpha = 800$, $\beta = 1.5$, $minatt=2$ and $maxatt=5$. Given a limit of 30 seconds for iterations, \InfoClus\ conducted 10 iterations and returned a partitioning with 4 clusters, shown in Fig. \ref{fig:mushroom_infoclus}.

Four visually coherent clusters emerge, differing from the dataset as a whole with respect to several attributes: for instance, large rings and orange print colors for cluster 1, low ring number and close gill spacing for cluster 2, purple caps and red gill colors for cluster 3. Unfortunately, a lack of background on mushrooms prevents us from further interpreting these findings.

\subsection{Hyper-parameter analysis  \label{sec:hyperparameters}}
We now investigate the sensitivity of \InfoClus' results to hyperparameter selection, using the \Cyto\ dataset. 
Since $t$, $minatt$ and $maxatt$ (respectively: imposed iteration time, minimum and maximum features per cluster) should be fixed based on users' preference, we focus on the role of hyper-parameters $\alpha$ and $\beta$. In the following experiment, we set $minatt=2$, $maxatt=5$, and $t=5$.

\begin{figure*}[ht]
\centering
    \begin{tabular}{|c|cccccc|}
    \hline
        \diagbox{$\beta$}{$\alpha$} &250 &500 &1000 &1500 &2000 &2500 \\ \hline
        1.4 & \includegraphics[width=0.15\linewidth]{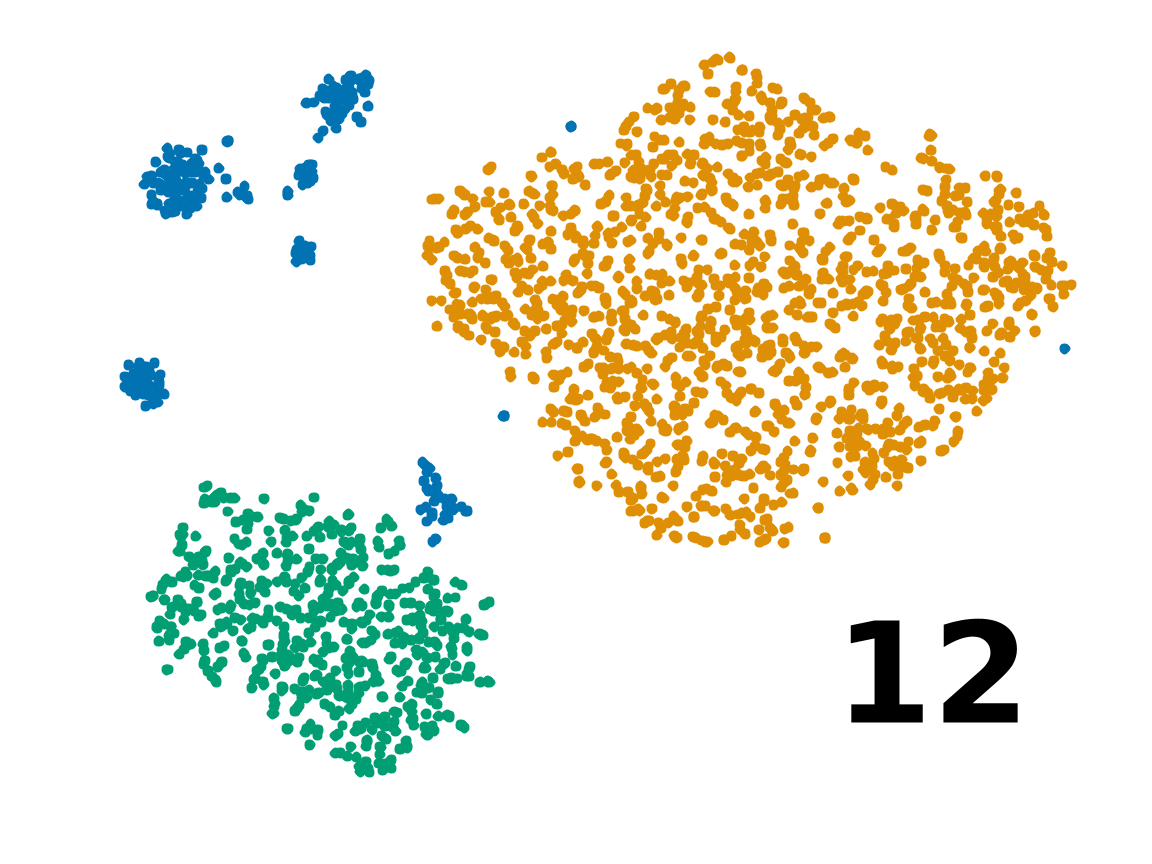}&\includegraphics[width=0.15\linewidth]{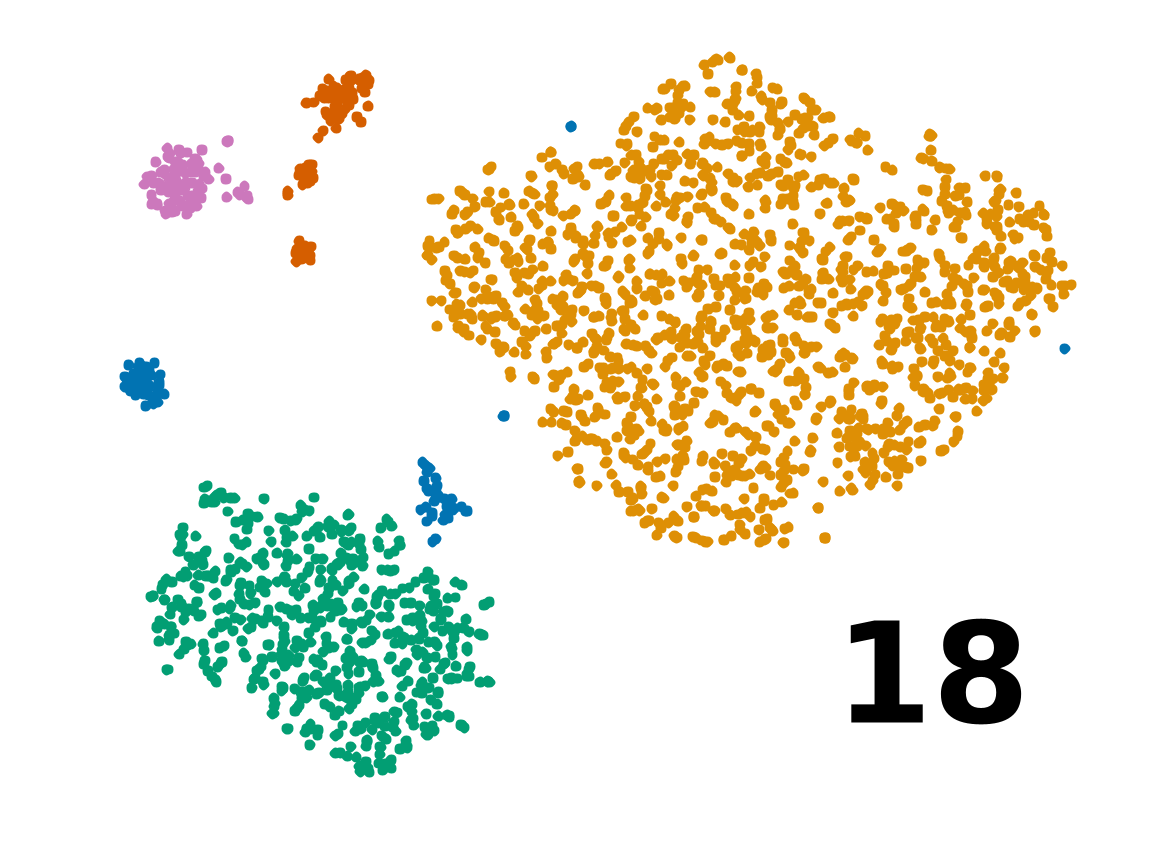} &\includegraphics[width=0.15\linewidth]{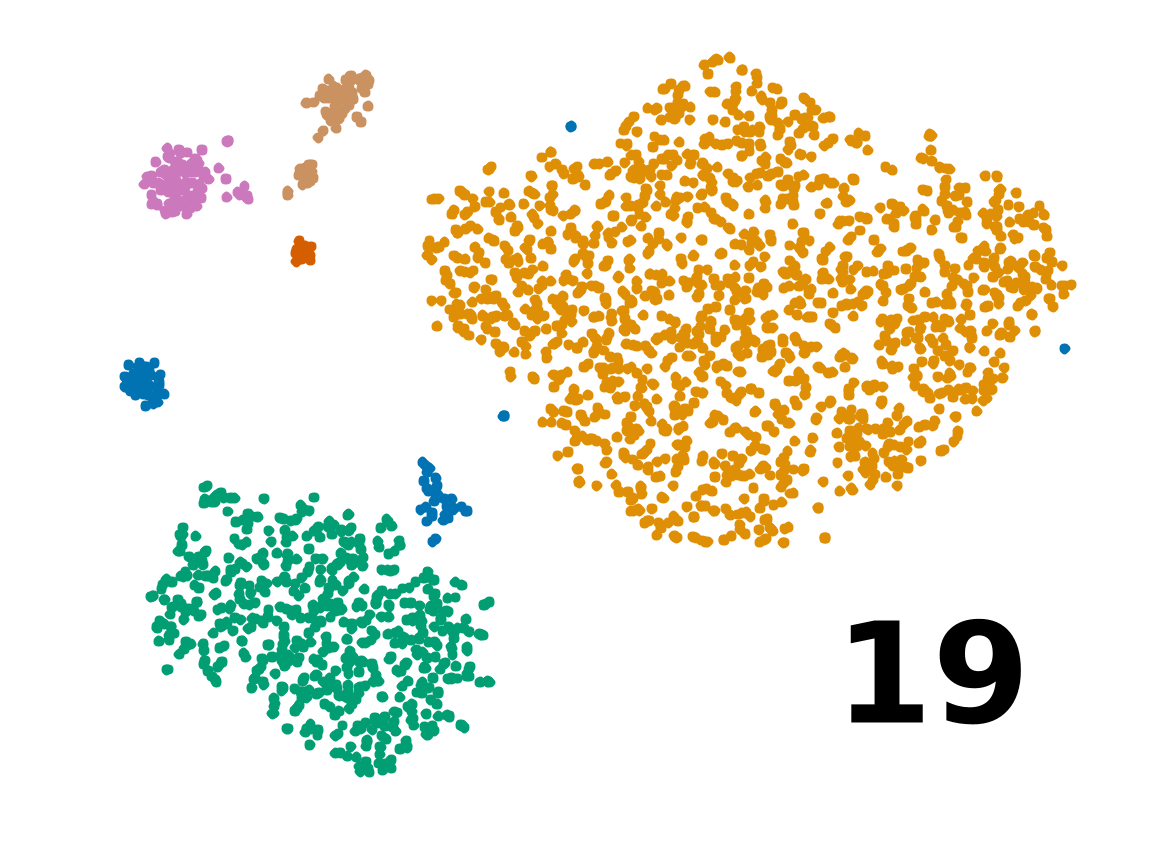} &\includegraphics[width=0.15\linewidth]{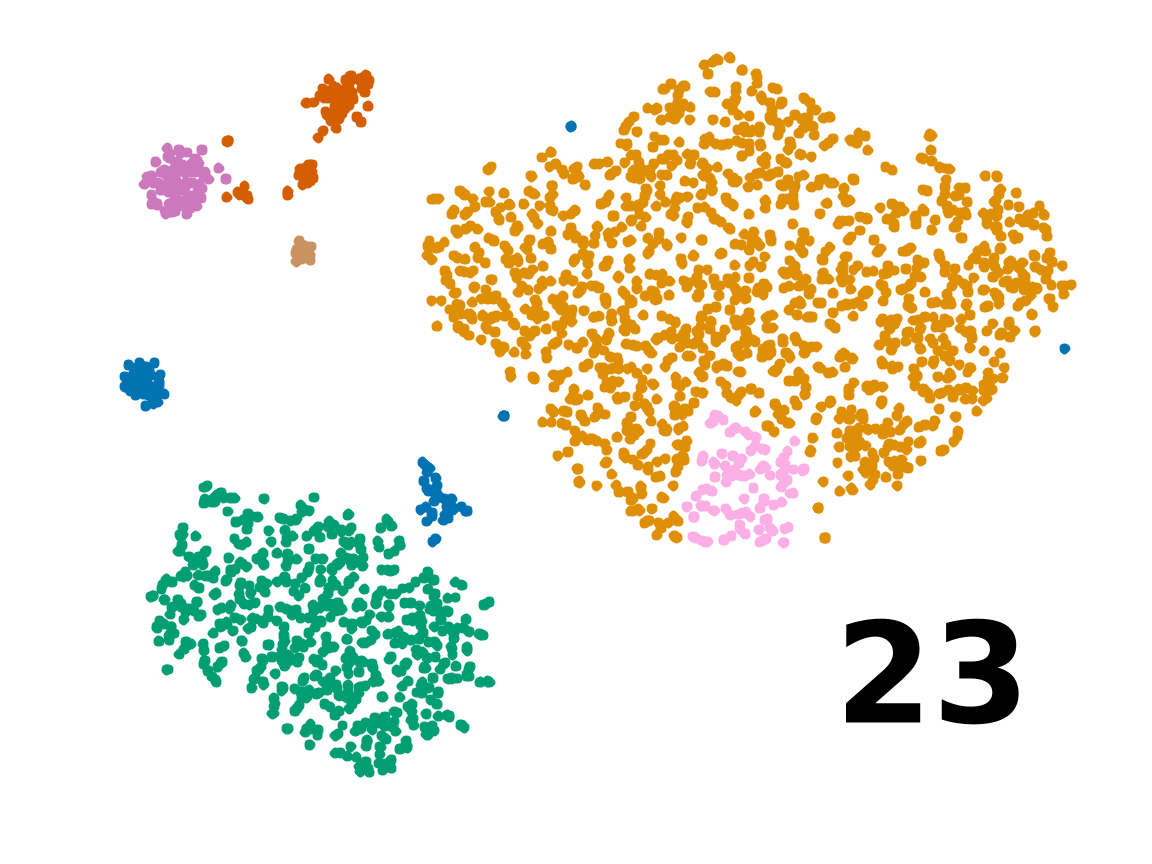} &\includegraphics[width=0.15\linewidth]{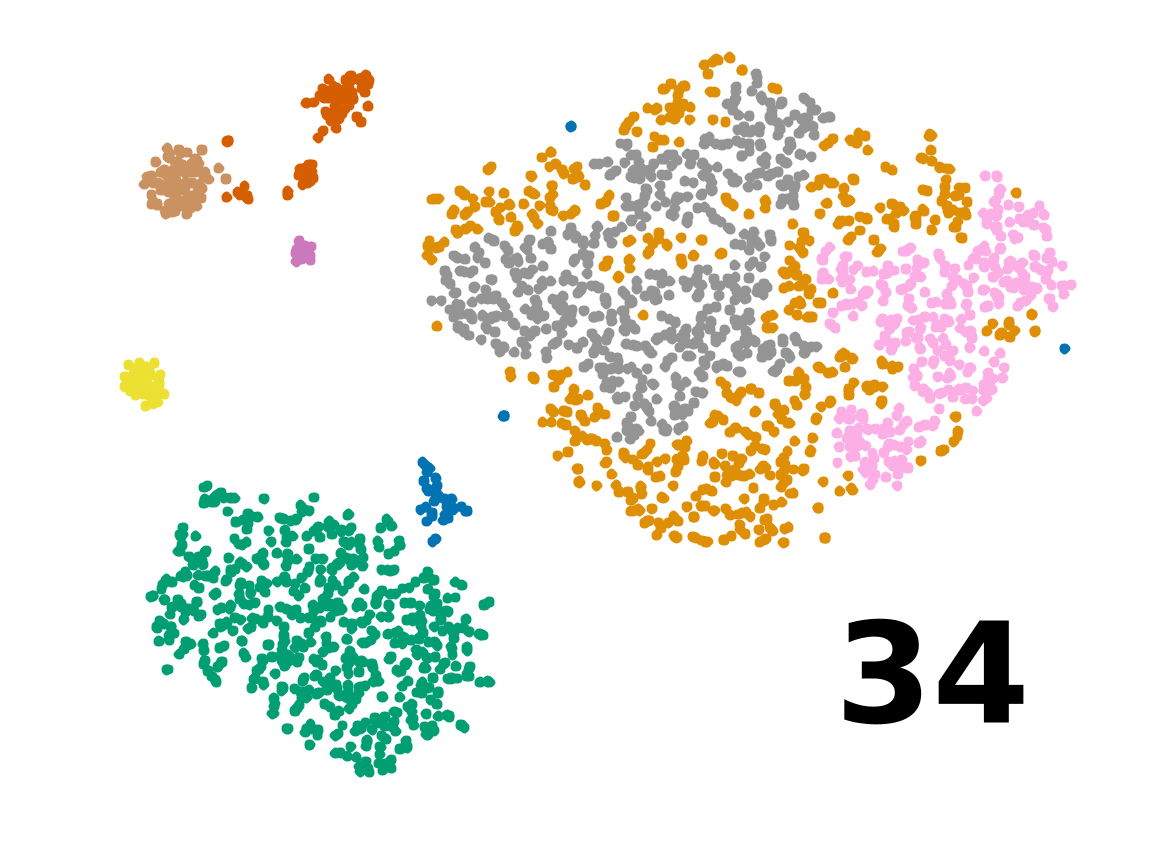} &\includegraphics[width=0.15\linewidth]{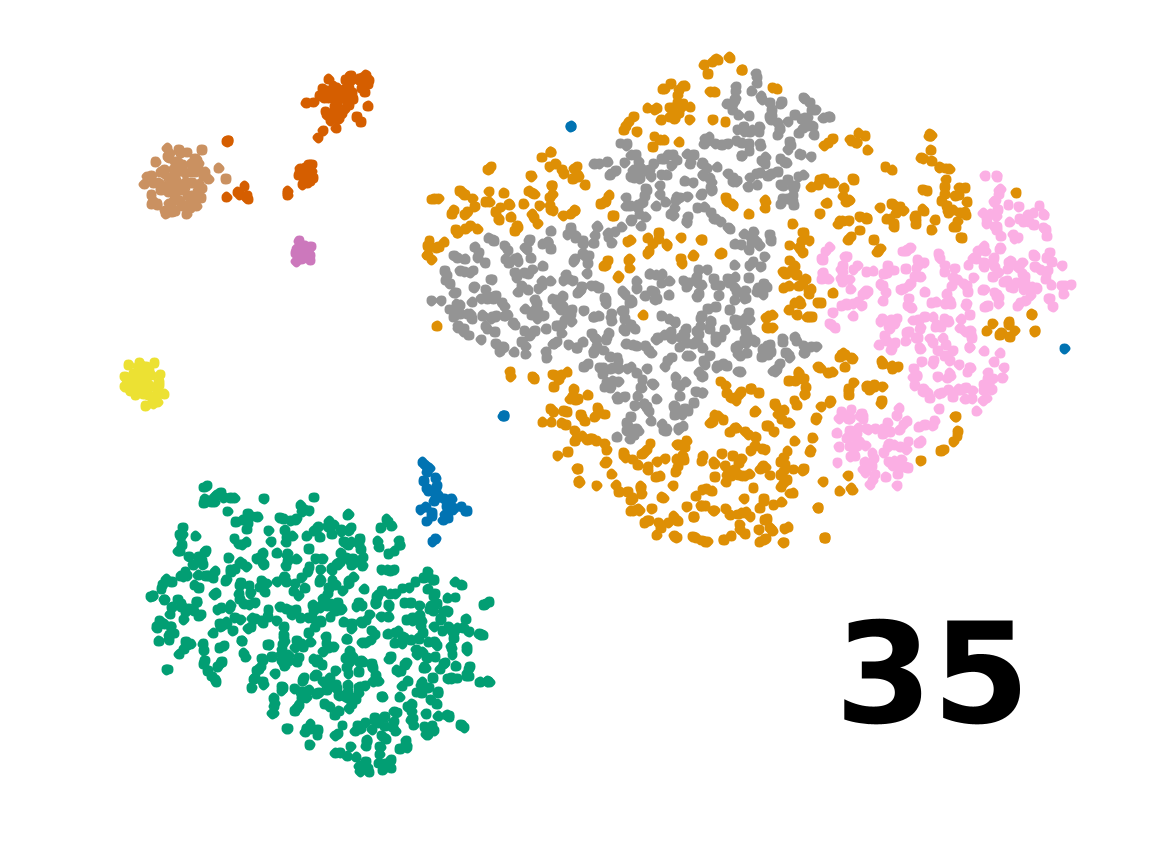} \\
        1.5 & \includegraphics[width=0.15\linewidth]{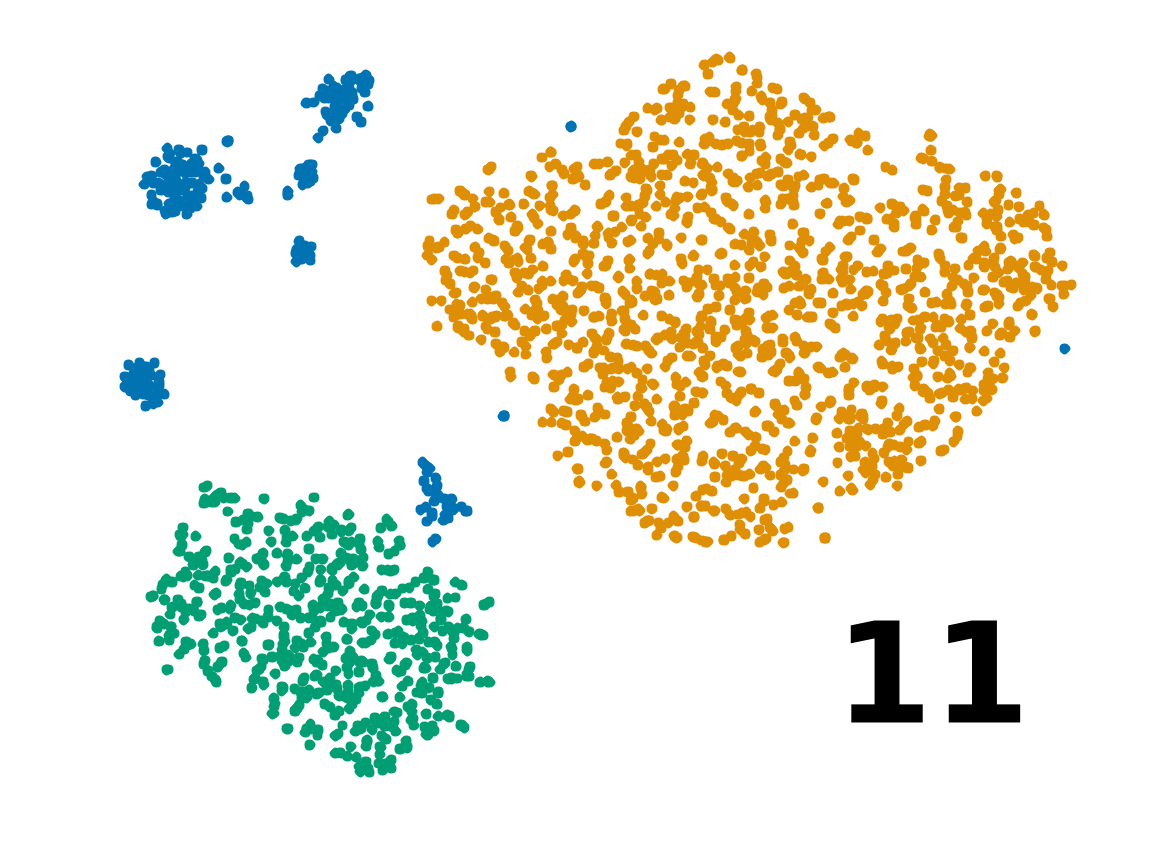}&\includegraphics[width=0.15\linewidth]{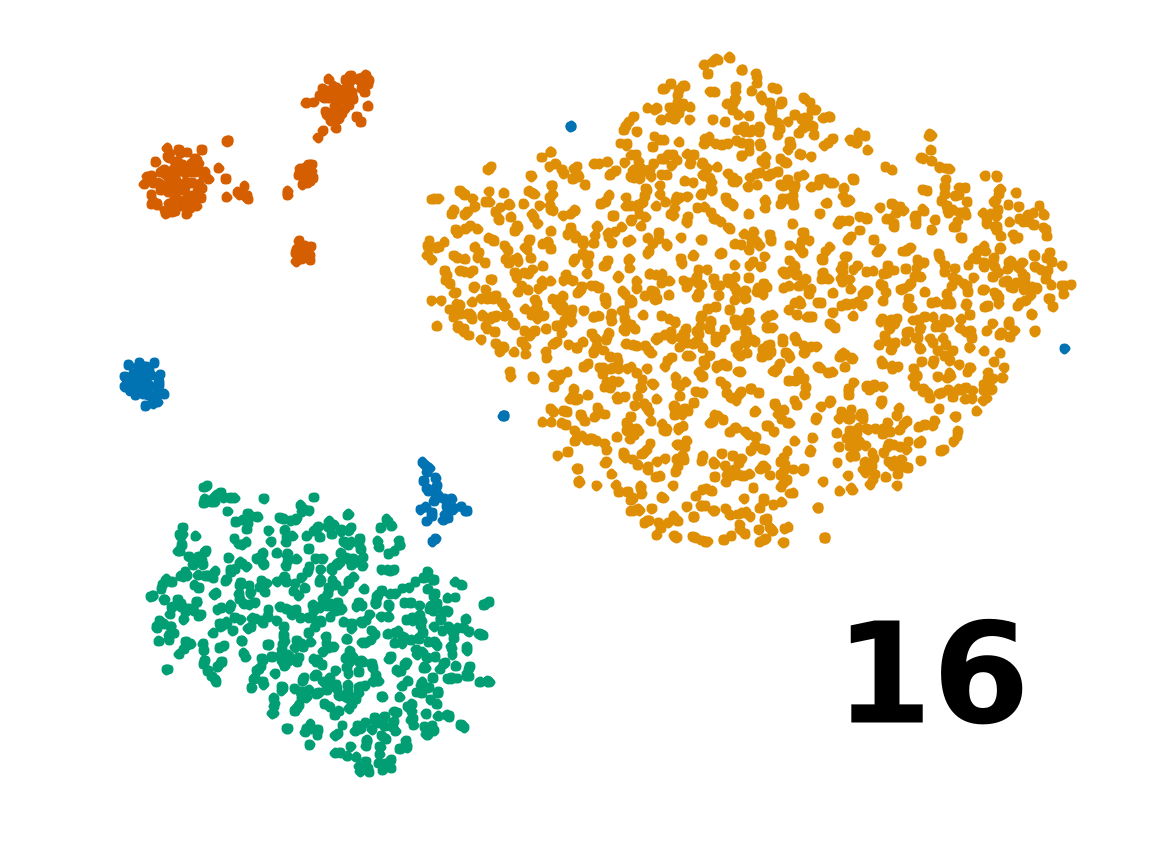} &\includegraphics[width=0.15\linewidth]{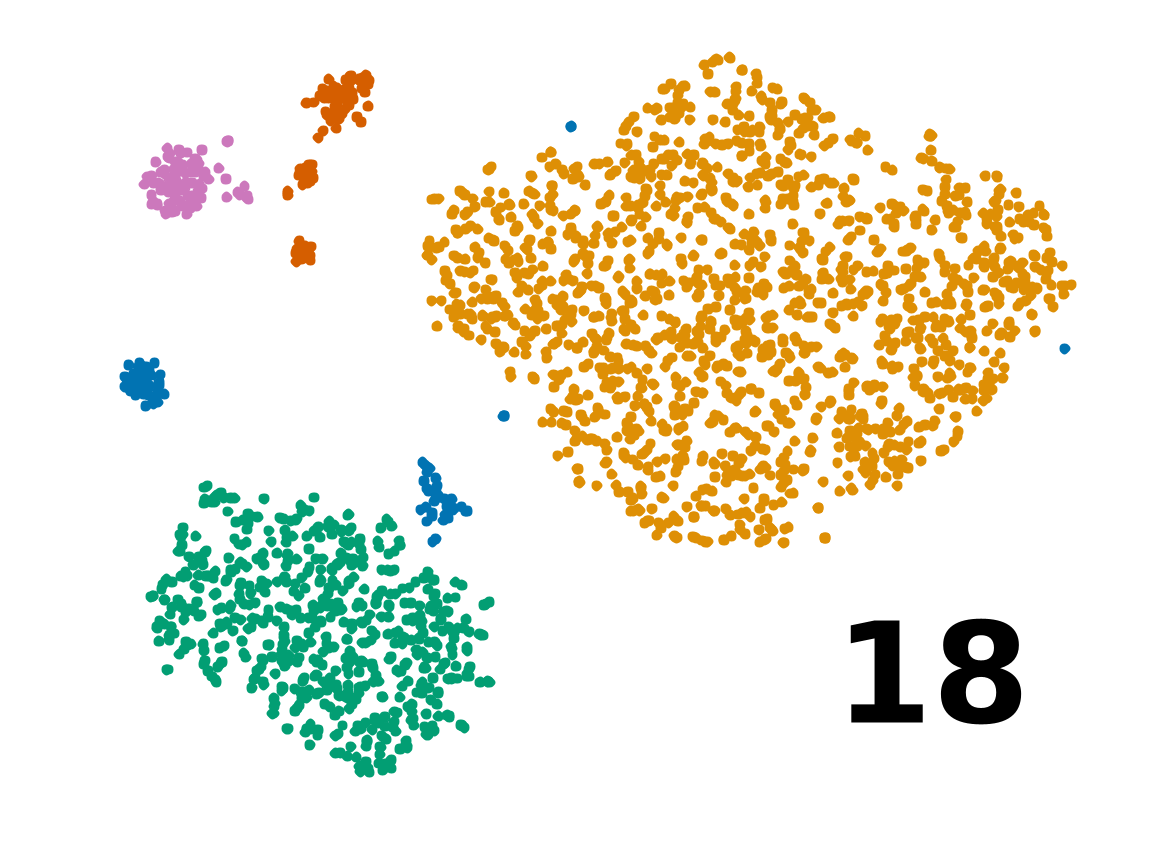} &\includegraphics[width=0.15\linewidth]{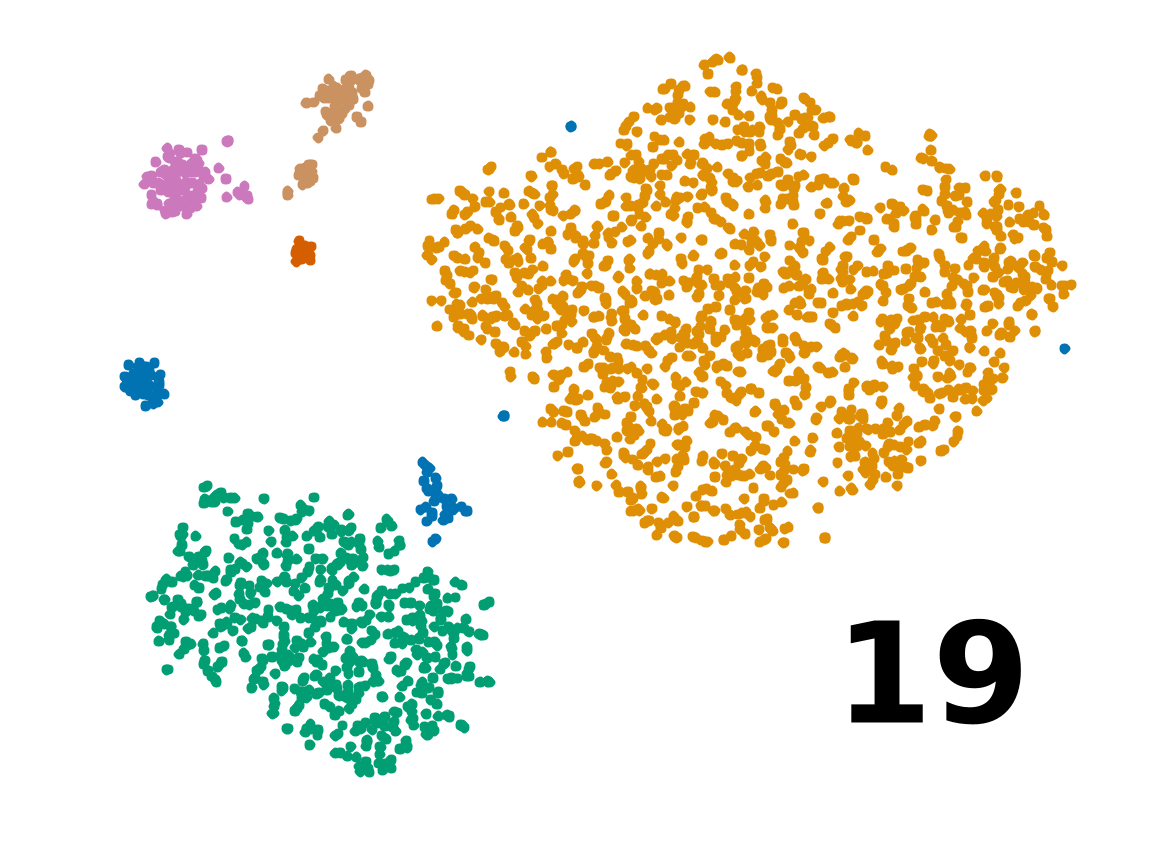} &\includegraphics[width=0.15\linewidth]{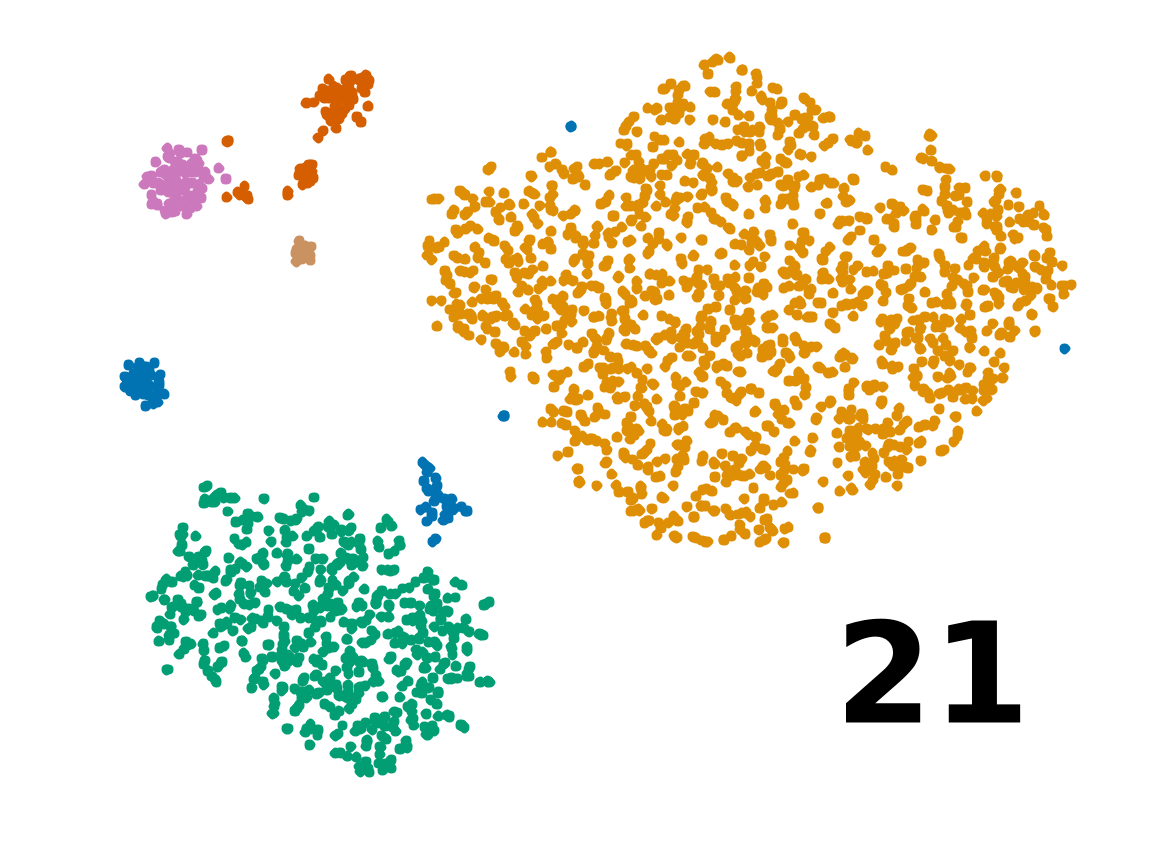} &\includegraphics[width=0.15\linewidth]{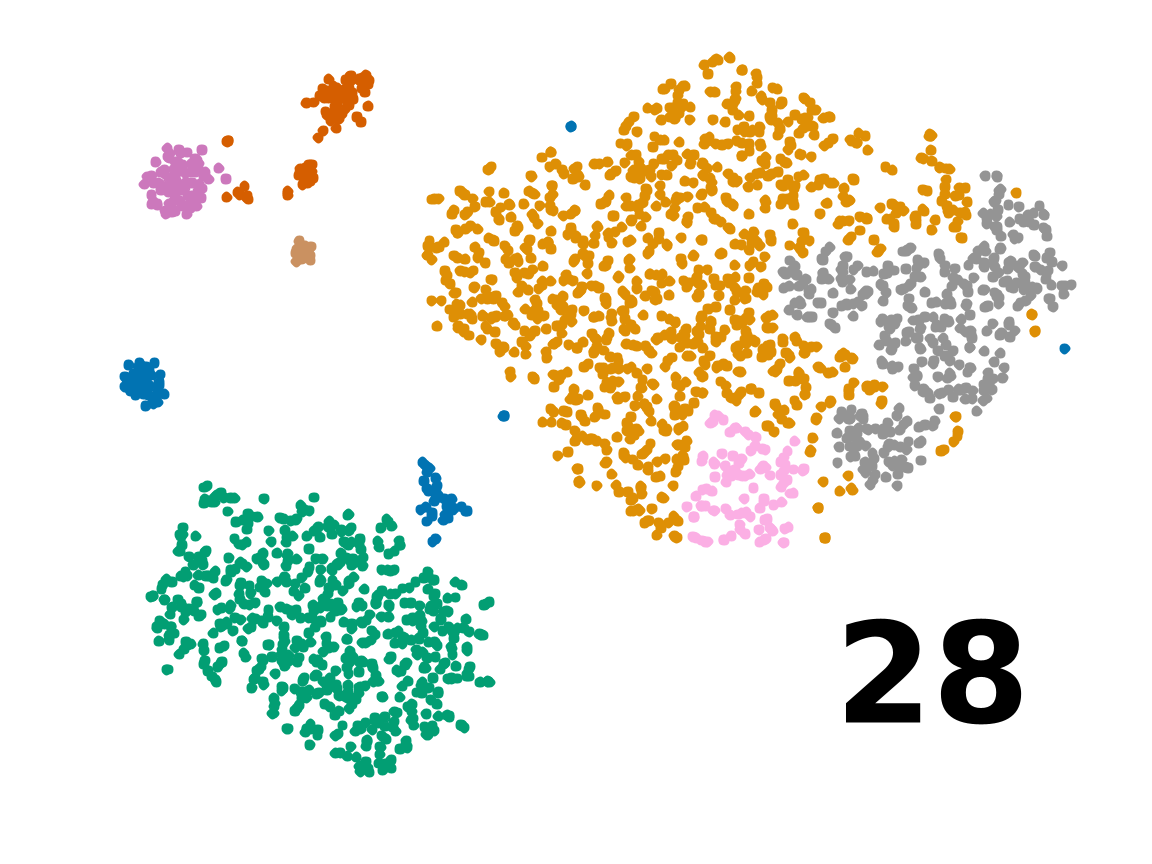} \\
        1.6 & \includegraphics[width=0.15\linewidth]{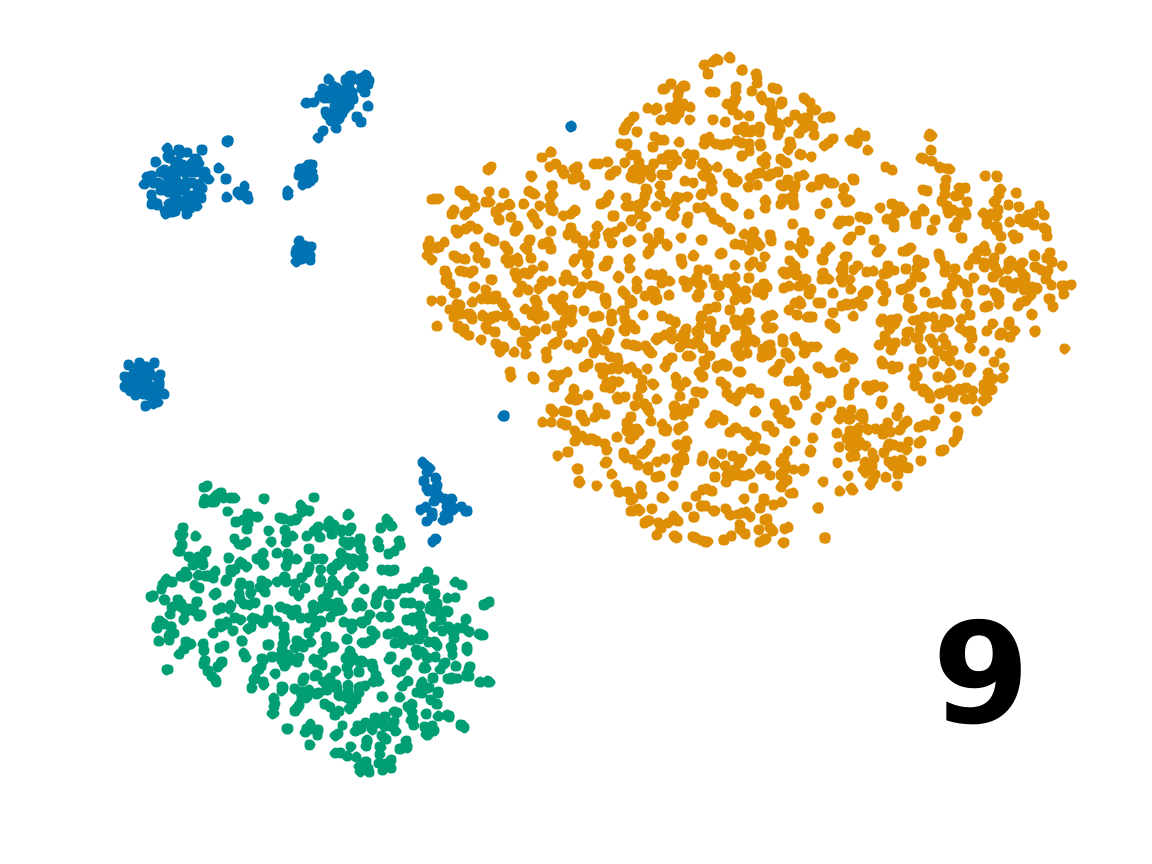}&\includegraphics[width=0.15\linewidth]{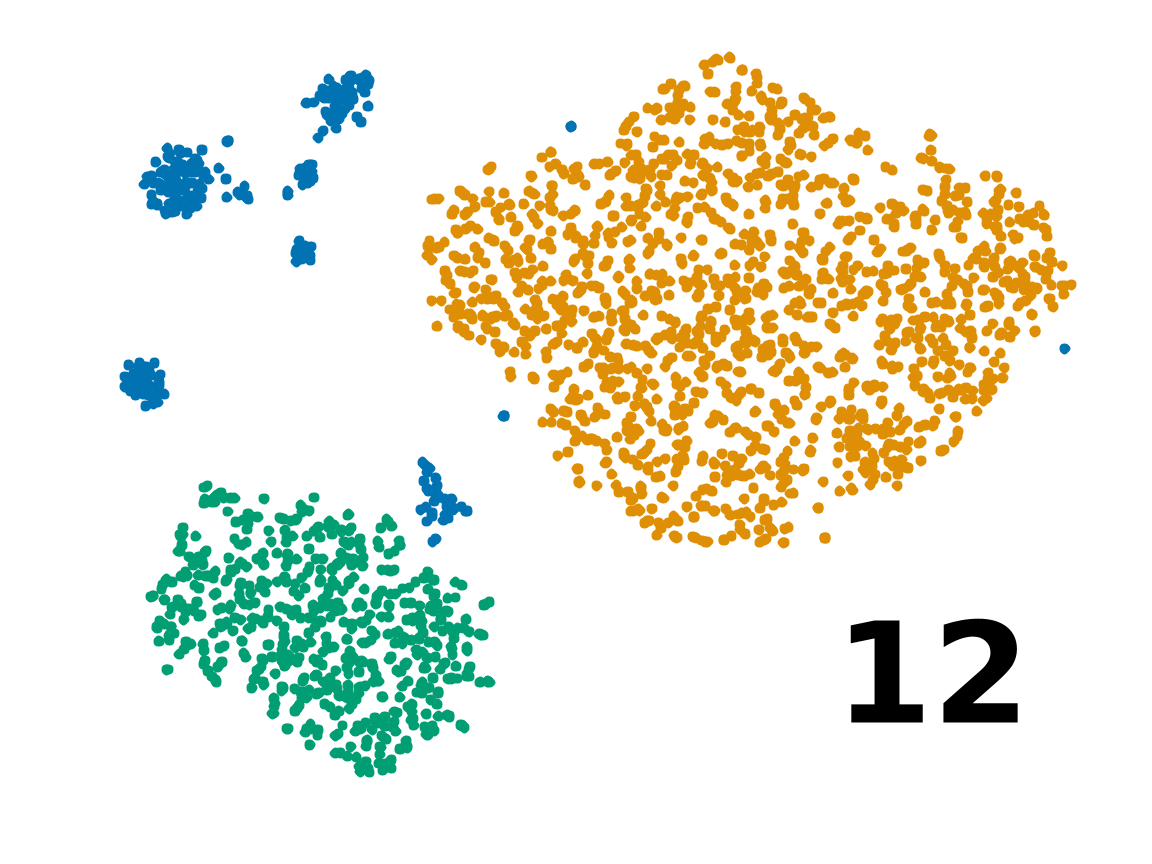} &\includegraphics[width=0.15\linewidth]{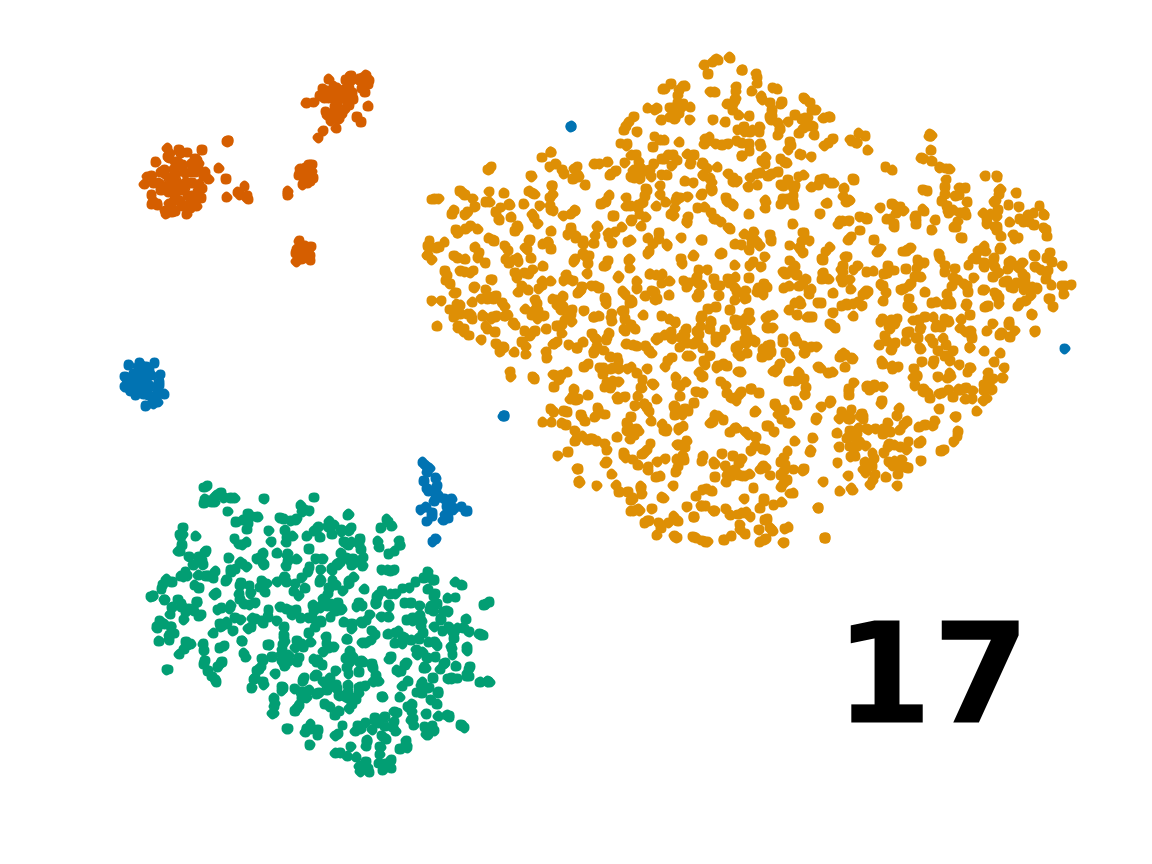} &\includegraphics[width=0.15\linewidth]{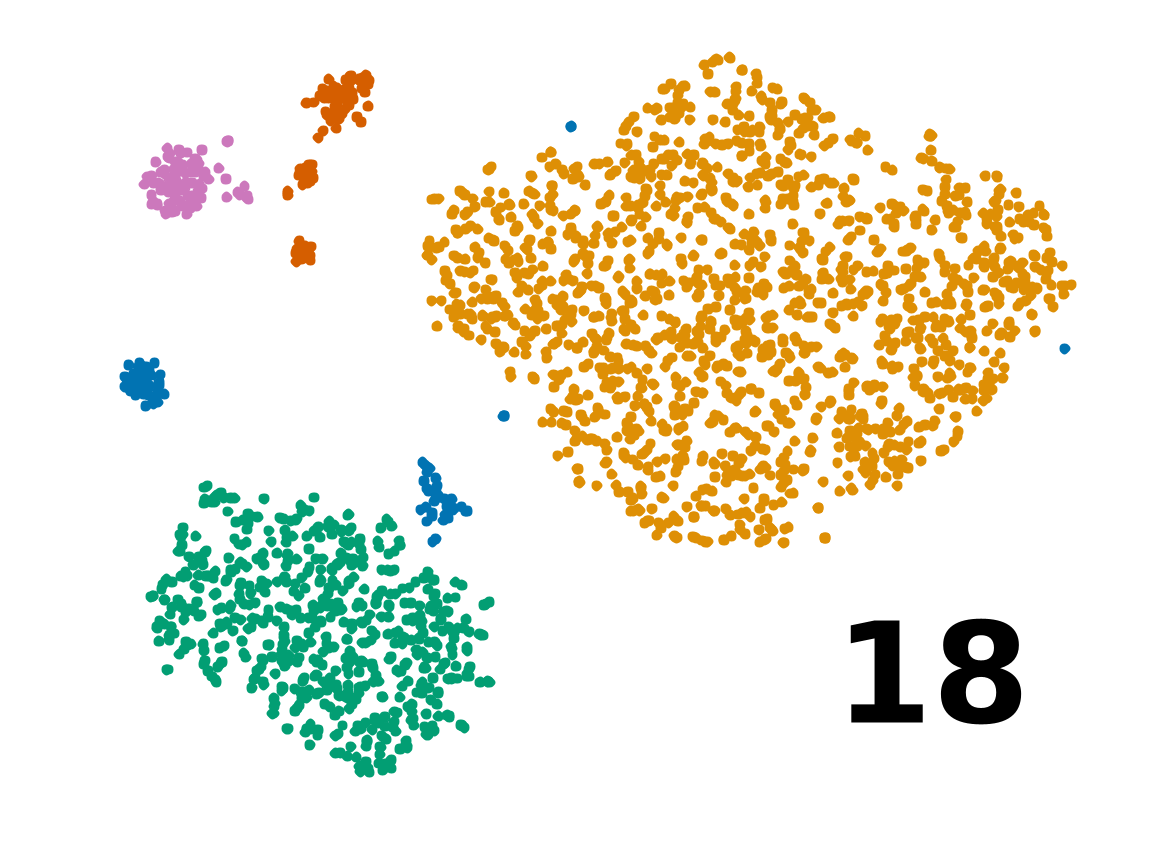} &\includegraphics[width=0.15\linewidth]{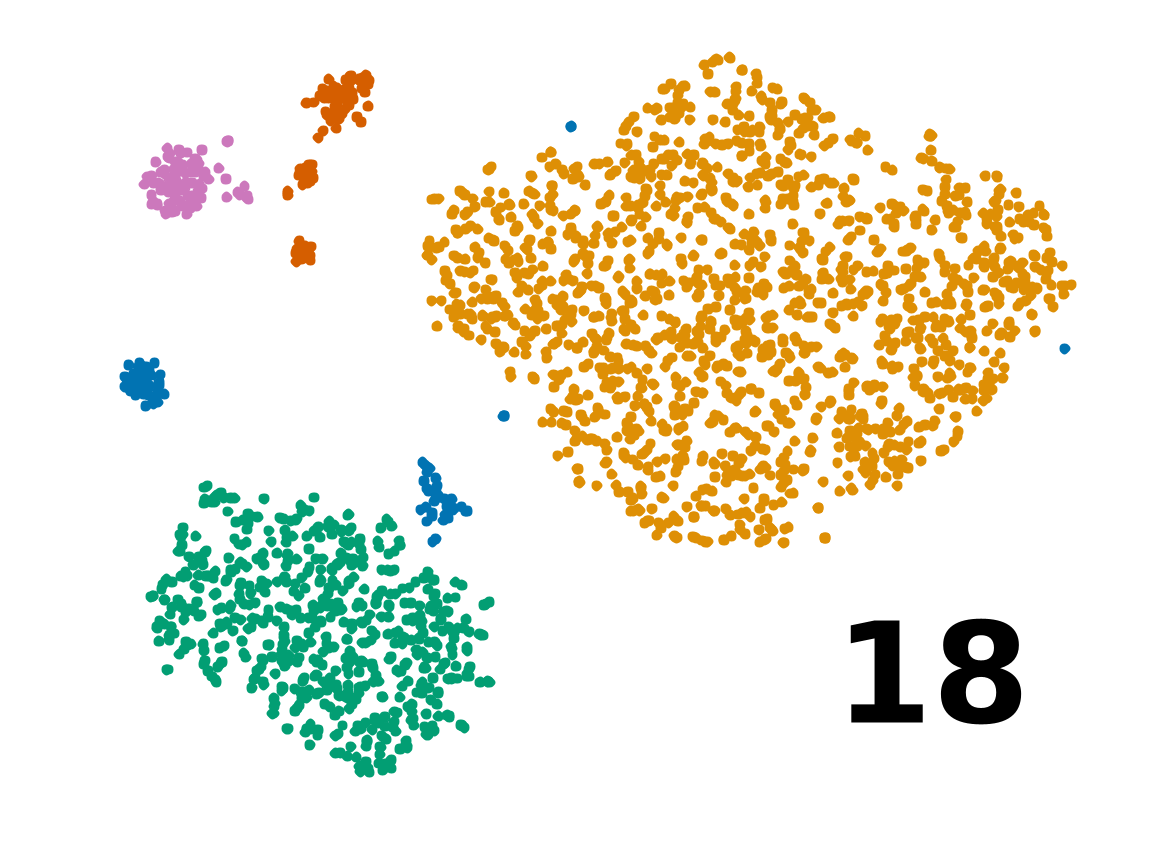} &\includegraphics[width=0.15\linewidth]{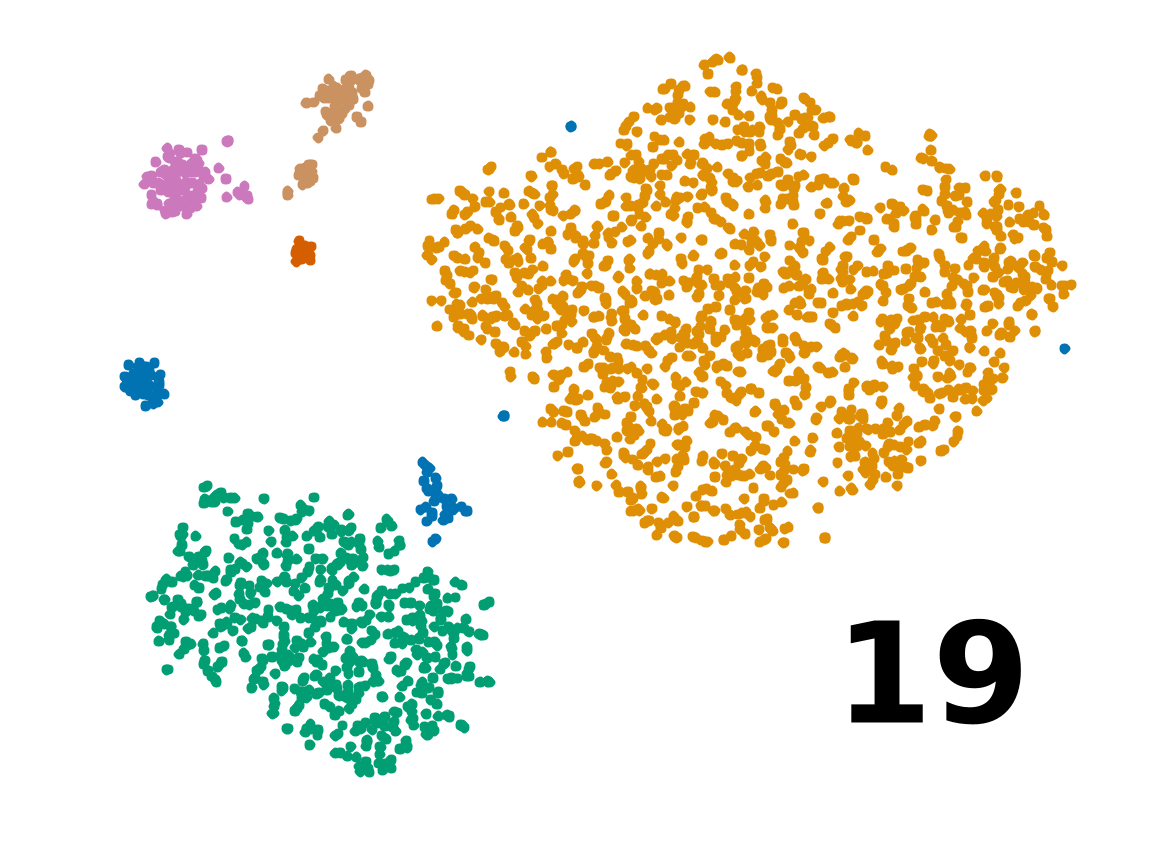} \\
        \hline
    \end{tabular}
    \caption{Hyper-parameter analysis on \Cyto. partitionings generated by \InfoClus\ under different $\alpha$ and $\beta$ are shown, along with the numbers of attributes selected for all clusters (bottom right corner). }
    \label{fig:sensitivity analysis}
    \vspace{-16pt}
\end{figure*}

Fig. \ref{fig:sensitivity analysis} displays the evolution of \InfoClus\ clusters and the total number of explanatory attributes selected for different values of $\alpha$ and $\beta$ ($\alpha$ varying from 250 to 2500, $\beta$ from 1.4 to 1.6).
The number of \InfoClus\ clusters increases when $\alpha$ increases, and decreases when $\beta$ increases.
Similarly, the number of selected attributes increases with $\alpha$ increases, and decreases when $\beta$ increases. 
This was to be expected given the definition of the \xratio\ ($R$) (Def. \ref{def:r}). $\alpha$ and $\beta$ both intervene in $R$'s denominator $\alpha + (\sum_{i=1}^r{\sum_{j=1}^{|e_i|}{|a_i^j|}})^\beta$. 
Increasing $\beta$ promotes sparser explanations.
When $\alpha$ increases, the weight of $(\sum_{i=1}^r{\sum_{j=1}^{|e_i|}{|a_i^j|}})^\beta$ decreases in the denominator, with similar effect to a decrease in $\beta$.

Nevertheless, 
partitionings does not change heavily when $\alpha$ and $\beta$ vary. It is a good signal for users that even they have few knowledge about \InfoClus , \InfoClus\ might also return a reasonable result. We also suggest users to select $\alpha$ between $n/10$ and $n$, and $\beta$ around $1.5$.



\subsection{Scalability Analysis  \label{sec:scalibility}}

We next turn to the analysis of \InfoClus' scalability.
\InfoClus' runtime can be divided into two components: initialization and searching for a good \pwx. The initialization proceeds: by i) computing embeddings, if none are user-provided; ii) computing a hierarchical clustering of the embeddings; iii) computing statistics used downstream for computing \xratio s (means and variances of candidate clusters), see Appendix~\ref{appendix:accelerate} for details. The search for the optimal \pwx\ then proceeds using these components.  

As users may impose an upper bound $t$ on total runtime, measuring the overall time spent on the search phase is meaningless. We instead list the time spent for initialization and describe the search phase by the time spent per iteration and per partitioning. The time spent per iteration corresponds to the average time spent to find the best $k$-cluster partitioning starting from a partitioning with $k-1$ clusters. The time spent per partitioning is the time used to select the top explanations of a partitioning with respect to the \xratio\ $R$.

Table \ref{tab:scalability} reports runtimes of \InfoClus\ components, computed on samples of increasing size drawn from the original \Cyto\ dataset.
The experiments were conducted on a machine equipped with a 13th Gen Intel(R) Core(th) i7-1365U (5.20 GHz) and 16.0 GB of RAM.

\begin{table}[t]
    \centering
    \begin{tabular}{|c|c|ccccccc|m{1.5cm}|}
    \hline 
    \multicolumn{2}{|c|}{Sample size ($\times10^{3}$)} &2.5 &5 &10 &20 &30 &50 &80 & \quad Trend\\
    \hline \hline
    \multirow[b]{2}{*}{ Time (s)} &\makecell{Initialization \\ rows 3-6 in Algo. \ref{algo:infoclus}} & 7.1 &15.6 &34.6 &71.8 &109 &190 &322& \begin{tikzpicture}
            \begin{axis}[
                width=3cm, height=2.5cm, 
                axis lines=middle, 
                xtick=\empty, ytick=\empty, 
                enlargelimits=false,
                xmin=0, ymin=0
            ]
                \addplot[mark=*,mark size=0.7pt, color=blue] coordinates {(2.5,7.1) (5,15.6) (10,34.6) (20,71.8) (30,109.0) (50,190.2) (80,322.0)};
            \end{axis}
        \end{tikzpicture} \\
        \cline{2-10}
    
    &\makecell{Avg. per iteration \\ rows 11-18 in Algo. \ref{algo:infoclus}}  &0.6 &1.4 &3.3 &6.1 &8.6 &16.7 &32.2 &\begin{tikzpicture}
            \begin{axis}[
                width=3cm, height=2.5cm, 
                axis lines=middle, 
                xtick=\empty, ytick=\empty, 
                enlargelimits=false,
                xmin=0, ymin=0
            ]
                \addplot[mark=*,mark size=0.7pt, color=blue] coordinates {(2.5,0.6) (5,1.4) (10,3.3) (20,6.1) (30,8.6) (50,16.7) (80,32.2)};
            \end{axis}
        \end{tikzpicture} \\
            \hline

    Time ($\times10^{-4}$s)&\makecell{Avg. per partitioning \\ rows 12-18 in Algo. \ref{algo:infoclus}}  &2.6 &2.8 &3.3 &3.0 &2.9 &3.3 &4.0 &\begin{tikzpicture}
            \begin{axis}[
                width=3cm, height=2.5cm, 
                axis lines=middle, 
                xtick=\empty, ytick=\empty, 
                enlargelimits=false,
                xmin=0, ymin=0
            ]
                \addplot[mark=*,mark size=0.7pt, color=blue] coordinates {(2.5,2.6) (5,2.8) (10,3.3) (20,3.0) (30,2.9) (50,3.3) (80,4.0)};
            \end{axis}
        \end{tikzpicture} \\
    \hline 
    \end{tabular}
    \vspace{2pt}
    \caption{Scalability of \InfoClus\ on samples from \Cyto\ with various size
    \label{tab:scalability}}
    \vspace{-24pt}
\end{table}

First, initialization is a rather costly process although it is only run once: 5 minutes are required to initialize a dataset with size of 80,000.
Second, we turn to the average times per iteration and partitioning in the \pwx\ search phase. These two quantities can be formally related: the cost per iteration is at most the product of the data size $n$ with the cost per partitioning\footnote{Partitionings are generated by splitting notes from the dendrogram generated by the hierarchical clustering. This dendrogram has at most $n-1$ nodes, including its root.}.
The time needed per partitioning seems to fluctuate as data size $n$ varies, with no obvious increase or decrease trend.
The absolute time for each partitioning is quite small with an order of magnitude of $10^{-4}$ seconds. 
Altogether, these results indicate that \InfoClus\ can credibly scale to problems of at least moderate size. 

\section{Discussion\label{sec:discussion}}

In this paper, we introduced \InfoClus, which finds a partitioning of an embedding that is cohesive in both the high-dimension and low-dimension space and also returns explanations in the form of attributes that stand out for each cluster. By constraining the solution with a hierarchical clustering, using greedy optimization with locally optimal cuts, 
we obtain an efficient and effective algorithm. By means of two hyperparameters, users can control how fine-grained (i.e., complex) the output is. We have found from the empirical results that the solutions can be helpful and that have distinct advantages over existing approaches.

The proposed method and experiments have several limitations that could be addressed in future work.
We do not know the computational complexity to find optimal \pwx s compatible with hierarchical clustering, it may be possible to identify better optimization strategies than greedy.
Also, the search strategy can lead to disconnected clusters because they can be nested. To prevent this, a different approach would be necessary.
Finally, it would be interesting to consider how to unify the \infoc\ for discrete and numeric variables, enabling the use of \InfoClus\ on mixed data. 

\begin{credits}
\subsubsection{\ackname} The research leading to these results has received funding from the Special Research Fund (BOF) of Ghent University (BOF20/IBF/117), from the Flemish Government under the ``Onderzoeksprogramma Artificiële Intelligentie (AI) Vlaanderen'' programme, from the FWO (project no. G0F9816N, 3G042220, G073924N). Funded by the European Union (ERC, VIGILIA, 101142229). Views and opinions expressed are however those of the author(s) only and do not necessarily reflect those of the European Union or the European Research Council Executive Agency. Neither the European Union nor the granting authority can be held responsible for them. For the purpose of Open Access the author has applied a CC BY public copyright licence to any Author Accepted Manuscript version arising from this submission.
EH is supported by a doctoral scholarship from the FWO (project number: 11J2322N)

\subsubsection{\discintname}
The authors have no competing interests to declare that are relevant to the content of this article. 
\end{credits}
%
%
%

\bibliographystyle{splncs04}
\bibliography{references}

\newpage
\section{Appendix}

\subsection{Accelerating Strategy. \label{appendix:accelerate}}
To increase the efficiency of the computation of \xratio\ and make full use of the dendrogram structure correlated with hierarchical clustering, we propose a strategy to recursively and efficiently compute mean and variance statistics for attributes over subtrees. The mean and variance statistics are key components in $KL$-divergence that are needed at every evaluation of the information content. We compute them efficiently as follows.

All single points in the dendrogram start with the mean and variance  equal to themselves and $0+\epsilon$ on all attributes, where $\epsilon$ is a small positive value used to avoid zero-variance. The mean and variance of intermediate nodes can be computed by the following method.
Assume we have two sets of data points, $X_1$ and $X_2$, of size $n_1$ and $n_2$, respectively. Let $\mu_1$ and $\mu_2$ be the means, and $\sigma^2_1$ and $\sigma^2_2$ be the variances of $X_1$ and $X_2$. When combining $X_1$ and $X_2$ into a new set $X$, the mean and variance of $X$ can be computed using
\begin{align}
&\mu = \frac{n_1*\mu_1+n_2*\mu_2}{n_1+n_2}\label{eqn:mean}\\
&\sigma^2 = \frac{n_1*\sigma^2_1+n_2*\sigma^2_2}{n_1+n_2}+\frac{n_1*n_2*(\mu_1-\mu_2)^2}{(n_1+n_2)^2}\,.\label{eqn:var}
\end{align}
Similar to above, the means and variances of two sets generated by splitting a subset out of a set can also be computed in a similar fashion.

\newpage
\subsection{Complementary materials. \label{appendix:complement}}

\begin{figure}
    \centering
    \begin{tabular}{cc}
        \includegraphics[width=0.45\textwidth]{figs_ecml/cytometry/labelled_by_RVX.png} & \includegraphics[width=0.45\textwidth]{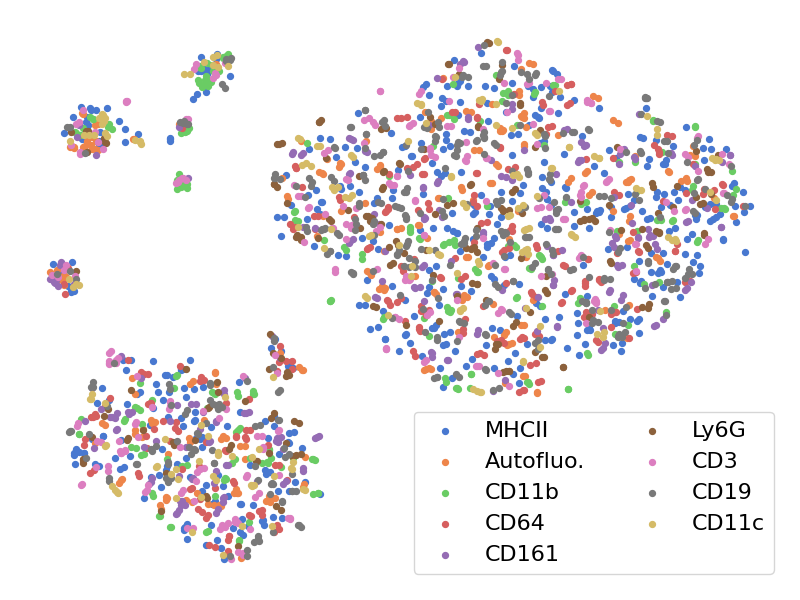} \\
        (a) $RVX$ & (b) $RV_\sigma X$
    \end{tabular}
    \caption{Methods from Thijssen et al. \cite{thijssen2023}: \underline{(a)}: Relative Value eXplanation, colors points
    in embeddings based on the attribute which has outlier values over their neighborhoods. \underline{(b)}:  colors points
    in embeddings based on the attribute which has different variance over their neighborhoods compared with global variance.}
    \label{fig:appen_rvx}
\end{figure}

\begin{figure}[h]
    \centering
    \begin{tabular}{c}
         \includegraphics[page=1, width=\textwidth]{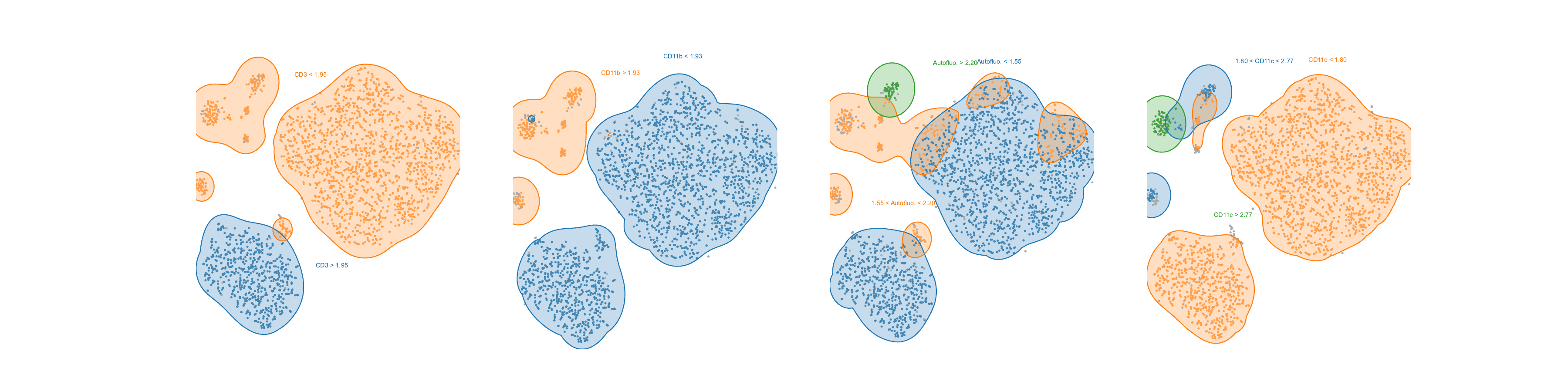}  \\
         (a) Contrastive panels \\
         \includegraphics[page=2, width=\textwidth]{figs_ecml/cytometry/cytometry_2500_descriptive.pdf}\\
         \includegraphics[page=3, width=\textwidth]{figs_ecml/cytometry/cytometry_2500_descriptive.pdf}\\
         (b) Descriptive panels
    \end{tabular}
    \caption{VERA \cite{policar2024}: \underline{(a)}: shows the most 4 interesting contrastive panels, explaining clusters by single attribute, selected by VERA, \underline{(b)}: shows the most interesting 4 descriptive panels, explaining clusters by multiple attributes, selected by VERA.}
    \label{fig:appen_vera}
\end{figure}

\end{document}